%% file: vmsr-corl.tex
\documentclass{article}

\usepackage[final]{corl_2019} 
\usepackage{times}
\usepackage{epsfig}
\usepackage{graphicx}
\usepackage{amsmath}
\usepackage{amssymb}

\usepackage{booktabs}
\usepackage{shortbold}
\usepackage{blindtext}
\usepackage{lib}
\usepackage[shortlabels]{enumitem}
\setlist[enumerate]{leftmargin=*}
\usepackage{alias}
\usepackage{cite}
\usepackage{epsfig}
\usepackage{graphicx}
\usepackage{balance}
\usepackage{rotating}
\usepackage{wrapfig}
\usepackage[font=small,labelfont=bf]{caption}
\usepackage{subcaption}
\usepackage[toc,page,header]{appendix}

\title{Learning Navigation Subroutines from \\ Egocentric Videos}
\author{Ashish Kumar$^1$ \; Saurabh Gupta$^{3}$ \; Jitendra Malik$^{1,2}$ \\
$^1$UC Berkeley \; $^2$Facebook AI Research \; $^3$UIUC \\
{\small \texttt{ashish\_kumar@berkeley.edu, saurabhg@illinois.edu,
malik@eecs.berkeley.edu}}}

\begin{document}
\maketitle


\begin{abstract}
Planning at a higher level of abstraction instead of low level torques improves the 
sample efficiency in reinforcement learning, and computational efficiency in classical planning.  We propose a method to learn such hierarchical abstractions, or subroutines from egocentric video data of experts performing tasks.  We learn a self-supervised inverse model on small amounts of random interaction data to pseudo-label the expert egocentric videos with agent actions. Visuomotor subroutines are acquired from these pseudo-labeled videos by learning a latent intent-conditioned policy that predicts the inferred pseudo-actions from the corresponding image observations.  
We demonstrate our proposed approach in context of navigation, and 
show that we can successfully learn consistent and
diverse visuomotor subroutines from passive egocentric videos.  We
demonstrate the utility of our acquired visuomotor subroutines by using them
as is for exploration, and as sub-policies in a hierarchical RL
framework for reaching point goals and semantic goals. We also demonstrate
behavior of our subroutines in the real world, by deploying them on a real
robotic platform. Project website: \href{https://ashishkumar1993.github.io/subroutines/}{https://ashishkumar1993.github.io/subroutines/}.
\end{abstract}


\input{intro}
\input{related}
\input{approach}
\input{subroutine_expts}

\input{experiments}

\input{conclusion}

\clearpage
\acknowledgments{Authors would like to thank Allan
Jabri, Shiry Ginosar, Devendra Singh Chaplot, Ashvin Nair and Angjoo Kanazawa for feedback on the manuscript. This work was supported by Berkeley DeepDrive.}


\bibliography{refs} 
\newpage

\renewcommand{\appendixpagename}{\centering Supplementary Material}

\begin{appendices}
\input{supp.tex}
\end{appendices}

\end{document}

%% file: intro.tex
\section{Introduction}
Every morning, when you decide to get a cup of coffee from the kitchen, you
go down the hallway, turn left into the corridor and then
enter the room on the right. Instead of deciding the exact muscle torques,
you reason at this higher level of abstraction by composing these reusable lower
level \textit{visuomotor subroutines} to reach your goal. These visuomotor subroutines are classically known as operators in STRIPS planning \citep{fikes1971strips}, or more recently as options in RL \citep{sutton1999between}. Once these subroutines are learned, they can be composed to solve novel tasks, e.g. exiting the building, finding an object, etc, enabling an agent to quickly learn new tasks by simply learning how to compose them together.   

These subroutines, or short-horizon policies with consistent behavior, can be manually designed as done in classical robotics or STRIPS, or can be learned through interaction by training a hierarchical agent through reward-based reinforcement learning. Learning through environment interactions is extremely slow, making it prohibitively expensive to operationalize in the real world. We propose a third way of learning these subroutines by using imitation learning on egocentric videos of experts performing tasks. We expect that these videos contain subroutines that have been appropriately combined to solve some tasks, and an appropriate clustering algorithm can be used to isolate and extract these subroutines. For example, in indoor navigation, such clusters could be exiting doors, walking down hallways; and for driving, they could be following a lane, changing lanes, etc. Once isolated, these subroutines just need to be fine tuned through reward-based RL for downstream tasks, which is very sample efficient, as we show in our experiments. 

To imitate an expert at a visuomotor task, we need to know both the perceptual input to the expert and the action taken. One way to do this is by instrumenting the agent to collect the perceptual input as well as the action executed, as done in autonomous driving \citep{xu2017end}. However, this limits the scalability of the data collection procedure. To scale it up, we could instead learn from videos of people performing tasks uploaded on websites such as YouTube. Such videos fall in two categories: first person (egocentric) or third person video. 
Third person videos have the benefit of having action information but don't have the
perceptual input. Skills learned from such videos don't depend on the perceptual input \citep{peng2018sfv,peng2018deepmimic}. But our focus is on navigation tasks for which the perceptual input is important. We thus face the opposite challenge when using egocentric video of a person performing a task (\eg biking with a head mounted GoPro
camera). The perceptual input to the agent is available, 
but the action information is typically missing. In this paper, we
will address this case and demonstrate our technique in the navigation domain.  

\begin{figure}
\centering
\includegraphics[width=1\linewidth]{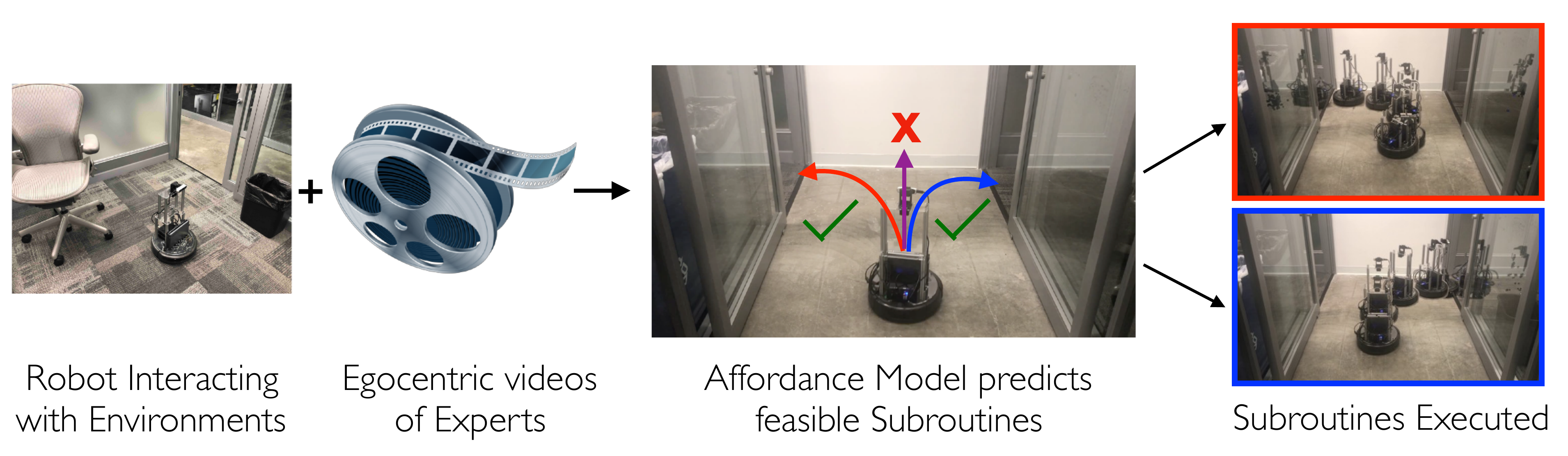}
\caption{\textbf{Approach Overview:} We propose an approach that combines
learning from direct environmental interaction, with learning from first-person
videos collected over the Internet. Inverse models built using a small number
of environmental interactions are used to interpret videos, and learn \textit{affordances} (what can I do) and \textit{subroutines} (how can I do it). Affordances predict which subset of subroutines are feasible given the current image. These subroutines can then be executed in novel environments.}
\figlabel{motivation}
\end{figure}

We start with egocentric videos of experts navigating to achieve some task unknown to us. Given these videos, we want our robot to learn meaningful and useful subroutines.

We obtain action labels for the egocentric videos by training a self-supervised inverse model on random interaction data. Egocentric videos can then be pseudo labeled by running the inverse model on consecutive image pairs. Note that the action space of the experts might be different from the action space
of the robot (for example if we were to download egocentric 
navigation videos available online). Hence, these pseudo-action labels are not the actual action taken, but an action imagined by the robot to make the transition between the
observations as closely as possible. (\secref{pseudolabel}).

Once we have the pseudo action labels, we need to label the subroutines in the videos and learn a controller which can be used in downstream tasks. For this, we slice up a trajectory into smaller sub-trajectories. The slicing length is a hyper parameter, that controls the complexity of the subroutines learned (longer sub-trajectories will lead to more complex subroutines). We then encode each sub-trajectory into a discrete latent variable which should be predictive of the action given the video frame, for every frame in the sub-trajectory. (\secref{vmsr}).

To effectively use the learned subroutines in downstream tasks, we must also infer which subroutines can be applied where. For this, we additionally train an affordance model to predict which subroutines \textit{can} be invoked for a given input image from our repertoire of learned subroutines. We do this by predicting the inferred one-hot latent encoding of the trajectory from the first image. (\secref{vmsr}).

We evaluate our learned subroutines and the affordance model on downstream navigation tasks, which are unknown to our method during the subroutine learning phase. We show that our learned subroutines can be composed together for zero-shot exploration in novel environments with a 50\% improvement in exploration over several learning and non-learning baselines. We also evaluate our learned subroutines on downstream point navigation and area goal tasks. We fine-tune our affordance model and subroutines through reward-based RL and observe a $4\times$ improvement in learning sample complexity over alternate initializations. (\secref{experiments}).

%% file: related.tex
\section{Related Work}
\seclabel{related}
\textbf{Classical Navigation.}
Classical approaches to navigation employ geometric reasoning to solve the
task \citep{Lavalle06book, thrun2005probabilistic}.
While most works optimize in the base action space of the agent, few works
employ hand-crafted motion primitives to speed up planning
\citep{hauser2008using}.  Dynamic Motion Primitives (DMPs) propose a
framework for specifying macro actions \citep{schaal2006dynamic}, which can be learned from a demonstration \citep{ijspeert2013dynamical} for a specific macro action. However, the set of macro actions are still manually specified. In contrast, these behaviors automatically emerge 
as a consequence of our algorithm.    

\textbf{Learned Navigation.}
Recent learning based-efforts use reinforcement learning or imitation learning
to learn policies for solving specific locomotion, navigation or manipulation tasks \citep{zhu2016target,
mirowski2016learning, gupta2017cognitive, sadeghi2019divis, sadeghi2016cad2rl, levine2016end, levine2018learning}.
While these works learn to leverage
high-level semantics, they still directly operate in the base action
space of the robot. Learned skills are task and environment specific, and it
takes a large number of interaction samples to even solve the same task in a
new environment (in navigation \citep{zhu2016target, mirowski2016learning} as well as in   manipulation \citep{levine2016end, levine2018learning}). To address this, some works \citep{pathak2017curiosity} use intrinsic rewards such as prediction error. However, these approaches don't distill out composable skills to solve novel tasks. Works like \citep{eysenbach2018diversity} distill out composable skills but do not scale to realistic setups as we show in our experiments.   
Moreover, as all training signal is derived from interaction with the
environment, skill acquisition is extremely expensive. In fact, our experiments
show that our use of passive videos for learning skills is more sample
efficient and results in better performance than such purely interactive
approaches.  

\textbf{Learning from State-Action Trajectories.} Several works
use learning from demonstration in scenarios where they have access to both the observations and the ground truth actions to solve the task 
\citep{argall2009survey, billard2008robot}. Works like
\citep{hausman2017multi} extend these formulations to work with trajectory
collections that have multiple modes.  However, this line of work relies on
ground truth action labels. In contrast, we only assume
observation data (without paired actions), and evaluate on novel tasks
in novel environments.

\textbf{Learning from State Only Trajectories.}
Contemporary works \citep{aytar2018playing, torabi2018behavioral,
edwards2018imitating} study the problem of learning from state only
trajectories, similar to our work here. However, all of these works only study
the scenario where the agent solves the task in \textit{exactly} the same
environment that they have state-only demonstrations for. In contrast,
we do not assume access to environments for which we have videos for, making
learning more challenging and rendering these past techniques
ineffective. Additionally, our goal is to learn subroutines that work in
previously unseen environments, which goes beyond the focus of these
works. Works like  \citep{pathak2018zero, finn2017one, yu2018one} focus on imitation from visual data and learn a monolithic policy from expert data for the task at hand. We focus on learning composable subroutines which can then be used in several downstream tasks.

\textbf{Sub-policies and Options in Hierarchical RL.} Hierarchical RL is an active area of research \citep{dayan1993feudal, barto2003recent, sutton1999between}, with a number of
recent papers (such as \citep{vezhnevets2017feudal, levy2017hierarchical}).
These works acquire sub-policies in a top-down manner while interacting with the environment to solve a reward based task. Our approach on
the other hand investigates a bottom-up development of subroutines, and can learn from relatively inexpensive unlabelled passive data. Our learned
subroutines are complementary to these frameworks and can be used to initialize any of these top-down HRL methods to accelerate learning.

\textbf{Affordance Learning from Videos.} Researchers have studied affordance
learning from Internet videos \citep{fouhey2014people} by leveraging
YouTube videos to learn about affordances. While this is a great first
step, it does not learn a controller that can be used in downstream tasks. 
In contrast, we learn a controller for each subroutine for the
the specific robot at hand, allowing immediate deployment.

%% file: approach.tex
\section{Pseudo-Labeling Egocentric Expert Videos}
\seclabel{pseudolabel}
We need action labels on expert videos to learn a controller for downstream tasks. 
We build a self-supervised inverse model which takes two consecutive image observations, $o_t$ and $o_{t+1}$, and predicts the action $\hat{a}$ which the agent took to transition from $o_{t}$ to $o_{t+1}$ . This inverse model is then used to pseudo-label egocentric videos of experts. However, since the videos may come from diverse sources (e.g. internet videos), the expert uploading the video may have a different action space than $S$. To handle this mismatch, we expect the inverse model to predict the action which $S$ \textit{should have taken} to go from $o_{t}^e$ to $o_{t+1}^e$ and not the action actually taken by the expert. 

\begin{wrapfigure}[17]{R}{0.6\textwidth}
\centering
\vspace{-1.45cm}
\insertWL{1.35}{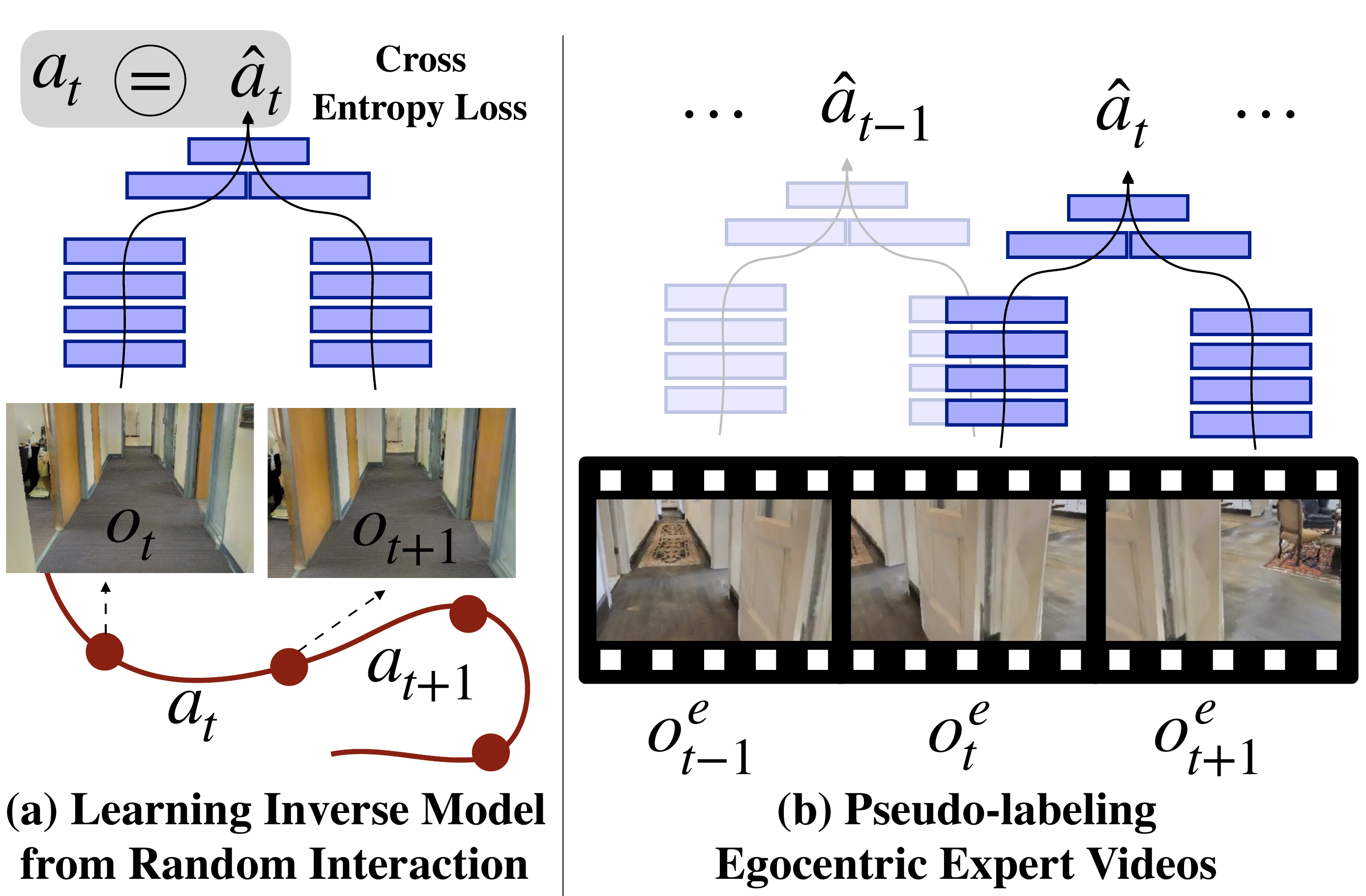}
\caption{\textbf{Pseudo-labeling:} \textbf{(a)} We execute random actions in
 an environment to obtain image-action sequences $\{\ldots, o_t, a_t,
o_{t+1} \ldots\}$. We use triplets $(o_t, a_t, o_{t+1})$ to train an inverse
model to predict action $a_t$ given consecutive images $o_t$,
$o_{t+1}$. \textbf{(b)} We use this inverse model to pseudo-label egocentric
videos of navigating agents. Grey circles with $=$ sign represent cross-entropy loss.}
\figlabel{pseudolabelling}
\end{wrapfigure}

\subsection{Inverse Model}
\seclabel{inverse}
We first build a self-supervised one-step inverse model $\psi$ \citep{jordan1992forward,agrawal2016learning} for the agent from random interaction data in an environment (simulation environment or the real world). More concretely, given a pair of consecutive image observations, $o_{t}$ and $o_{t+1}$, 
we train the model $\psi$ to predict
action $a_{t}$ that was executed to transition from image $o_{t}$ to $o_{t+1}$ as follows: $\hat{a}_{t} = \psi(o_{t},o_{t+1})$.
The agent $S$ collects data \{\ldots, $o_{t}, {a}_{t}, o_{t+1}, {a}_{t+1}$ \ldots\} to
train $\psi$ by sampling $a_t$ uniformly from \{{\small\texttt{left, right,
forward}}\} and executing it in the environment conveying 
it from $o_{t}$ to $o_{t+1}$.
\subsection{Pseudo-Labeling Video Data}
\seclabel{video}
We then use this learned inverse model $\psi$ to pseudo-label the dataset
$\mathcal{D}$ of egocentric expert videos. Given a sequence of images $\{o_1^e, o_2^e, \ldots \}$ from a video, we evaluate $\psi$ on consecutive pairs of images to obtain $\hat{a}$ = $\psi(o_{t}^e,o_{t+1}^e)$. We use observations as a means of implicitly mapping equivalent actions between agents. This
generates a pseudo labeled dataset $\mathcal{\hat{D}}$, that contains 
image action sequences, $\{o_1^e, \hat{a}_1, o_2^e, \hat{a}_2, \ldots, o_T^e\}$.

\section{Visuomotor Subroutines and Affordances}
\seclabel{vmsr}
We formally define subroutines and the affordance model for an agent $S$ as $\langle \alpha, \{\pi_i\}_{i=1..N} \rangle$, where $N$ is the number of subroutines available to the $S$, $\pi_i$ is the $i^{th}$ subroutine and $\alpha$ is the affordance model. The affordance model predicts the probability distribution $\tilde{p_t}$ given an input observation $o_t$, where $(\tilde{p}_t)_i$ is the probability that $\pi_i$ is applicable given the observation $o_t$. Each subroutine $\pi_i$ is a closed loop policy which takes the current observation $o_t$ and predicts a distribution over actions $\tilde{a_t}$. Thus,	
$\tilde{p_t} = \alpha(o_t)$ and $\tilde{a_t} = \pi_i(o_t)$.

We isolate these subroutines from $\mathcal{\hat{D}}$ by clustering them to improve the action prediction accuracy of the visuomotor trajectories in $\mathcal{\hat{D}}$. Intuitively, if the observation contains a T-junction with a possible left and right turn, they need to be clustered separately to unambiguously predict the future given the observation and the cluster id.

To effectively use these subroutines, we learn another model to infer which subroutines are applicable in what scenario. For example, given an image of a hallway with two doors, one on the left and other on the right, the model should learn to assign high probabilities to both \textit{go into left door} and \textit{go into right door} subroutines, whereas when there are no doors, it should simply peak on
the subroutine \textit{go down a hallway}. 

\subsection{Learning Formulation}
\seclabel{conditional}
We slice up each trajectory (from the pseudo-labeled dataset $\mathcal{\hat{D}}$) into smaller overlapping trajectories of fixed length $T$, where $T$ is a hyper-parameter which determines the complexity of the subroutines.\footnote{A smaller $T$ leads to simpler subroutines. We show ablations over $T$ in supplementary.} We encode actions in each sub-trajectory into a discrete latent vector $z$. This $z$ is then used to predict actions corresponding to different frames in the video.
\begin{wrapfigure}[25]{R}{0.68\textwidth}
\vspace{-0.0cm}
\insertWL{1.35}{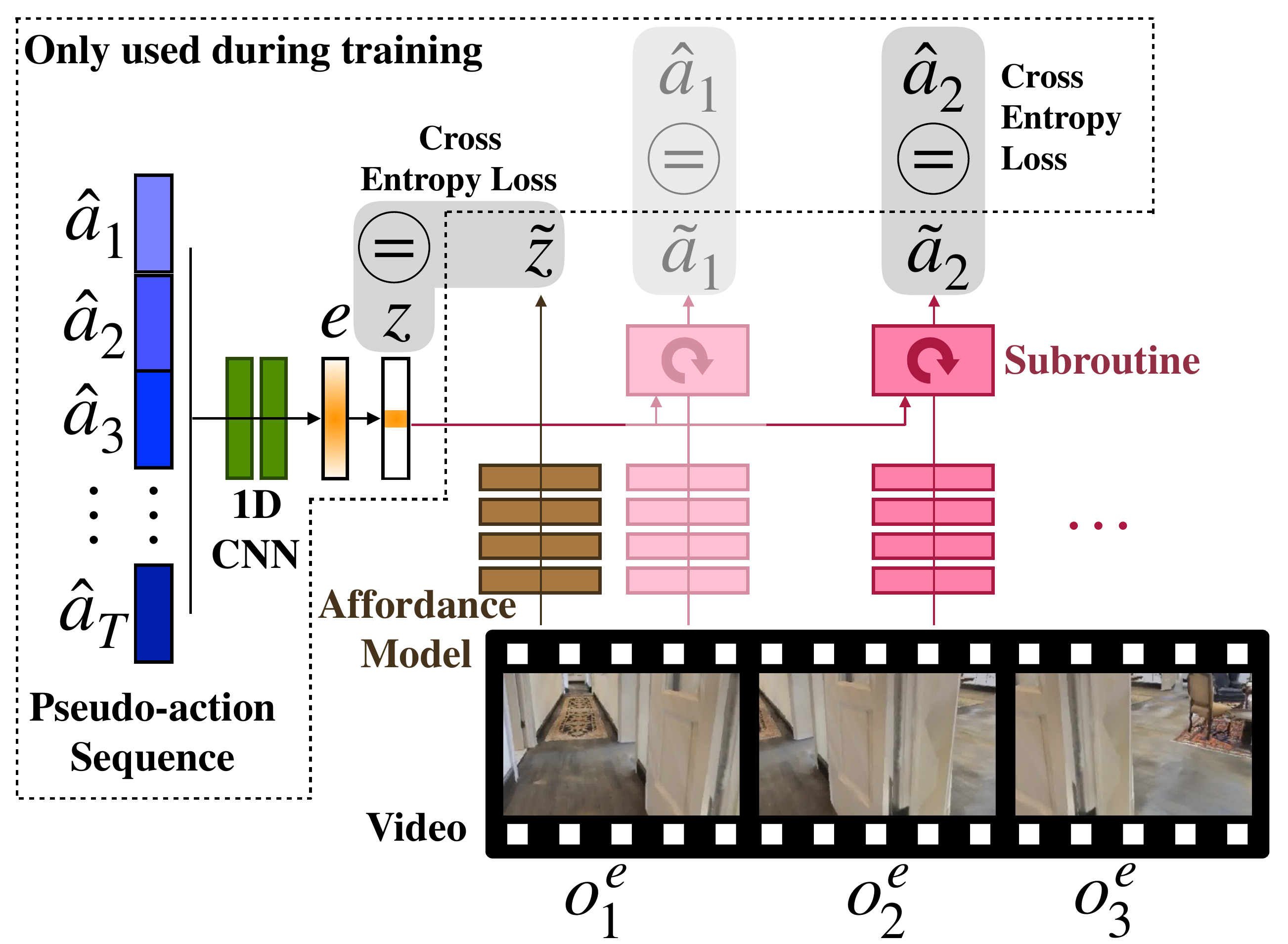}
\caption{\textbf{Learning Visuomotor Routines and Affordances:} We want to mine visuomotor routines with the ability to explicitly
invoke them. We implement this as recurrent network that takes in
the current observation and a one-hot vector $z$ that specifies the
subroutine to invoke. Since we don't have the labels
for subroutine id $z$, we obtain it by jointly training two networks:
one that looks at the entire future sequence of actions to predict
the subroutine id, and other which takes the subroutine id $z$ and
the image as input to predict the action to take. Both the networks
are jointly trained to minimize cross-entropy loss. Finally, we also train an affordance model that predicts
the inferred subroutine id $z$ from the first image.}
\figlabel{operator}
\end{wrapfigure}
We implement the trajectory encoder as network $f$ and the subroutines as a network $\pi$ parametrized by the subroutine-id $z$:
\begin{eqnarray}
  e &=& f(\hat{a}_1, \hat{a}_2, \ldots, \hat{a}_T)  \\
  z &\sim& \text{softmax}(e) \\
  \tilde{a}_t, h_{t+1} &=& \pi(o_t, z, h_t) \quad \forall t \in \{1 \ldots T-1\}
\end{eqnarray}

where state $h_t$ and $h_{t+1}$ are the current and updated hidden states respectively, $o_t$ is the current observation, and $z$ is a 
discrete latent vector which specifies the subroutine to invoke. 
We train the affordance model \am to predict the subroutine id given the \textit{first}
image of the video sequence:
\begin{eqnarray}
    \tilde{z} = \text{\am}(o_1).
\end{eqnarray}

At test time, \am takes the current observation as input and predicts 
which subroutines are applicable, and $\pi$ then
executes the selected subroutine in a closed loop, 
executing the predicted action $\tilde{a}_t$ and receiving the next 
observation $o_{t+1}$ as it proceeds. 

\textbf{Training}: Both the networks, $\pi$ and $f$, are jointly optimized to maximize the likelihood of the pseudo-labeled action sequence. \am is optimized to maximize the likelihood of $z$ given first observation of the video sequence. The subroutine id $z$ is sampled from the trajectory encoding $e$ through a Gumbel-Softmax distribution \citep{jang2016categorical}.
This allows estimating gradients for parameters of $f$ despite the sampling.
\figref{operator} shows the network diagram.

%% file: subroutine_expts.tex
\section{Experiments (Learning Subroutines and Affordances)}
\seclabel{exptsubr}
\input{experiment_setup}

\input{experiment_subroutines}

%% file: experiment_setup.tex
\subsection{Experimental Setup}
Our experiments involve use of \textit{environments} (where the \textit{agent}
can actively interact with the environment) \Einv and
\Etest, and a \textit{dataset} of first-person videos \VD. We describe choices
for the environment, agent and this video dataset: 

\textbf{Environments}: We model environments using a visually
realistic simulator derived from scans of real world indoor environments from
the Stanford Building Parser Dataset \citep{armeni20163d} (SBPD) and the
Matterport 3D Dataset \citep{Matterport3D} (MP3D). These scans have been used to
study navigation tasks in \citep{gupta2017cognitive, kumar2018visual, swedish2018deep}, and we adapt publicly available simulation code from
\citep{gupta2017cognitive}.  We split these environments into four disjoint sets: \Einv,
\Evideo, \Eval and \Etest. \Einv is used to train the inverse model, \Eval is used
for development of policies for down-stream tasks, and \Etest is used for
evaluating performance of our policies on down-stream tasks. \Evideo is used to 
create a dataset of egocentric videos. 

\textbf{Agent Model: } Our agent is modeled as a cylinder with 4
actions: a) stay in place, b,c) rotate left or right by $\theta$ ($=30^\circ$),
and d) move forward by $x$ ($=40\text{cm}$). The robot is equipped with a \rgb
camera mounted at a height $h$ ($=120\text{cm}$) from the ground with an elevation $\phi$ ($=-5^\circ$) from the horizontal. 

\textbf{Dataset \VD}: We create \textit{MP3D Walks Dataset} of 
egocentric videos. \textit{MP3D Walks Dataset} is auto-generated using the \Evideo
environments, by rendering out images along the path taken by an expert navigator to navigate between given pairs of random points. We implement this expert as an analytical path planner which has access to the ground truth free space map.  
We additionally ensure that experts have a different action space than 
our agent (see supplementary for specifics). 
\textit{MP3D Walks Dataset}
consists of around $217K$ clips of $40$ steps each, without any action labels.

%% file: experiment_subroutines.tex
\subsection{Training Details} 
\textbf{Inverse Model Training and Pseudo-labeling.} The agent starts at $1.5K$ different
locations spread over 4 environments ($\mathcal{E}_{train}$) and executes random actions for $30$ steps. The collected data ($45K$
interaction samples) is used to train the inverse model. See supplementary for ablations over the number of interaction samples and the generalization performance of the inverse model over different camera heights of the test images. 

This model is then used to pseudo-label videos in \VD to obtain
dataset $\mathcal{\hat{D}}$ as described in \secref{video}. 
$\mathcal{\hat{D}}$ is used to learn subroutines $\pi(., z)$
and the affordance model. (\secref{conditional}).

\input{realrobot}
\textbf{Subroutine Training}: We slice each of the $217K$ videos into clips
of length $10$ steps with a sliding window of $5$. This gives us a total of
$2.2M$ clips to train our subroutines. 
We experiment with using $4$ subroutines (\ie the $z$ vector is
$4$-dimensional. We show ablations over the number of subroutines as well as the length of each subroutine in supplementary.). 
This model is trained by
minimizing the cross-entropy loss between the actions output by the policy
($\tilde{a}$) and the pseudo-labels ($\hat{a}$) obtained from the inverse
model. 
\input{ops_aff_fig}

\textbf{Affordance Training}: We train the affordance model to predict the
 inferred subroutine id $z$ given the first image of the length $10$ trajectory
 by minimizing cross-entropy loss over the inferred $z$ label.

\subsection{Results}
\textbf{Behavior of Subroutines}: We deployed our subroutines (learned in 
simulation) in the real world on a real robot (\texttt{iCreate2}
platform equipped with a \rgb camera). 
\figref{realrobot1} shows the diversity between two of our learned subroutines \subR{1} and \subR{2}. We observe that \subR{1} prefers turning rightward into doors and corridors, and \subR{2} prefers turning leftward. The subroutines show robustness to perturbations in the starting location and consistently enter the door. \figref{realrobot2} shows that \subR{2} consistently turns left into doors and corridors across different starting locations. See supplementary for simulation results showing diversity and consistency of our learned subroutines.
\textbf{Affordance Prediction}: We show observations from \Etest for which 
the affordance model prediction is high for \subR{1} and \subR{2} 
in \figref{affordanceboth}. Top row shows observations that cause a high
prediction for \subR{1}, while bottom row shows images that excite \subR{2}.

%% file: realrobot.tex
\begin{figure}
\centering
\begin{subfigure}[b]{0.6\textwidth}
\insertWL{1.3665}{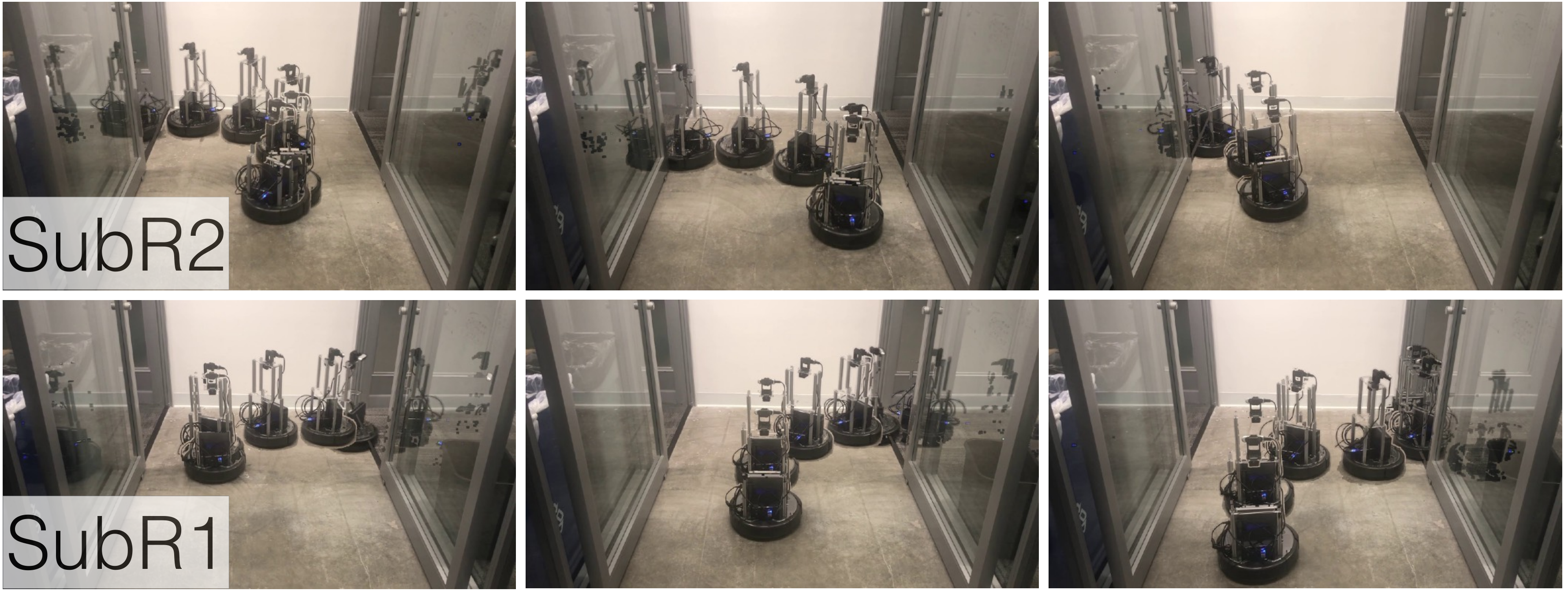} 
\caption{\textbf{Robustness and Diversity of Subroutines}}
\figlabel{realrobot1}
\end{subfigure}
\begin{subfigure}[b]{0.39\textwidth}
\centering
\insertWL{0.92}{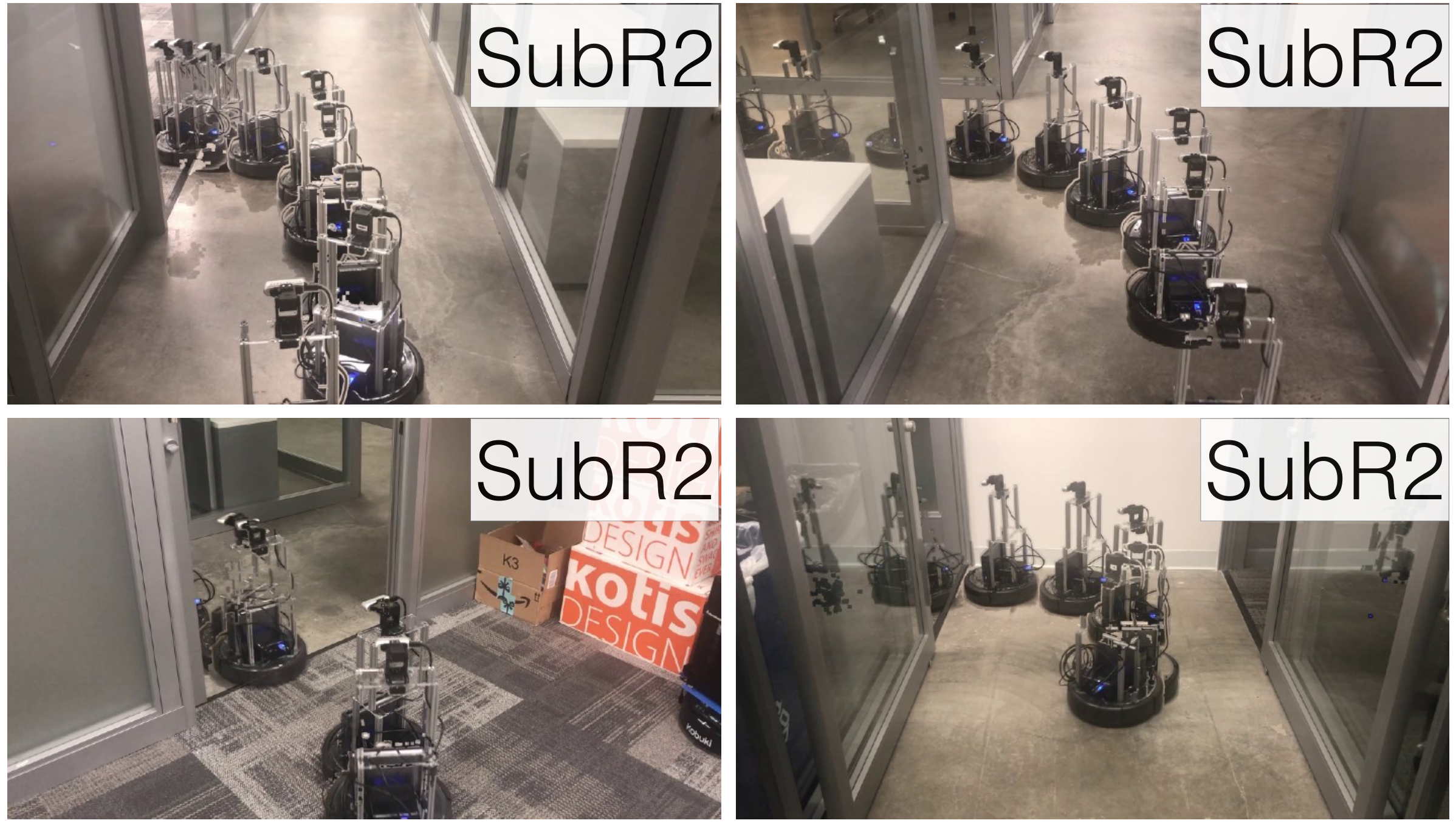} 
\caption{\textbf{Consistency of Subroutines}}
\figlabel{realrobot2}
\end{subfigure}
\caption{\textbf{(a) Robustness and Diversity of Subroutines}: Subroutines \subR{2} (top row) and \subR{1} (bottom row) when deployed on a real robot demonstrate robustness and consistency over perturbations to starting location. \textbf{(b) Consistency of Subroutines}: Learned Subroutines when deployed on a real robot demonstrate consistent behavior over different starting locations, as demonstrated for \subR{2}.}
\end{figure}

%% file: ops_aff_fig.tex
\begin{figure}
\centering
\includegraphics[scale=0.194]{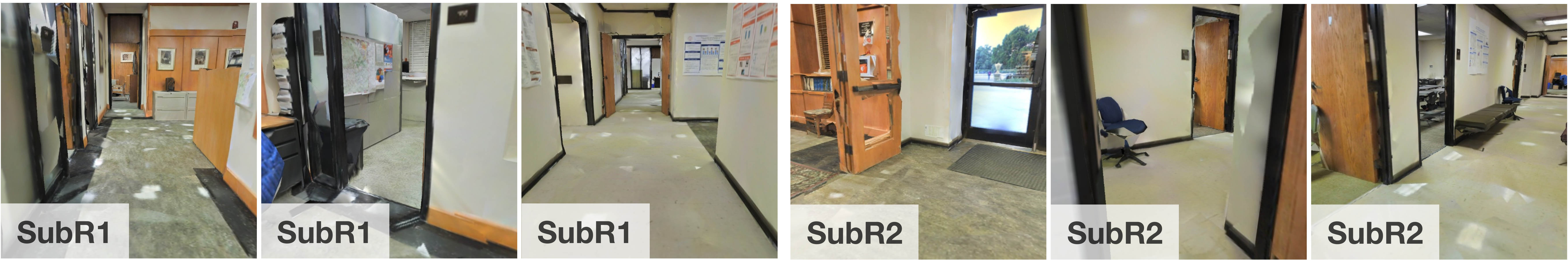}
\caption{\textbf{Affordance Model Visualization:} The images in the first 3 columns were assigned high probability for \subR{1} by the affordance model 
(that goes rightwards), while images in the next three columns were
assigned high probability for \subR{2} (that goes leftward). 
These images, indeed afford the predicted subroutines. 
}
\figlabel{affordanceboth}
\end{figure}

%% file: experiments.tex
\section{Experiments (Using Affordances and Subroutines)}
\seclabel{experiments}
\input{exploration}

\input{rl_plots}

\input{hrl}

%% file: exploration.tex
\subsection{Exploration via \methodname}
The exploration task requires the agent to explore a novel environment
efficiently.
\input{cov_table}
\textbf{Task Setup}: We randomly initialize the agent at $100$ different
locations in the novel test environment \Etest. For each location, we do $5$
random executions (each from a randomly chosen initial orientation) of
length $408$ each. 

\textbf{Metrics}: We measure different aspects of exploration via
the following metrics: 
\textbf{\# Samples $\downarrow$}: Number of environment interactions used for training. 
\textbf{Average Distance to Trajectory (ADT) $\downarrow$}: Given executed trajectories
from a given starting location, we compute the mean geodesic distance
of points in the environment to the closest point on the trajectory. If we
wanted to visit a point in the environment, this metric measures how much
we will need to go off the trajectory to get to this point, in expectation. We report the average over all starting locations. 
\textbf{Maximum Distance $\uparrow$}: Measures how far the executed trajectories convey the agent. For each trajectory, we measure the maximum geodesic distance from the starting location to all points on the trajectory. We report the average maximum geodesic distance over trials. 
\textbf{Collision Rate $\downarrow$}: Fraction of forward actions that result in collisions. We emphasize that \methodname is not trained to optimize for
any of these metrics.

\textbf{Exploration via Subroutines}: Given the visual observation from the
current location, we repeat the following two steps: a) we use the affordance
model \am to sample the subroutine $z$ to execute, b) we execute the sampled
subroutine for 10 steps. 
\textbf{Baselines.} We compare with three hand-crafted baselines: a) Random
policy (randomly execute one of the 4 actions), b) Forward bias policy (biased
to more frequently execute forward action), and c) Always forward but rotate on
collision policy. We also compare to the state-of-the-art unsupervised RL-based
skill learning methods d) DIAYN \citep{eysenbach2018diversity}, and e) Curiosity
\citep{pathak2017curiosity}. These learning based techniques were trained with
comparable networks (ResNet~18 models pre-trained on ImageNet) for over $10M$
samples. More details about these baselines are in supplementary.

\textbf{Results.}
\tableref{explore-results} shows that \methodname outperforms all the baselines
on all three metrics. 
\methodname successfully learned to bias
towards forward action in navigation and the notion of obstacle avoidance,
leading to a high maximum distance, and a low collision rate. It outperforms
hand-crafted baselines that were designed with these insights in mind.
Furthermore, it outperforms state-of-the-art learning based techniques for
learning skills \citep{eysenbach2018diversity, pathak2017curiosity}, that were
trained on $200\times$ more interaction samples (45$K$ \vs
10M). These past works have only been shown to perform well for
low-dimensional state-spaces and simple game environments, and in our experiments they either learn a trivial solution, or suffer with high-dimensional inputs such as
real world images (further discussed in supplementary).

%% file: cov_table.tex
\begin{table}
\centering
\caption{\textbf{Exploration Metrics:} \methodname beats 3 hand-crafted
baselines and two state-of-the-art learning based
techniques~\citep{eysenbach2018diversity, pathak2017curiosity}. See text for
details.}
\begin{tabular}{lccccc}
\toprule
\textbf{Method}& \textbf{\# Samples} & \textbf{ADT} & \textbf{Maximum} & \textbf{Collision} 
\\
& $\downarrow$ & $\downarrow$ & \textbf{Distance $\uparrow$} & \textbf{Rate (\%) $\downarrow$}
\\ \midrule
Random                                                 & 0       &  18.09 & 7.5   & 65 
\\ Forward Bias Policy                                 & 0    &   15.25 & 13.11 & 82
\\ Always Forward, Rotate on Collision          & 0       &  14.89 & 13.31 & 72
\\ Skills from Diversity \citep{eysenbach2018diversity} & 10$M$ &  17.63 & 7.85  & 67
\\ Skills from Curiosity \citep{pathak2017curiosity}    & 10$M$ &  17.68 & 7.87  & 64
\\ \methodname (Ours)                   & 45$K$   & \textbf{7.73}  & \textbf{27.78} & \textbf{12}
\\ \bottomrule
\end{tabular}
\tablelabel{explore-results}
\end{table}

%% file: rl_plots.tex
\begin{figure}
\begin{subfigure}[b]{0.66\textwidth}
\centering
\insertWL{0.78}{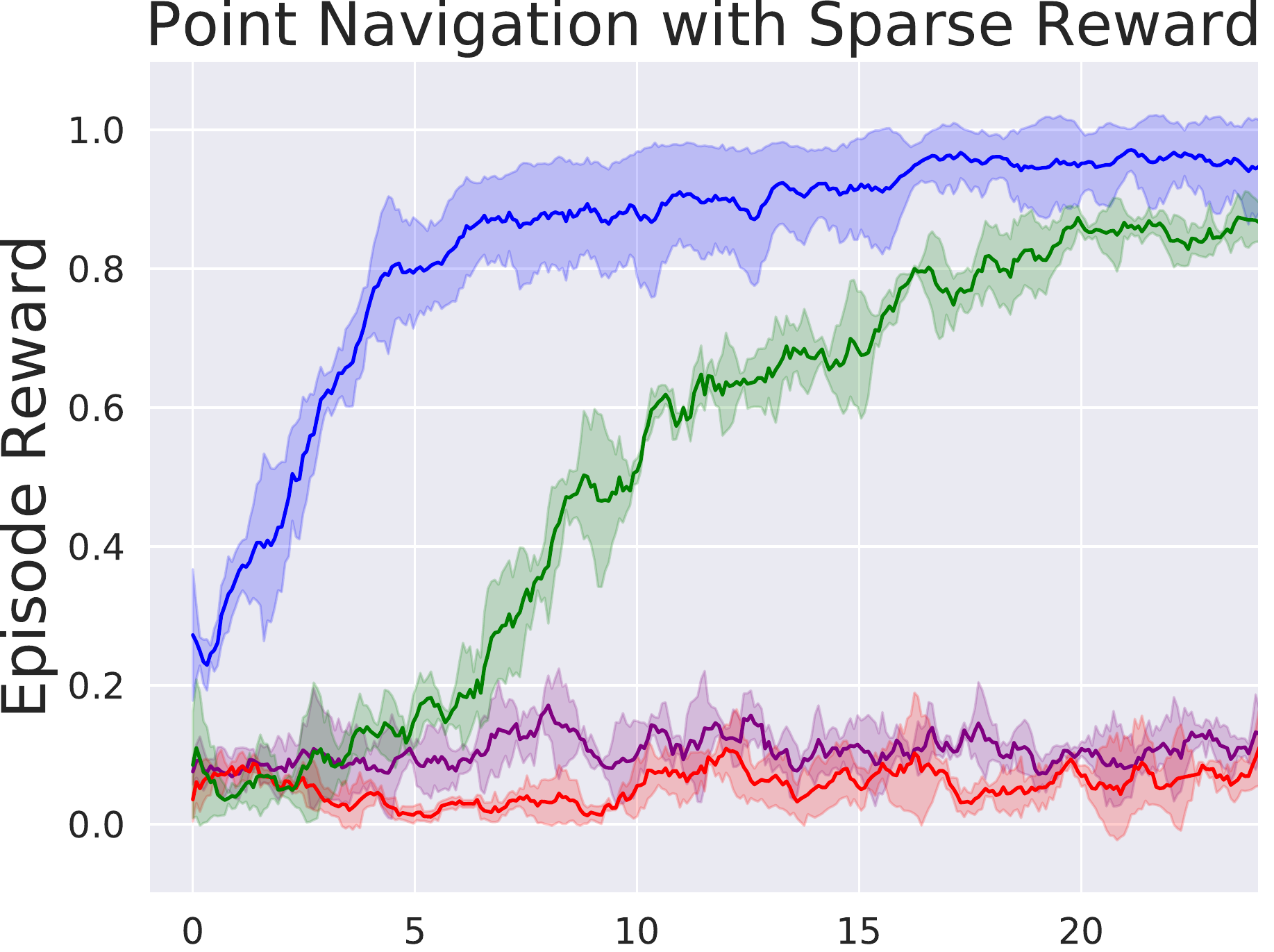} 
\insertWL{0.7}{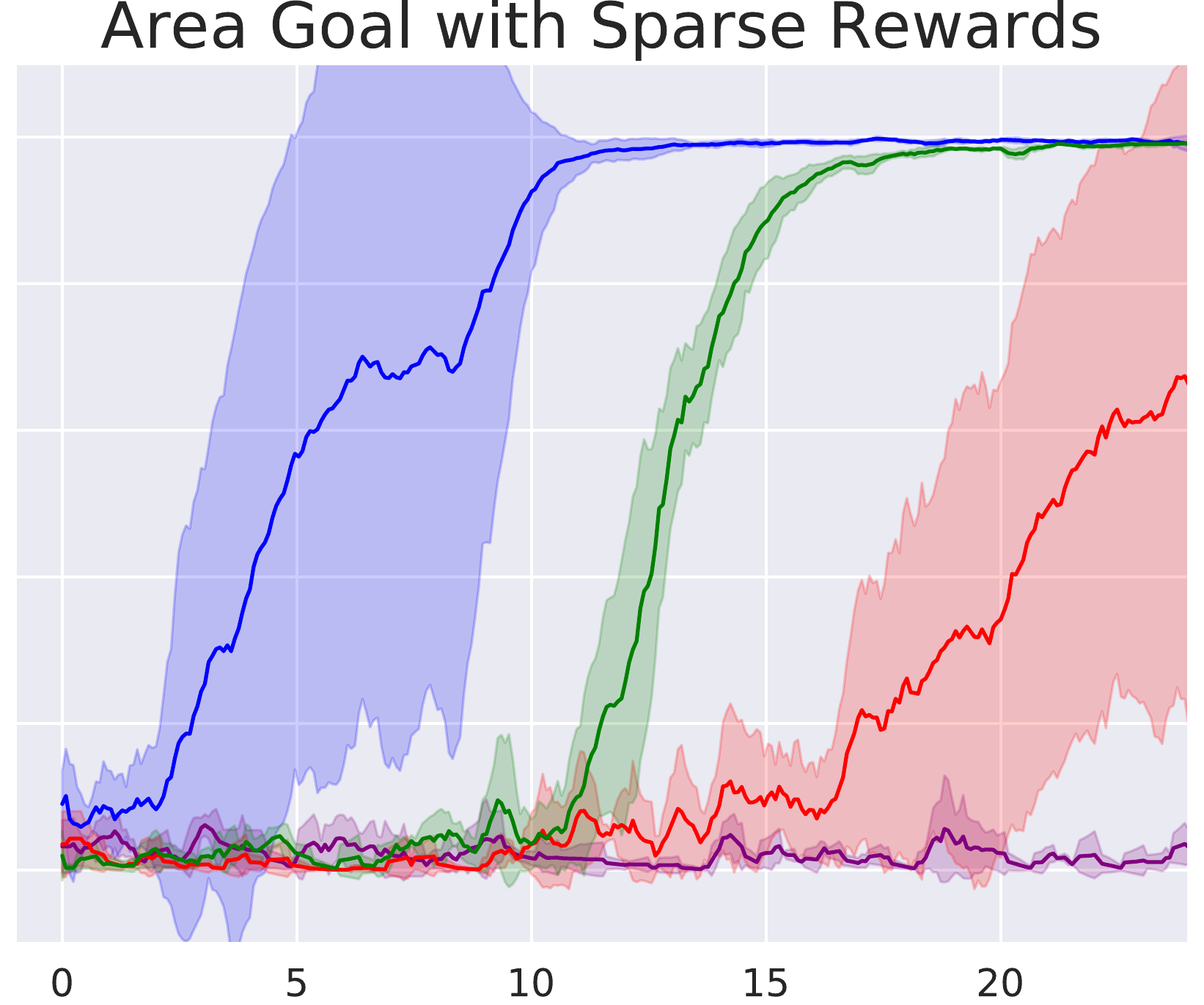} \\
\insertWL{0.78}{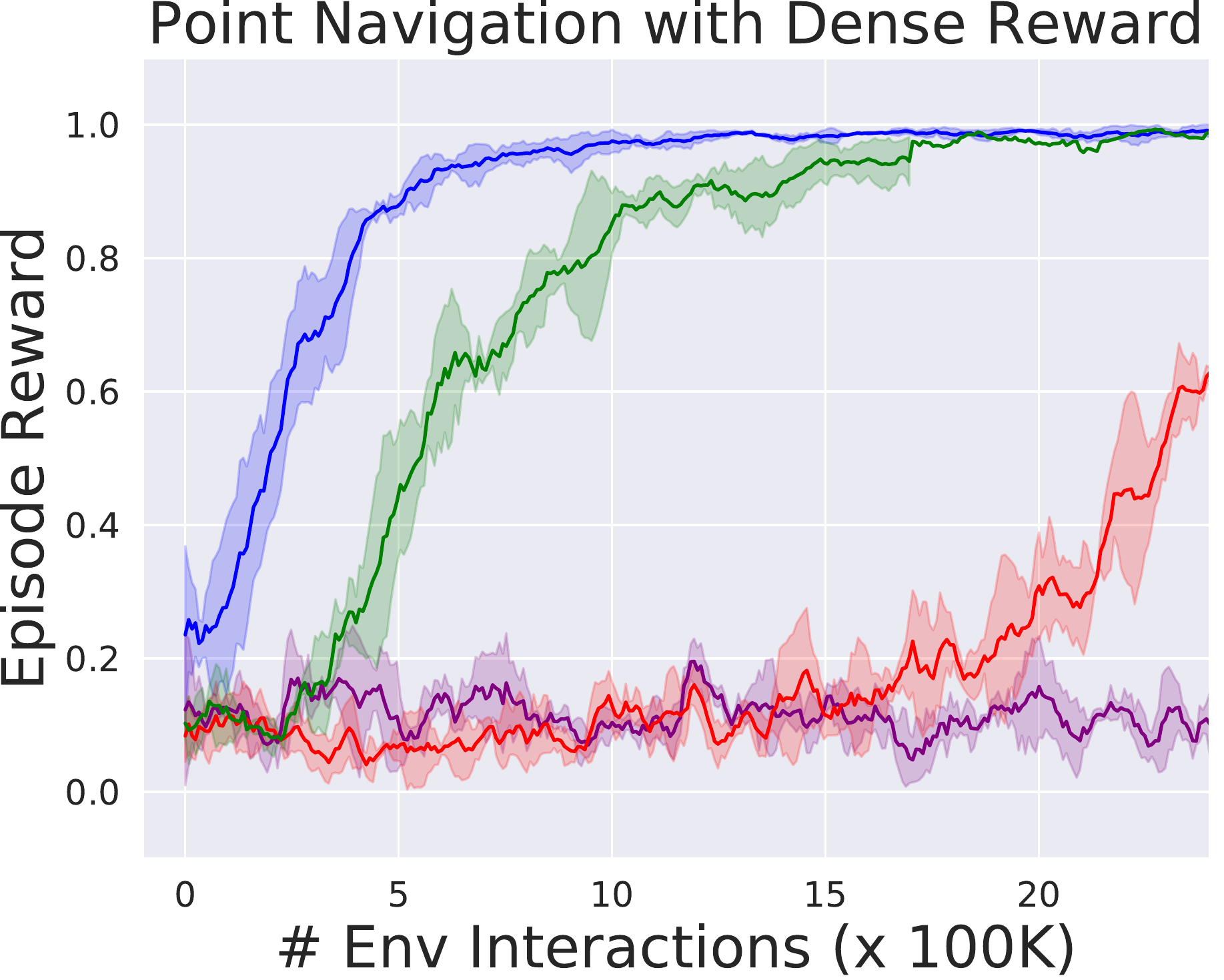}
\insertWL{0.7}{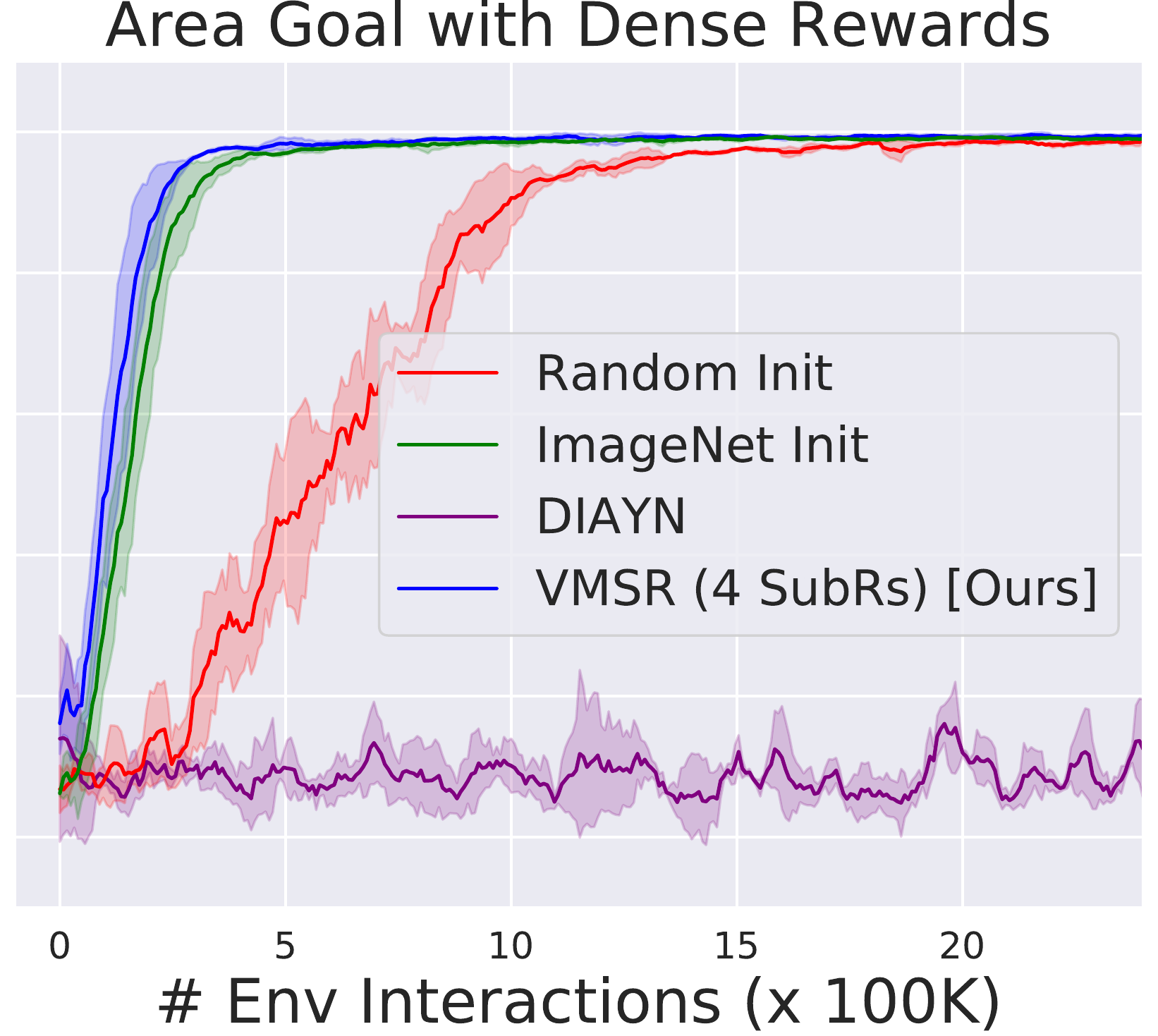}
\caption{\textbf{Subroutines and Affordances for Hierarchical RL}}
\figlabel{rltask-hrl}
\end{subfigure}
\begin{subfigure}[b]{0.33\textwidth}
\centering

\insertWL{0.705}{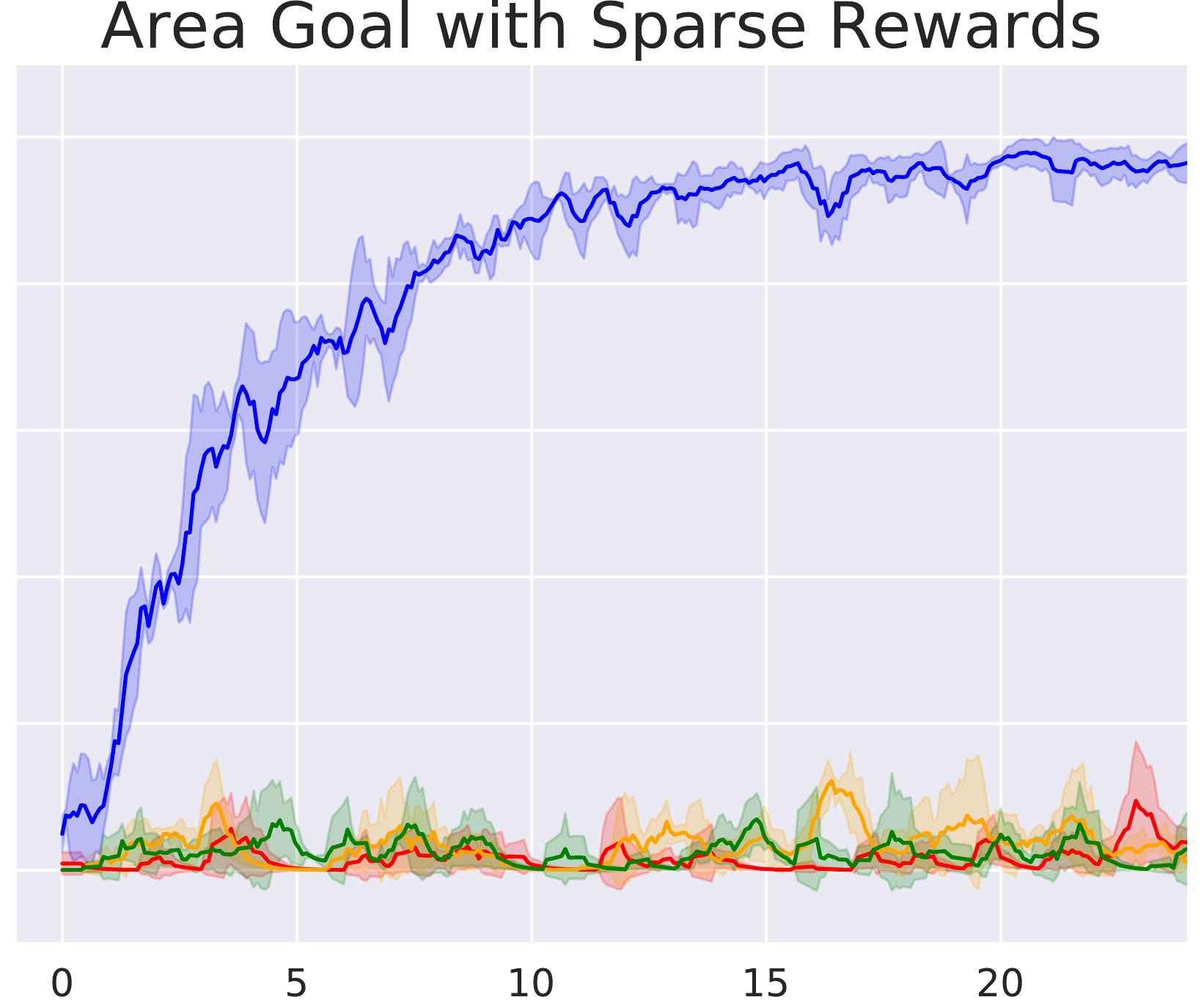}
\insertWL{0.7}{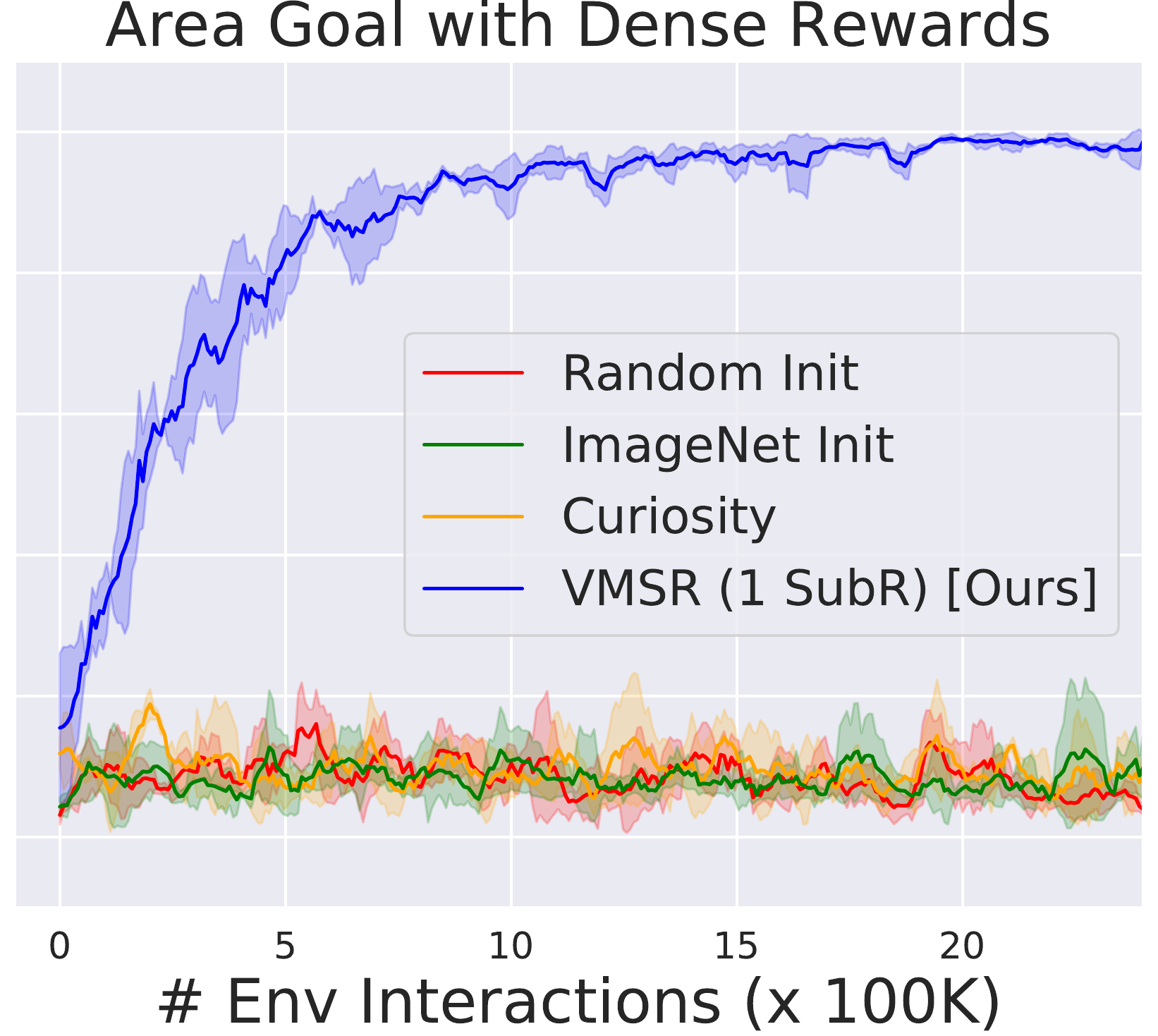}

\caption{\textbf{Subroutines for RL}}
\figlabel{rltask-rl}
\end{subfigure}
\caption{\textbf{(a) Subroutines and Affordances for Hierarchical RL}:
Initializing from \methodname leads to upto 4x more sample efficient learning in downstream navigation tasks. First Column shows results for
PointGoal (go to $(x,y)$ coordinate), second column shows results for AreaGoal (go
to washroom). We see improvements across these tasks for both sparse and dense
reward scenarios, with larger gains in the harder case of sparser rewards. \textbf{(b) Subroutines for RL}:
Initializing a flat RL policy with \methodname (with only 1 \subRgen) leads to improved
sample complexity for AreaGoal navigation (go to washroom),
compared to alternate initializations.} 
\end{figure}

%% file: hrl.tex
\subsection{PointGoal and AreaGoal via \methodname Initialization in HRL}
We next investigate how we can use \methodname to solve goal-driven tasks. 
We do this by setting up hierarchical RL
policies based of our learned subroutines and affordance models.

\textbf{Task Setup:} We setup two goal driven navigation tasks, PointGoal and
AreaGoal as defined in \citep{anderson2018evaluation}. For PointGoal task, the
agent is required to reach a given goal location (specified as a relative
offset from robot's current location).  For
the AreaGoal task, the agent is required to go to the washroom. 
We study both tasks in sparse and dense reward settings. RL and HRL policies
are developed on the validation environment \Eval, and finally trained 
on the test environment \Etest with 3 random seeds to assess sample
efficiency for learning. 

\textbf{HRL via Subroutines and Affordance Model:} We follow the framework in \citep{barto2003recent} and initialize the sub policies with our subroutines and meta-controller with our affordance model. We then fine tune the sub policies and the meta-controller via reinforcement learning.
 
\textbf{Comparisons:} 
We compare with the following alternates for
initializing the meta-controller and the sub-policies: a) \textit{Random
Initialization},  b) \textit{ImageNet Initialization}, and c)
\textit{Initialization from skills via DIAYN \citep{eysenbach2018diversity}
pre-training}. (c) doesn't provide an affordance model, so we initialize the
meta-controller image CNN with the sub-policy CNN.
We can also compare \methodname initialization to initialization obtained from
Curiosity \citep{pathak2017curiosity}. Pathak \etal \citep{pathak2017curiosity}
use a monolithic policy (\ie without any handle to control what they do), and
study an AreaGoal task. Thus, for a fair comparison, we limit the comparison to
the AreaGoal task and use a monolithic RL policy instead of the hierarchical
policy. We report three training curves: a) \textit{Random Initialization}, b)
\textit{Initialization from Curiosity \citep{pathak2017curiosity}}, and c)
\textit{\methodname} with 1 subroutine (to obtain a monolithic policy).

\textbf{Results:} Training rewards are plotted in \figref{rltask-hrl} and 
\figref{rltask-rl}. \figref{rltask-hrl} shows the comparison among hierarchical policies. 
We observe upto $4\times$ faster training when initialized with \methodname and affordance models as compared to the next best baseline (which is ImageNet 
initialization). Improvements are generally larger
for the harder case of sparser rewards. DIAYN \citep{eysenbach2018diversity}
based initialization entirely fails, as it collapses to a trivial policy (more
details in supplementary). Even among non hierarchical policies (\figref{rltask-rl}), initializing with \methodname (\methodname (1 \subRgen)) performs the best in AreaGoal Tasks,
outperforming random initialization and initialization from curiosity policy
\citep{pathak2017curiosity}, which also learns a trivial policy (see supplementary).

%% file: conclusion.tex
\section{Discussion}
In this paper, we developed a technique that combined learning from interaction with learning from videos, to extract meaningful and useful subroutines from egocentric videos of experts performing different tasks. We showed how these extracted subroutines can be used as is for exploration, or can be specialized using hierarchical RL for solving other downstream navigation tasks. We believe advances made in this paper will enable scaling up of policy learning in robotics.

%% file: supp.tex
\renewcommand{\thefigure}{A\arabic{figure}}
\renewcommand{\thetable}{A\arabic{table}}
\setcounter{figure}{0}
\setcounter{table}{0}
\appendix
\renewcommand{\thesection}{A\arabic{section}}
\setcounter{section}{0}

\vspace{1cm}
\section{Subroutines and Affordances Full Training Details}
\textbf{Inverse Model Training and Pseudo-labeling.} The agent starts at $1.5K$ different
locations spread over 4 environments ($\mathcal{E}_{train}$) and executes random actions for $30$ steps. The collected data ($45K$
interaction samples) is used to train the inverse model. We use cross-entropy
loss between the actual action and the predicted action. We use Adam
\citep{kingma2014adam} with $64$ batch size and $0.001$ learning rate. Ablations over number of interaction samples by varying number of starting locations and number of steps per starting location is shown in \figref{ablations}.  

This model is then used to pseudo-label videos in \VD to obtain
dataset $\mathcal{\hat{D}}$. $\mathcal{\hat{D}}$ is used to learn subroutines $\pi(., z)$
and the affordance model. 

\textbf{Subroutine Training}: We slice each of the $217K$ videos into clips
of length $10$ steps with a sliding window of $5$ (ablations over the length of subroutines is shown in \figref{ablations}). This gives us a total of
$2.2M$ clips to train our subroutines. 
We experiment with using $4$ subroutines (\ie the $z$ vector is
$4$-dimensional) and show ablations over this hyper-parameter \figref{ablations}. 
This model is trained by
minimizing the cross-entropy loss between the actions output by the policy
($\tilde{a}$) and the pseudo-labels ($\hat{a}$) obtained from the inverse
model. 

\textbf{Affordance Training}: We train the affordance model to predict the
 inferred subroutine id $z$ given the first image in length $10$ trajectory
 by minimizing cross-entropy loss over the inferred $z$ label.

\section{Consistency and Diversity Visualizations}

We unroll different subroutines from different locations in the test environment \Etest,
and visualize the trajectories followed by each of them in
the top view in \figref{consistency}. We show multiple rollouts of each subroutine from each of the starting locations. Randomness in behavior comes from the sampling of the actions from the network output. The three top view figures in each column
of \figref{consistency} correspond to one subroutine at three different starting locations and we rollout 8 trajectories from each starting location. 
Thus, each column demonstrates that a specific subroutine does
similar things when initialized at different locations, showing the consistency of our learned subroutines. For example, SubR1 always turns right, SubR2 always turns left. Rollouts
shown in different rows of \figref{consistency} show that different
subroutines show diverse behaviors when started from the
same location. This shows the diversity of across our learned subroutines.

We also quantitatively compute disentanglement: we unroll the different subroutines from the same starting location and compute the intersection over union between trajectories from \subR{$_i$} and \subR{$_j$} (for example, IoU between the green region and blue region in plots in the top row of Figure A2). A higher IoU implies similar areas are traversed by two sets of sub-policies, lower IoU implies the two sub-policies are distinct. Thus, we should expect a higher IoU between the trajectories from the same subroutine and a lower IoU between different subroutines. The average IoU between different subroutines is 0.42, and the average IoU between trajectories from the same subroutines is 0.58. Thus, indeed different subroutines are disentangled.

\section{Affordance Model Entropy Visualization}
\input{entropy_viz}
We also look at the entropy of the output of affordance model in \figref{entropy} in top view. The arrow shows the direction in which the agent is facing and we plot the entropy of the prediction of the affordance model when the first-person (egocentric) observation is given as input. A higher entropy implies that more subroutines apply in the given scenario. The observed entropy is consistent with our expectations, as explained in the figure caption.

\section{Baselines for Exploration}
\seclabel{explr-baseline}
\begin{enumerate}[noitemsep]
\item 
\textit{Random Policy}: We randomly sample an action from the four
possible actions (stay, left, right, forward) at every step. 
\item 
\textit{Forward Bias Policy}: Since motion is typically dominated by forward
motion, we compare to another policy that samples the forward action more
preferably. We use the distribution of actions in the MP3D Walks Dataset,
probabilities for {\small \texttt{stop, turn left, turn right and forward}}
were $[0.0, 0.17, 0.17, 0.66]$ respectively.
\item 
\textit{Always Forward, Rotate on Collision}: This baseline repeats the
following procedure: rotate by a random angle sampled from $(-\pi, \pi]$,
move straight till collision.
\item 
\textit{Diversity Policy (DIAYN) \citep{eysenbach2018diversity}}: We use
the state-of-the-art RL-based unsupervised skill learning algorithm from
Eysenbach \etal \citep{eysenbach2018diversity} to learn 4 diverse skills on
\Einv environments. We test the learned skills for exploration by randomly
sampling a skill, and then executing it for $10$ steps, where we sample actions
from the probabilities output by the selected skill. Policy architecture is
same as those for our subroutines, discriminator is based of a ResNet~18 model.
Both models are initialized from ImageNet. Policy is trained for over 10
million interaction, best performance occurs at around 1M interaction samples.
\item 
\textit{Curiosity Policy \citep{pathak2017curiosity}}: We train a
curiosity-based agent that seeks regions of space where its forward model has
high prediction error \citep{pathak2017curiosity}. Policy architecture is same
as that for our subroutines (except that it does not take in the latent vector
$z$), and initialized from ImageNet. Forward model is learned in the $conv5$
average pooled feature space of a fixed Resnet~18 model pre-trained on
ImageNet. Trajectories are executed by sampling from the action probabilities
output by the policy. Once again, policy is trained for over 10 million
interaction, best performance occurs at around 1M interaction samples.
\end{enumerate}

\input{vis_operators}

\textbf{Curiosity Model:} Pathak \etal \citep{pathak2017curiosity} proposed use
of prediction error of a forward model as an intrinsic reward for learning
skills using RL.  We were surprised at the rather poor performance for the
curiosity model. We found that the model converges to the policy of simply
rotating in-place. Such a degenerate solution makes sense as rotating in-place
has higher prediction error than staying in-place and moving forward. In-place
rotations cause new parts of the environment to become visible which makes for
a harder prediction task. Staying-in-place and moving forward cause only minor
changes to the image or no changes at all. Thus, the curiosity model rightly
learns to simply rotate in-place. We saw this same behavior across different
runs with different hyper-parameters and different architectures: policies will
collapse to outputting just the rotation actions. Entropy based regularization
is used to prevent such a collapse. We used such regularization and
cross-validated various choices for the trade-offs in loss between entropy
regularization and policy gradient loss, but didn't find it to alleviate this
issue. We selected the best model for the task of exploration across different
runs and different number of training iterations. This selected model ended up
being a heavily regularized model that would pick actions almost uniformly at
random, as that would get higher performance than simply rotating in-place.
As both extremes (taking actions randomly, or picking only the rotate in-place
action) are trivial solutions, the curiosity model starts to ignore the image
and consequently performs on-par with uninitialized models for reinforcement
learning tasks. 

\textbf{Diversity Model:} The diversity model from Eysenbach \etal
\citep{eysenbach2018diversity} seeks to classify states with the skill id that
was used to get to it (see Algorithm 1 in \citep{eysenbach2018diversity}). While
this works well for the environments studied in \citep{eysenbach2018diversity},
it breaks down for visual navigation. This is because, the same state can be
reached via different skills depending on the starting state. This causes the
skill classifiers $q$ to only perform at chance.  Consequently, the reward for
the skill policies is uniform, causing the policies to collapse (all actions
produce the same reward, and hence no learning happens). We observed this
empirically in our experiments as well: accuracy for state classification was
at chance (25\% for four skills), and the reward stayed constant. Best
performing policy (based on validation for exploration metrics) always
predicted the following probabilities for different actions for different
skills: $[0.246, 0.232, 0.237, 0.285]$ (for {\small \texttt{stop, left, right,
forward}} respectively). As this can be done without looking at the image, the
policy learns to ignores the image. Thus, the model perform on-par with
uninitialized models for hierarchical reinforcement learning experiments.

\input{vis_coverage}

\input{ablation_plots}

\section{Exploration Visualization}
We show the coverage for each method in \figref{coverage}.  Figure 10 overlay trajectories executed by different policies onto the map (only used for visualization). We see a wider coverage for \methodname over other methods and also observe that the trajectories avoid the walls of the hallways when going down them. 
\section{Ablations}
We show ablations over 4 hyper-parameters in \figref{ablations} (see caption for more details). We compare \methodname initialization to inverse features initialization for downstream HRL task in \figref{invfeats}. We show the ability of the trained model to generalize across various camera heights of the reference images in \figref{inv-ht}.

\input{rl_setup}
\section{Video Results}
The enclosed video \texttt{vmsr.mp4} contains video results of real robot deployment followed by an explanation of our method. We use \citep{pyrobot2019} for real robot deployment. Note that along with the 4 primitive actions (rotate left, rotate right, move forward and stay in place), the robot also moves slightly backward incase of a collision.

\begin{figure}
\centering
\insertWL{1.0}{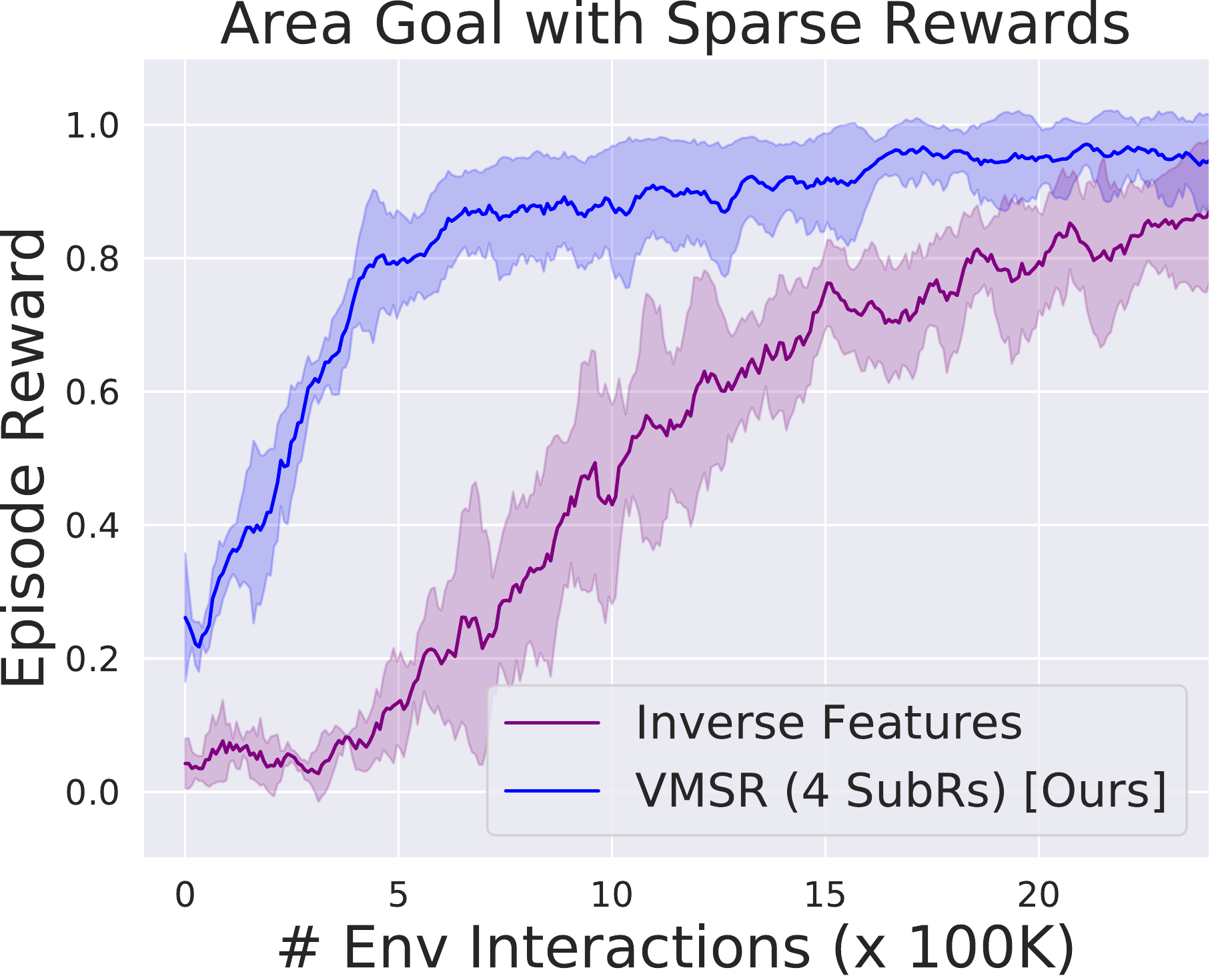} 
\caption{We compare \methodname initialization to initializing the image features of the sub-policies with the features from the inverse model for the downstream HRL task of PointGoal with sparse rewards. We see that \methodname is ~3x more sample efficient compared to this baseline.}
\figlabel{invfeats}
\end{figure}

\begin{figure}
\centering
\insertWL{1.0}{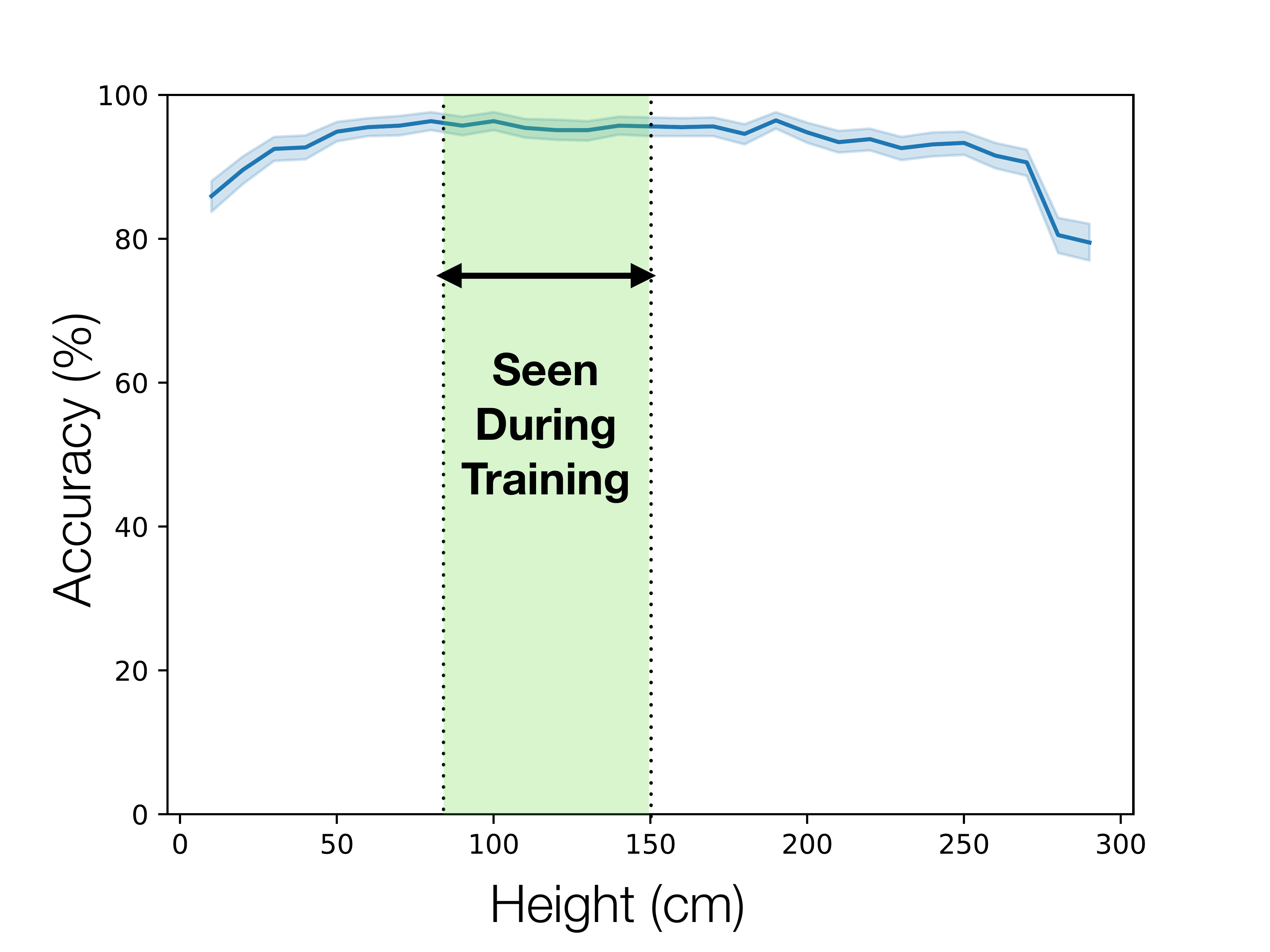} 
\caption{We test the generalization of the learned inverse model on images from \Eval, which is unseen during training. We plot the prediction accuracy (y axis) as we increase the camera height from the ground (x axis). The agent is trained on heights from 90cm to 150 cm during training in \Einv and evaluated on heights from 10cm to 300cm. We observe a very consistent performance even in the range not seen during training. Note that the agent starts touching the ceiling of the room in some places at 300cm.}
\figlabel{inv-ht}
\end{figure}

\section{Area Splits and Agent Settings}
We give details of area splits and the action space of the experts which generate reference videos in \tableref{agent-settings}. In the table, step size ($x$) refers to the length of a single forward step, $\theta$ refers to the rotation angle for left/right turn, $\phi$ refers to the elevation angle of the onboard RGB camera from the horizontal and $h$ refers to the height of the robot from the ground.  
\input{supp-settings}
\newpage

%% file: entropy_viz.tex
\begin{wrapfigure}[15]{R}{0.6\linewidth}
\vspace{-0.5cm}
\centering
\insertWL{1.4}{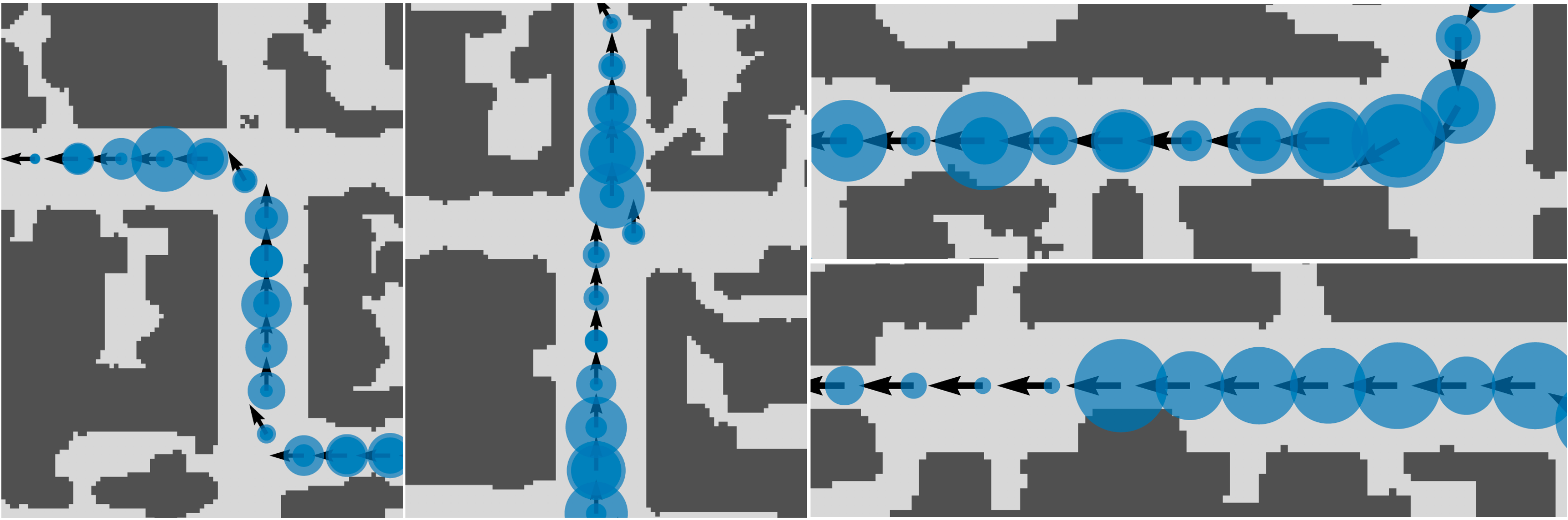}
\caption{\textbf{Multi-modality in Affordance Predictions}: We visualize the
entropy of the distribution output by the affordance model
in the \textit{test} environment. A larger circle
denotes a higher entropy meaning more subroutines can be invoked at that
location. We observe that the affordance model has a higher entropy as the agent approaches hallway intersections, or room entrances.
This multi-modality collapses as the agent crosses the decision junctions.}
\figlabel{entropy}
\end{wrapfigure}

%% file: vis_operators.tex
\begin{figure}
\centering
\insertWL{0.56}{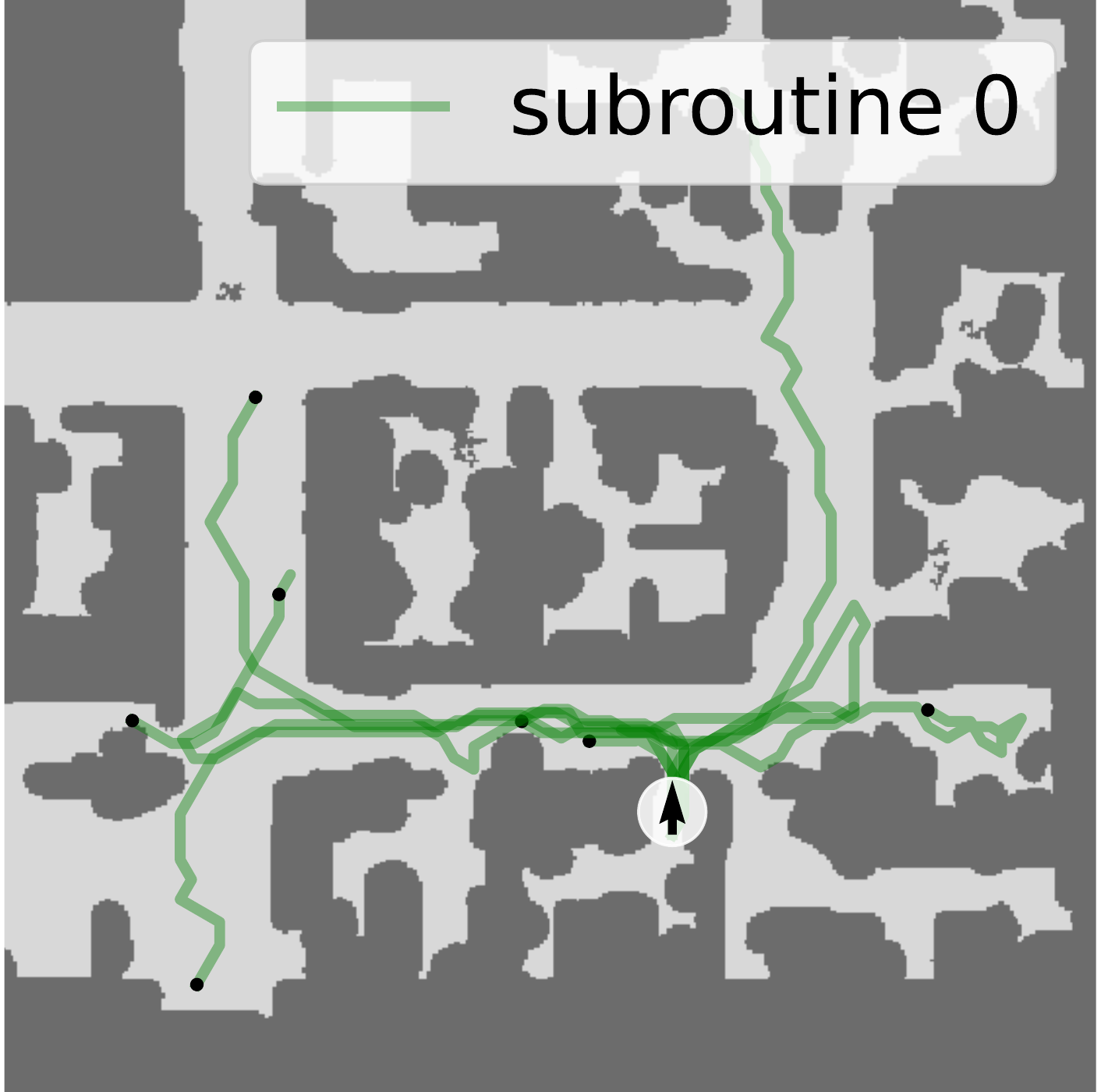}
\insertWL{0.56}{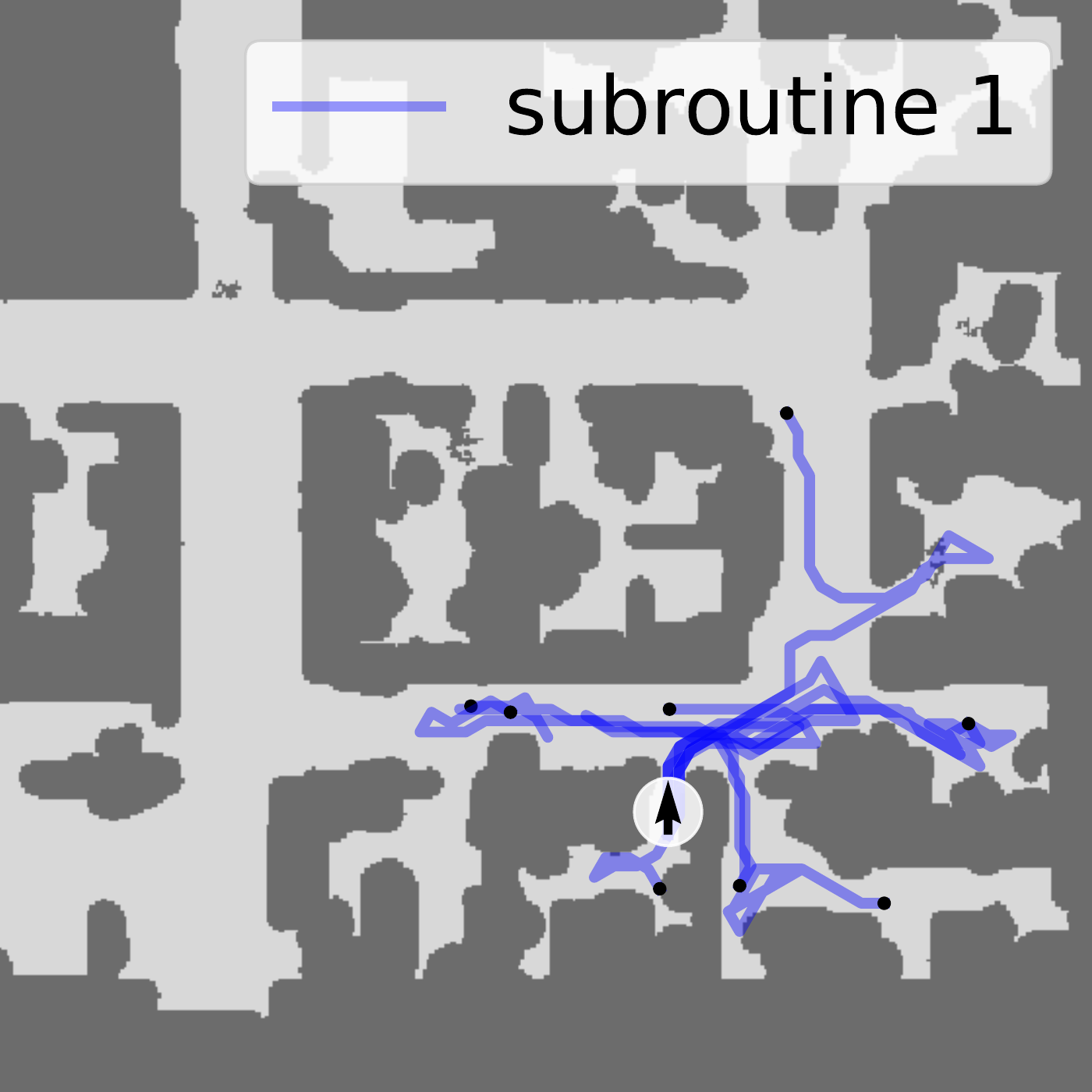}
\insertWL{0.56}{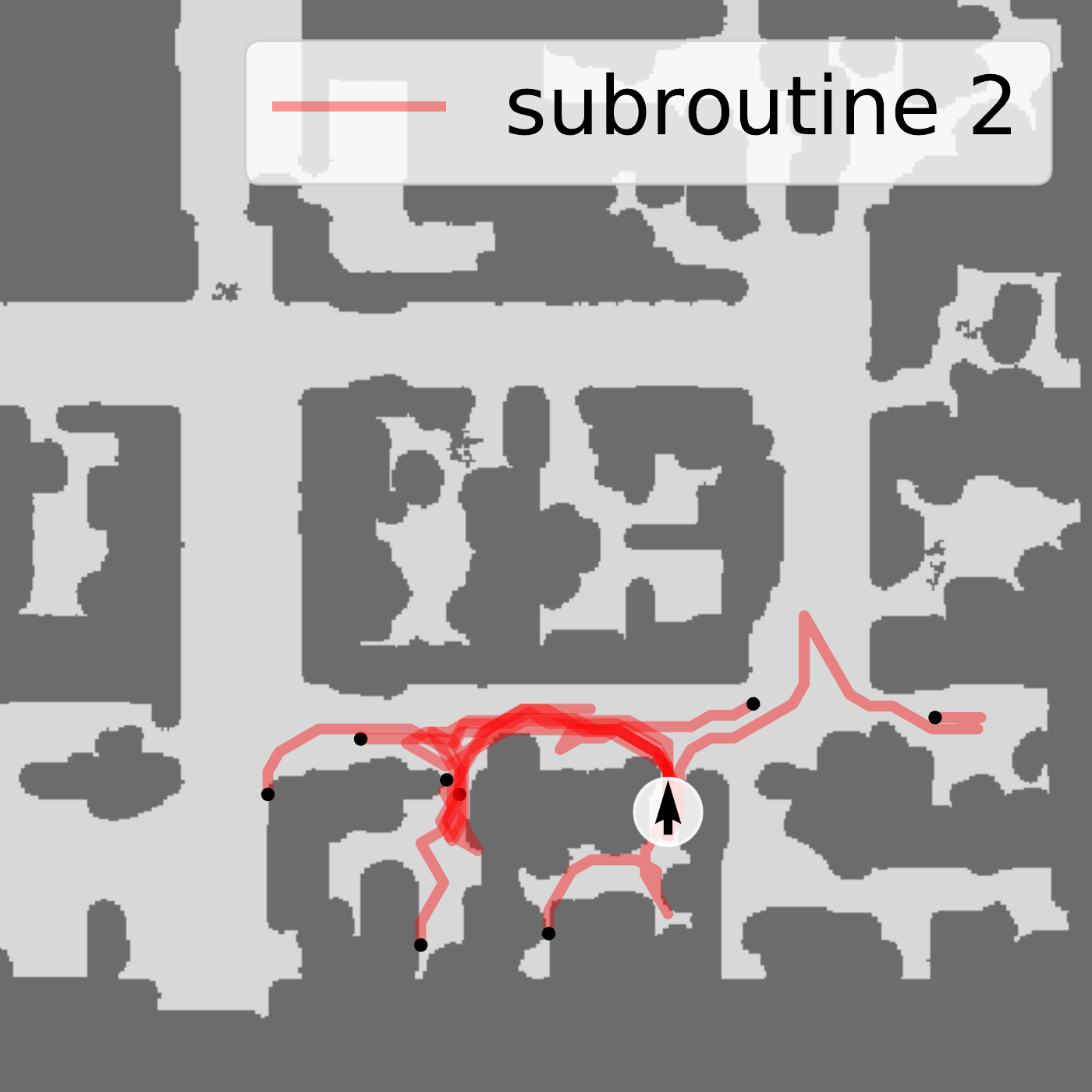}
\insertWL{0.56}{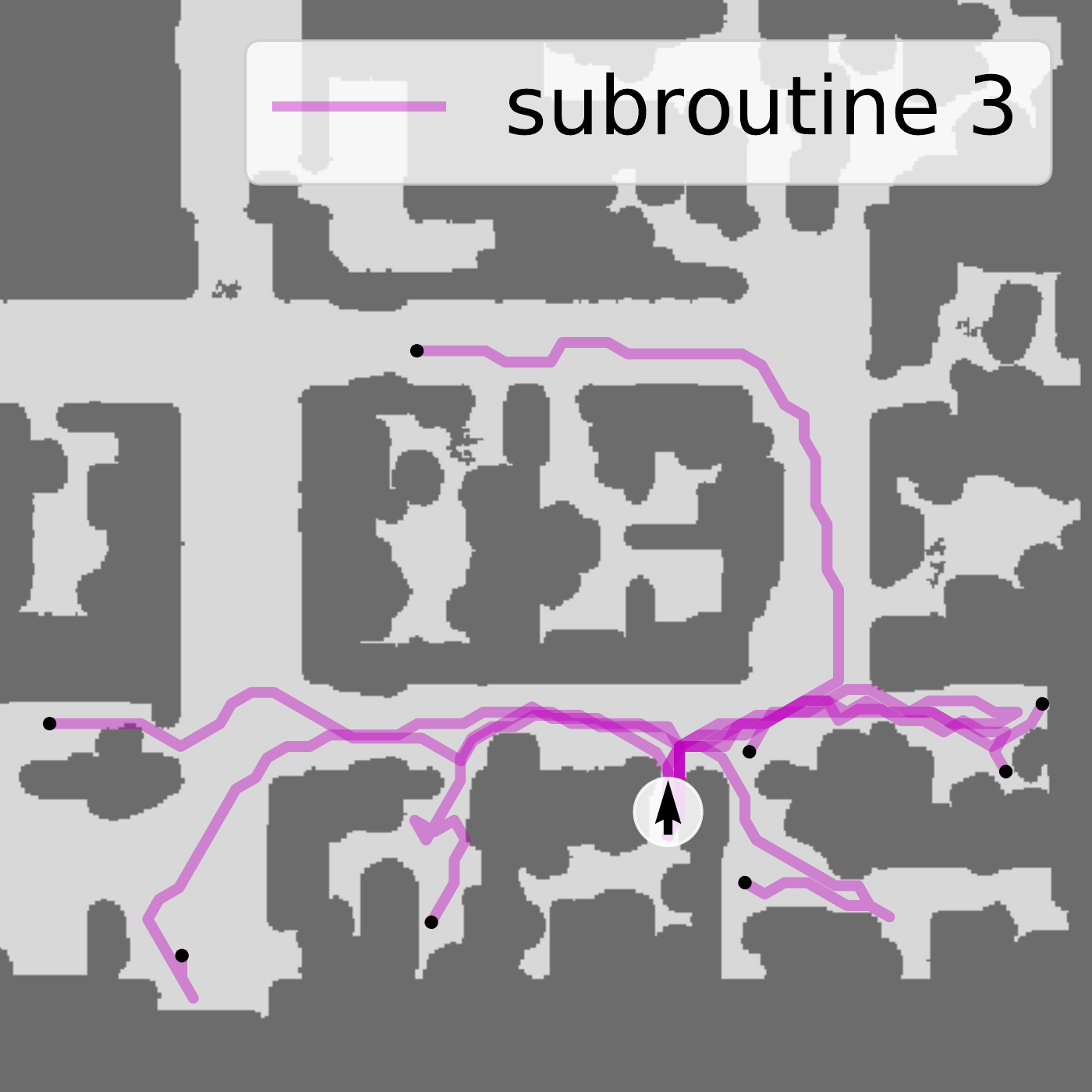}
\\
\insertWL{0.56}{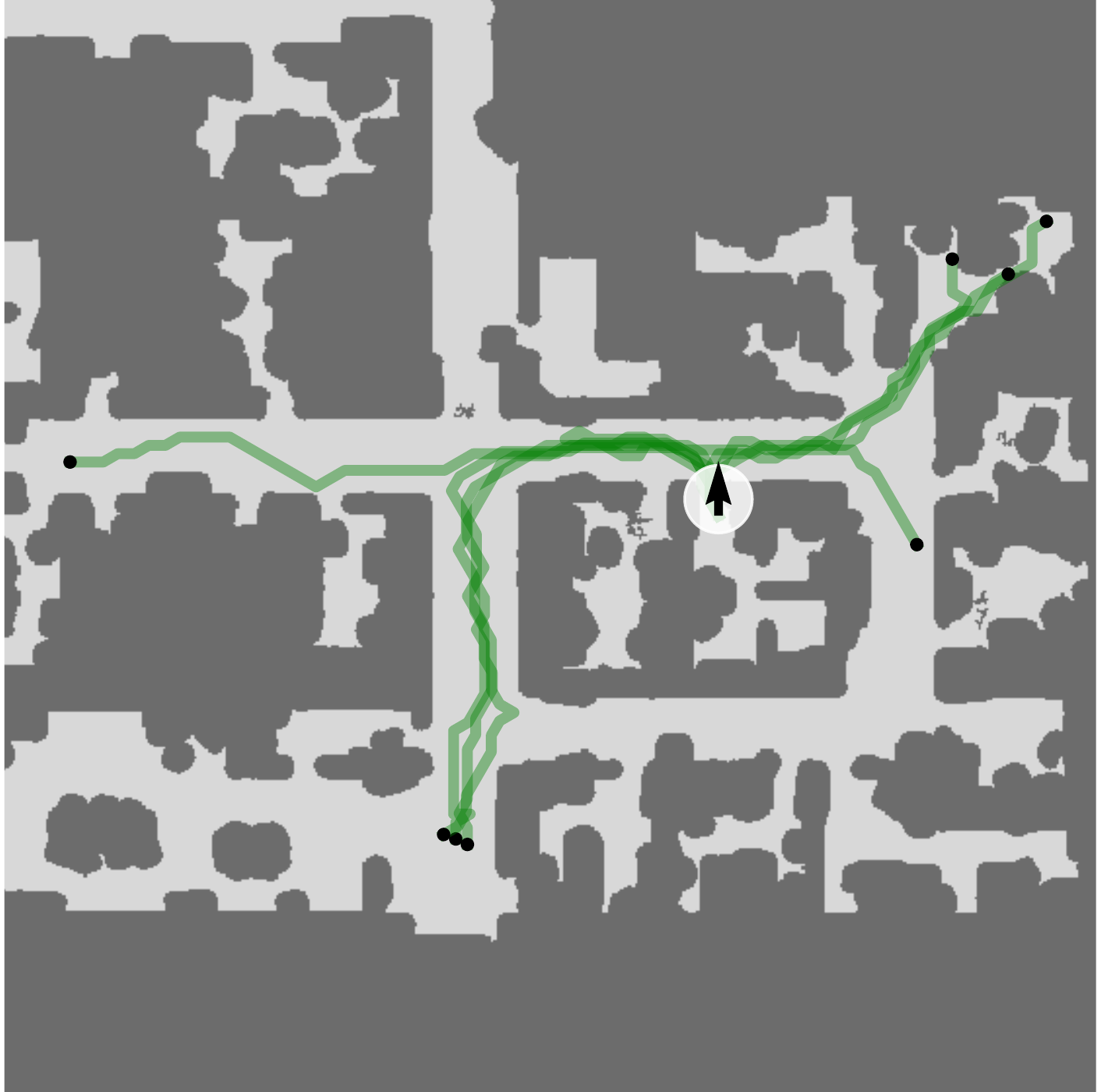}
\insertWL{0.56}{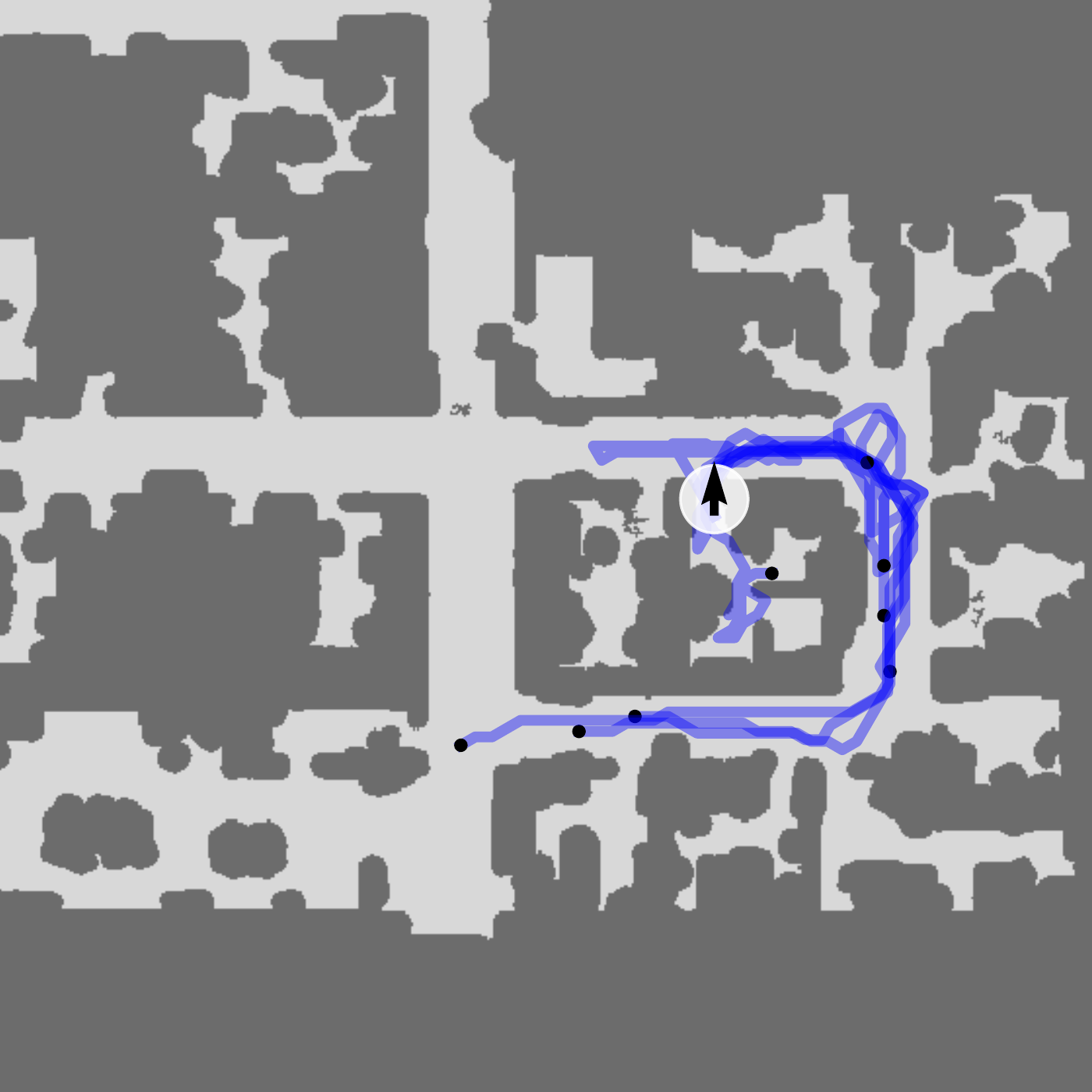}
\insertWL{0.56}{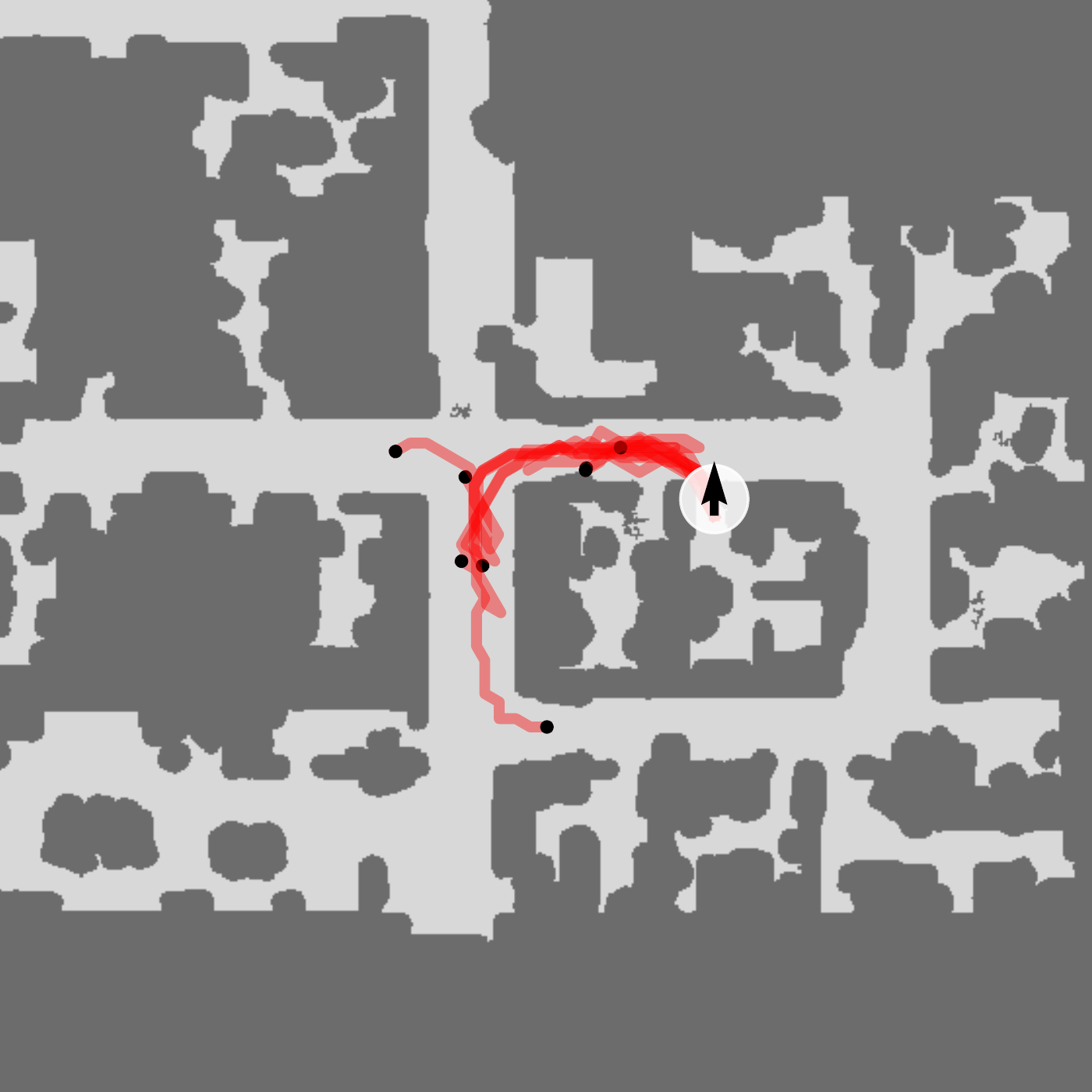}
\insertWL{0.56}{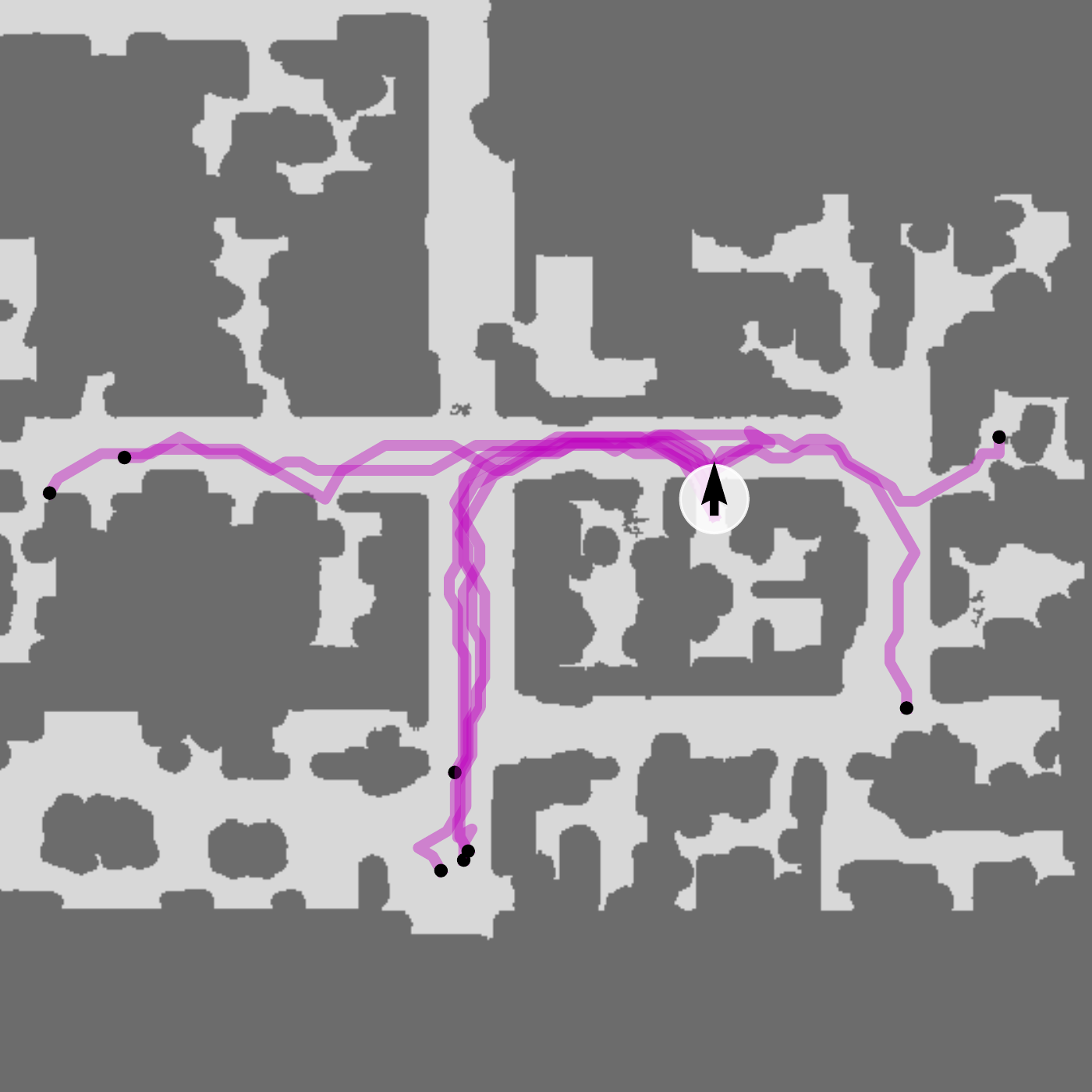}
\\
\insertWL{0.56}{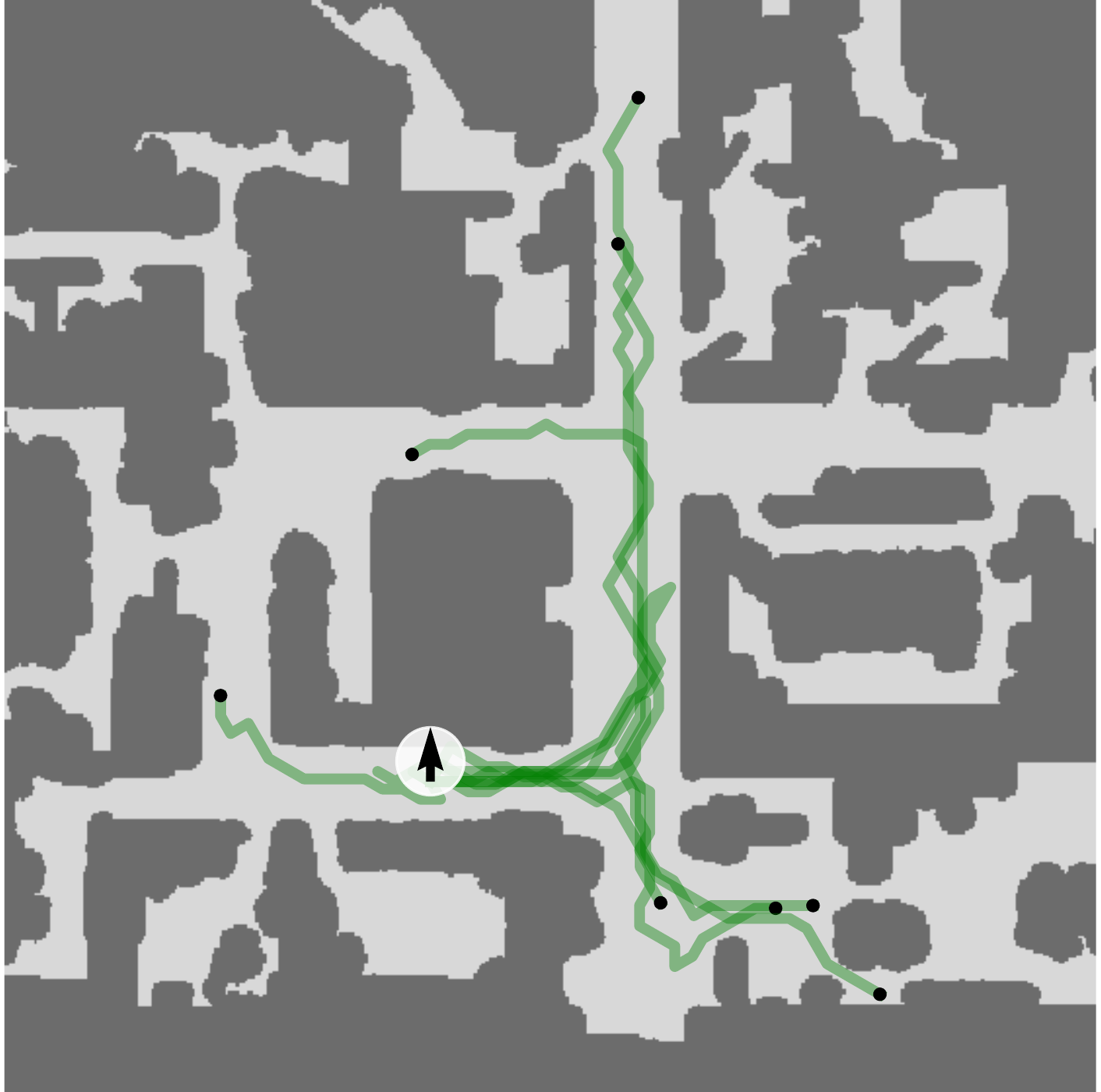}
\insertWL{0.56}{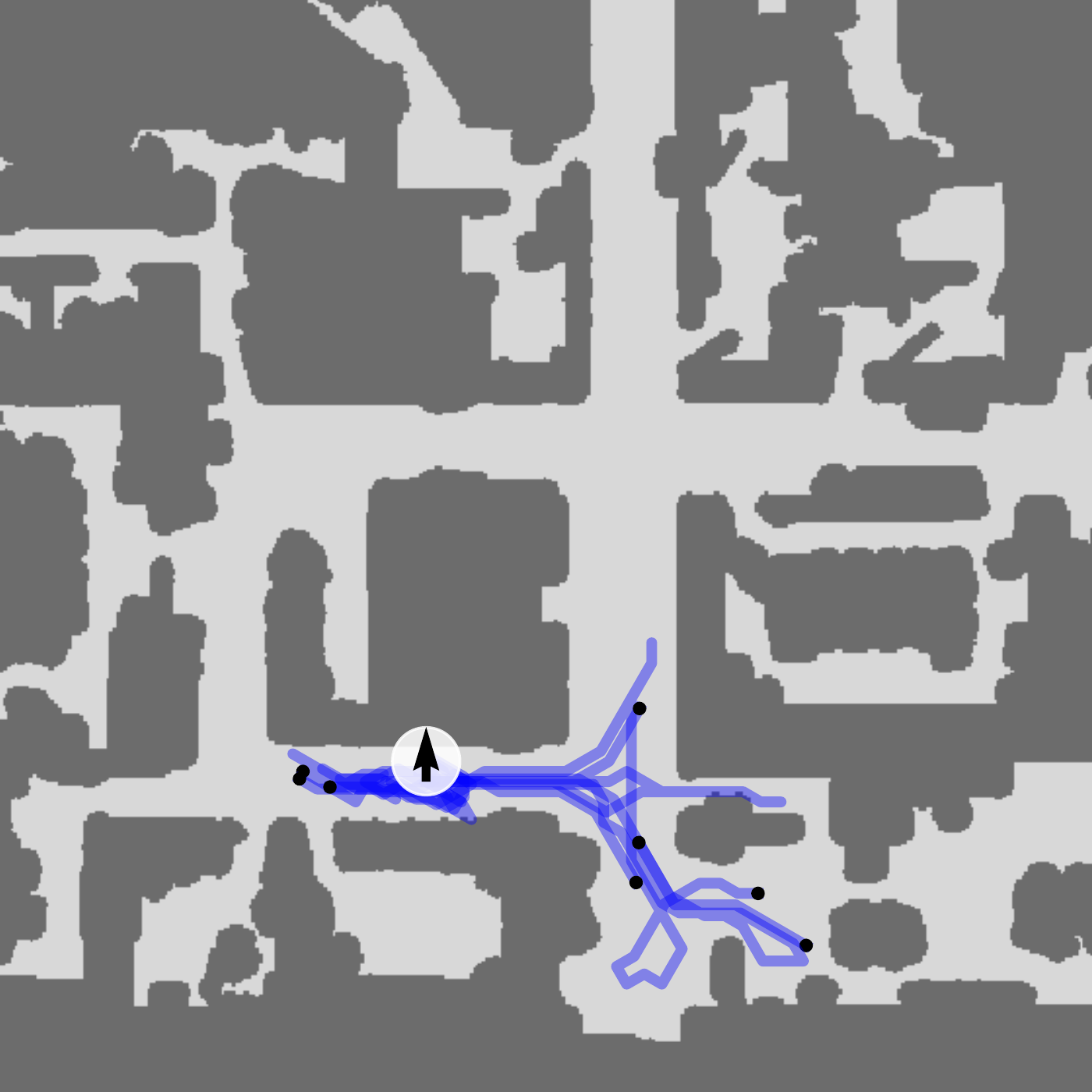}
\insertWL{0.56}{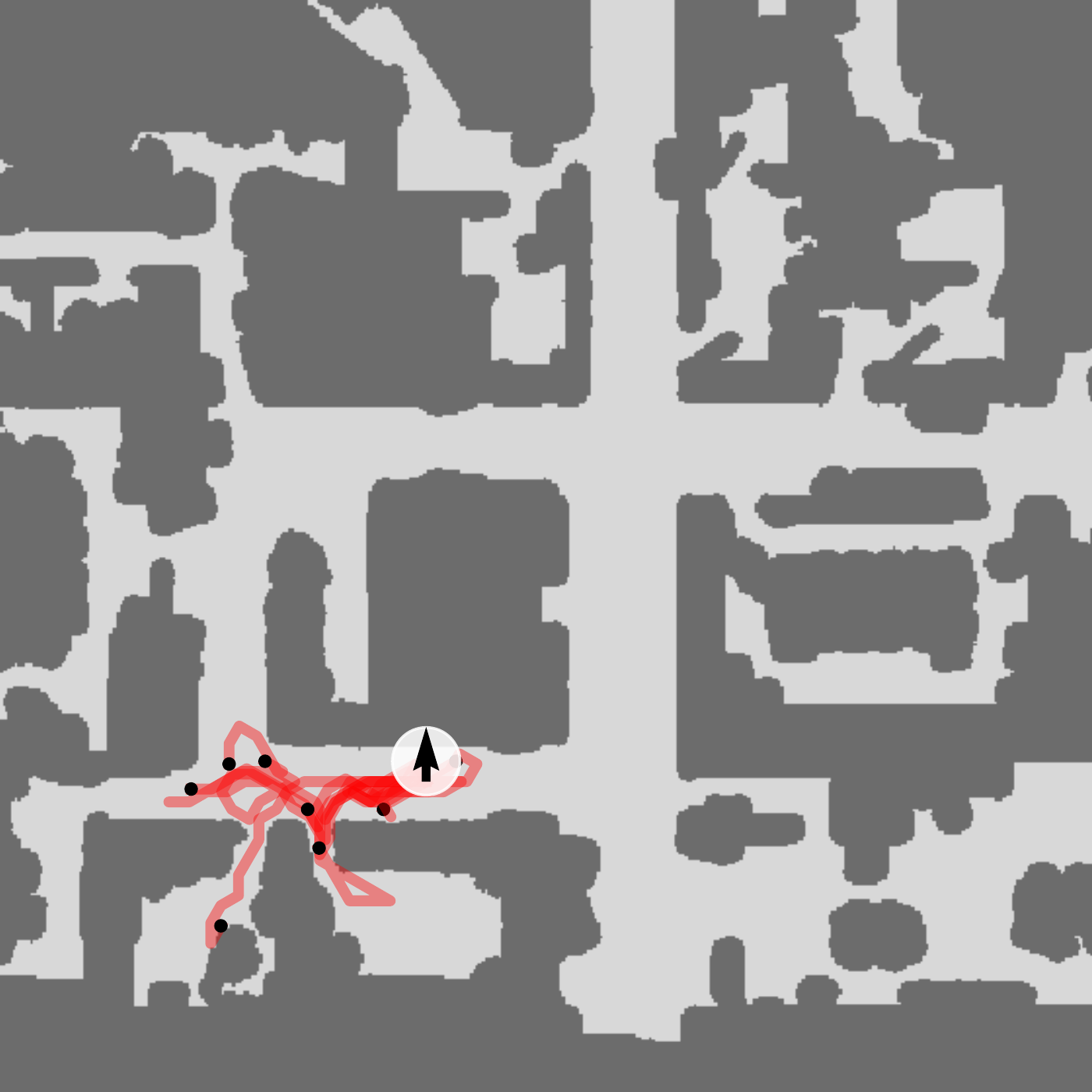}
\insertWL{0.56}{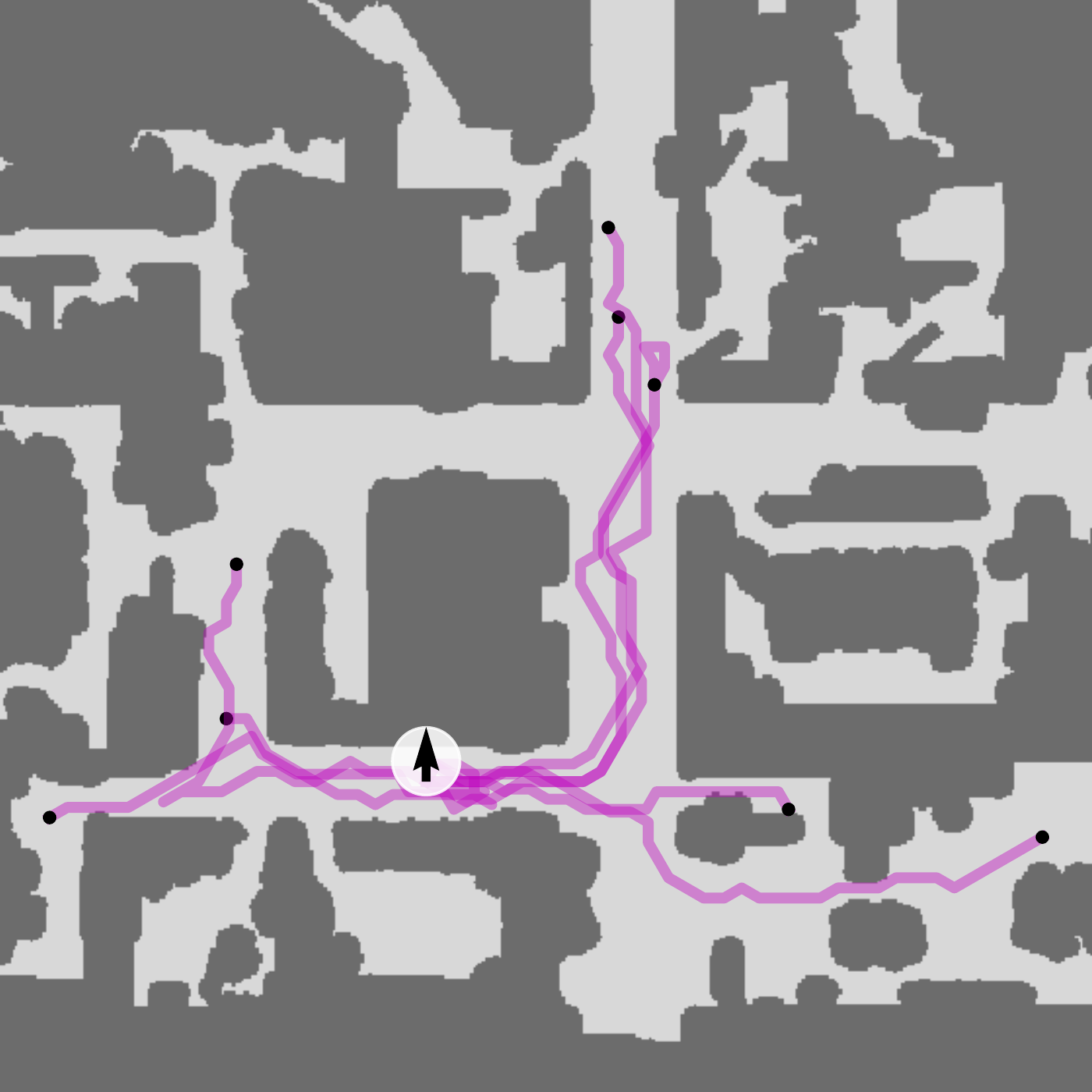}
\caption{\textbf{Subroutine Consistency and Diversity:} Each 
top-view figure shows multiple roll-outs of a subroutine from a given location. 
The black arrow in white circle shows starting position and the black dots shows the ending location of the rollouts. Columns show the same subroutine over different starting locations, illustrating the consistency of our subroutines while rows show different subroutines unrolled from the same location illustrating their diversity. It appears that \subR{1} prefers turning right and, \subR{2} prefers turning left. Note that policies only use first person views.}
\figlabel{consistency}
\end{figure}

%% file: vis_coverage.tex
\begin{figure}
\centering
\includegraphics[trim={0.35cm 0 0.28cm 0},clip,width=0.32\linewidth]{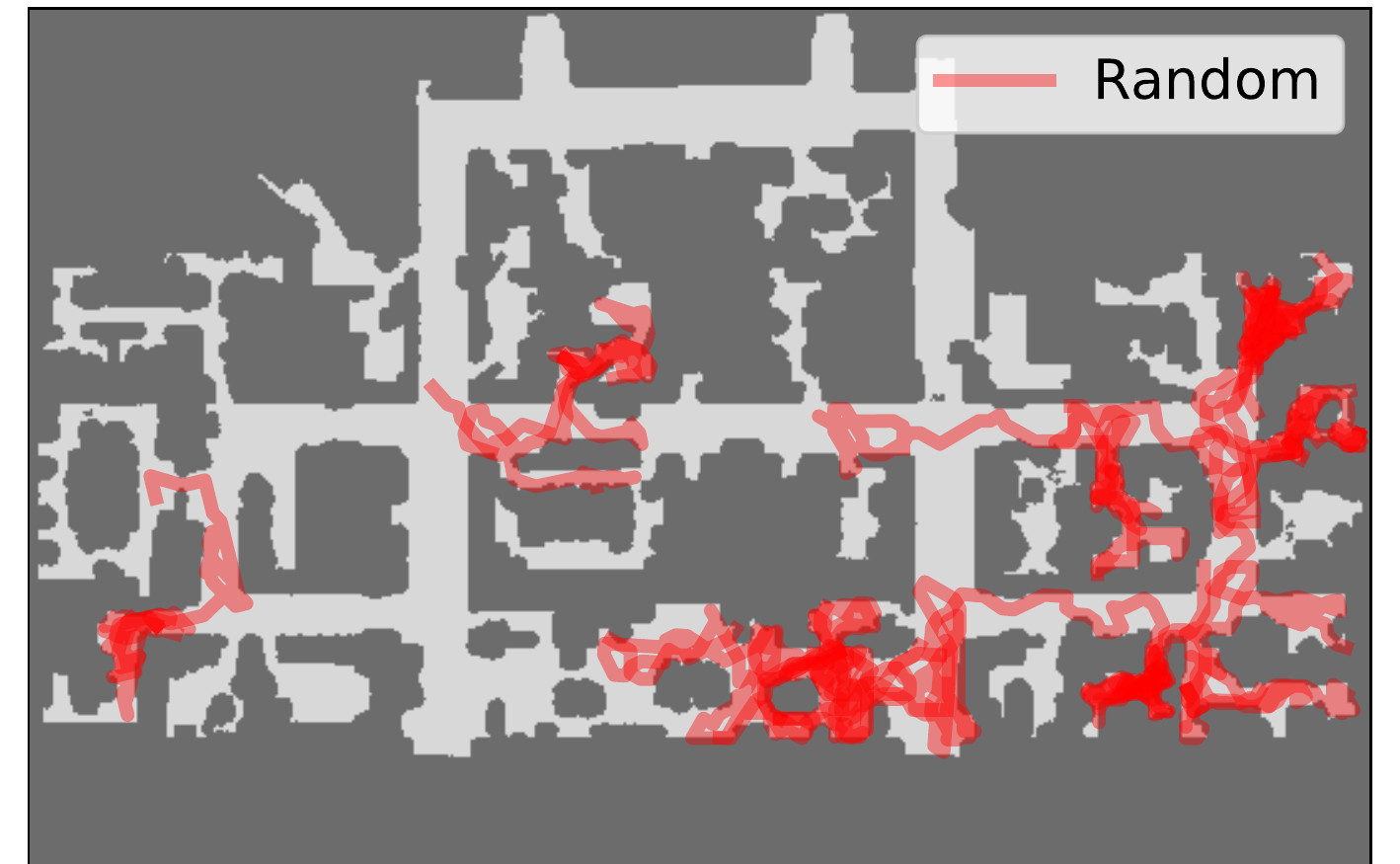}
\includegraphics[trim={0.35cm 0 0.28cm 0},clip,width=0.32\linewidth]{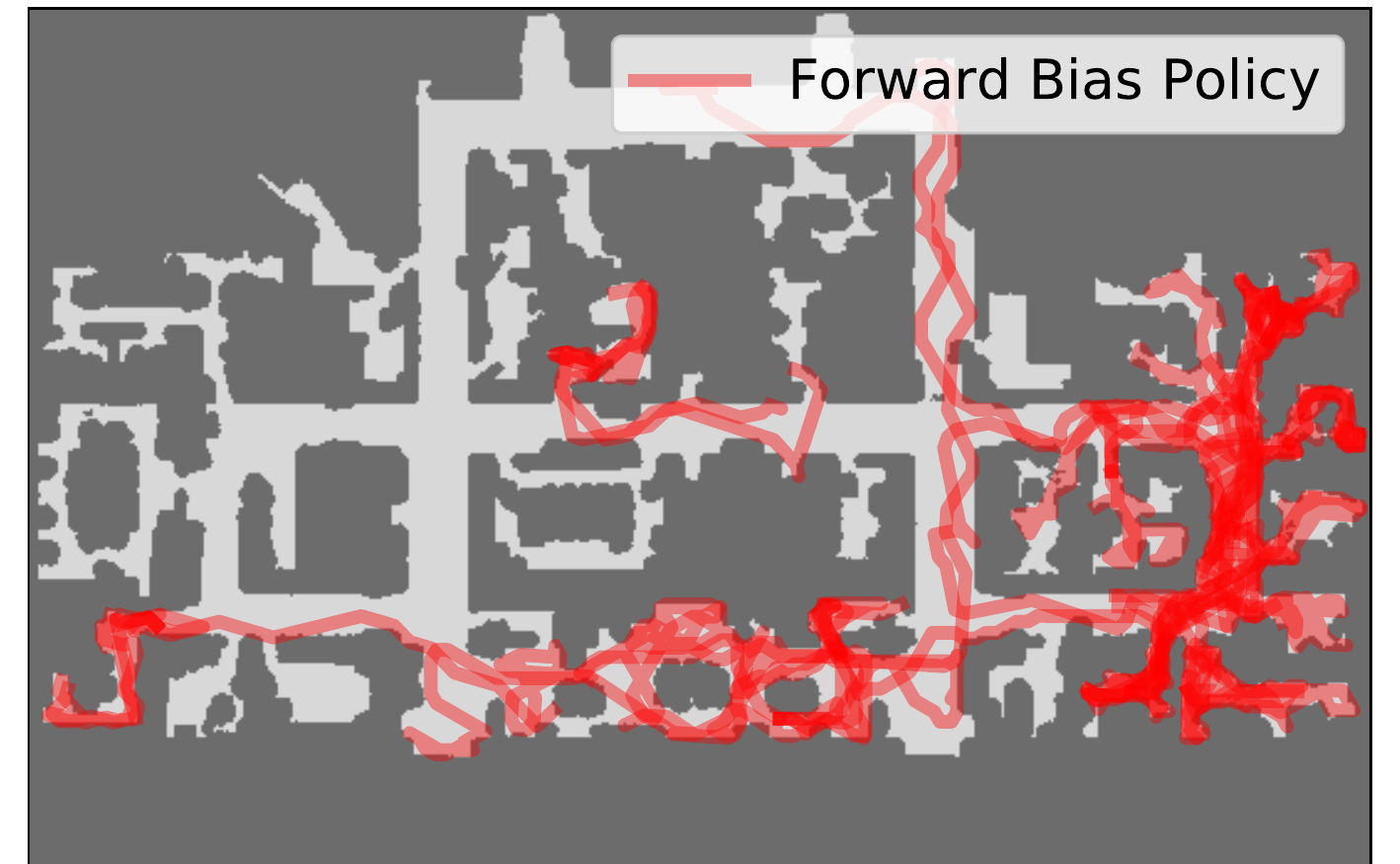}
\includegraphics[trim={0.35cm 0 0.28cm 0},clip,width=0.32\linewidth]{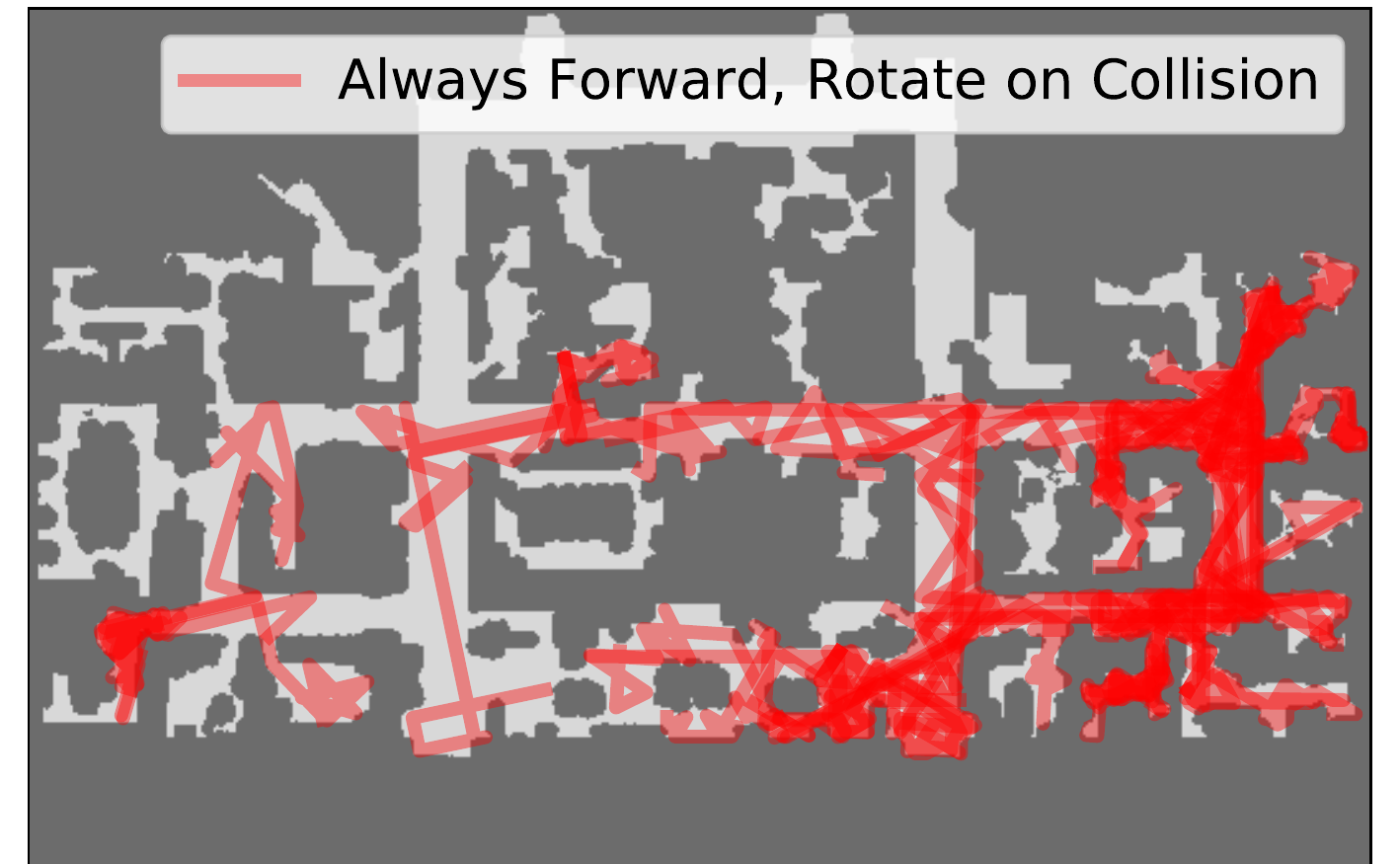} \\
\includegraphics[trim={0.35cm 0 0.28cm 0},clip,width=0.32\linewidth]{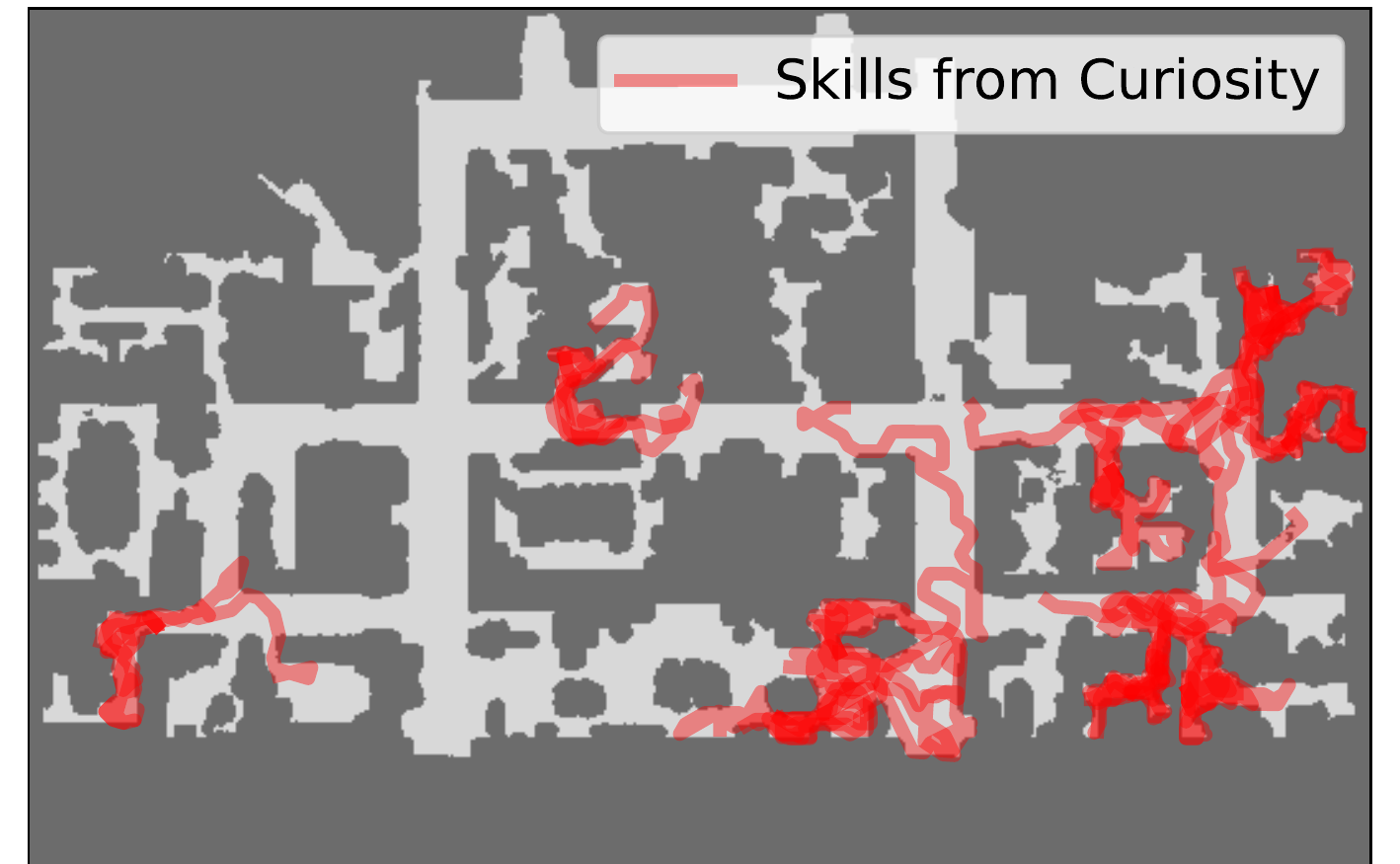}
\includegraphics[trim={0.35cm 0 0.28cm 0},clip,width=0.32\linewidth]{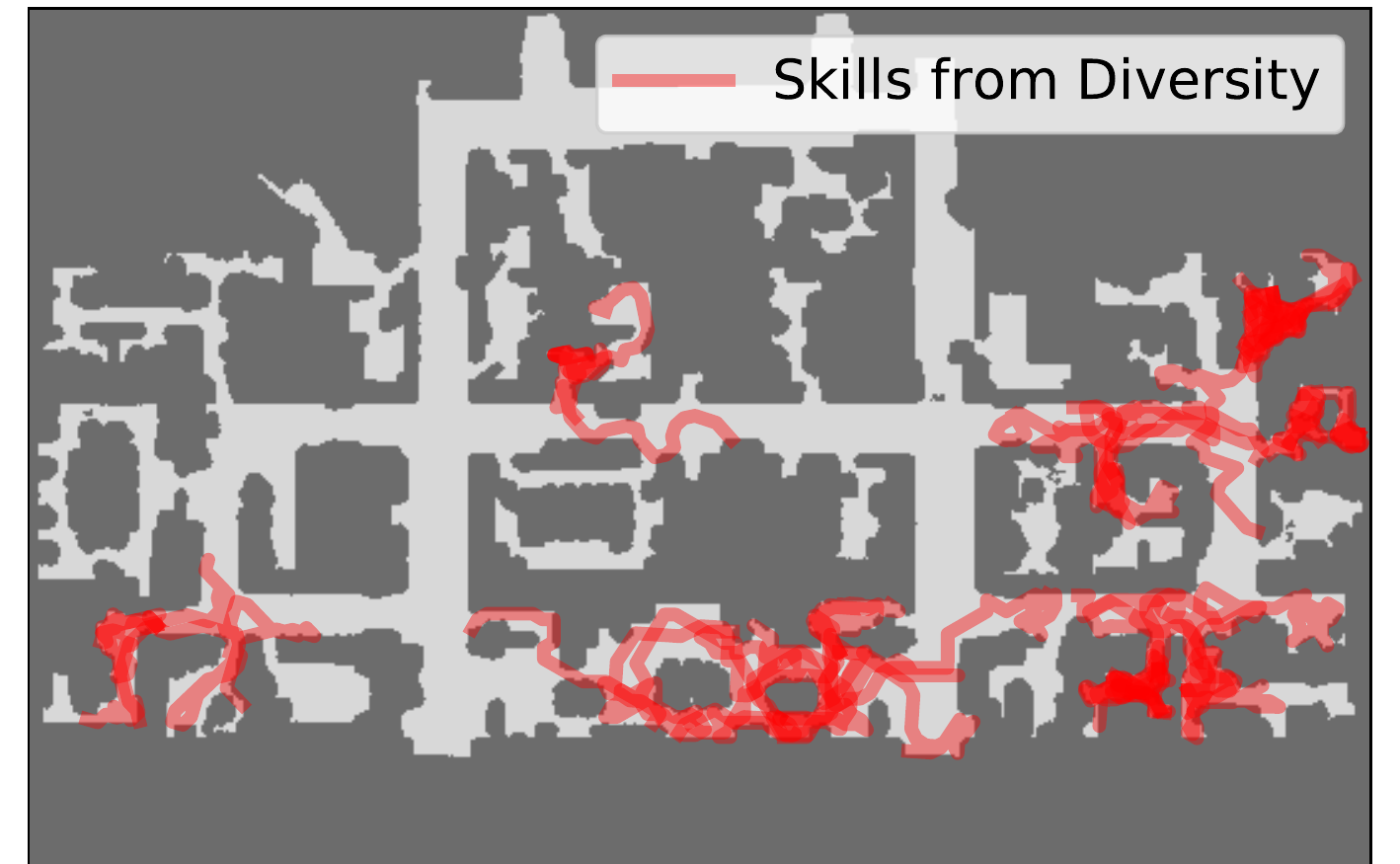}
\includegraphics[trim={0.35cm 0 0.28cm 0},clip,width=0.32\linewidth]{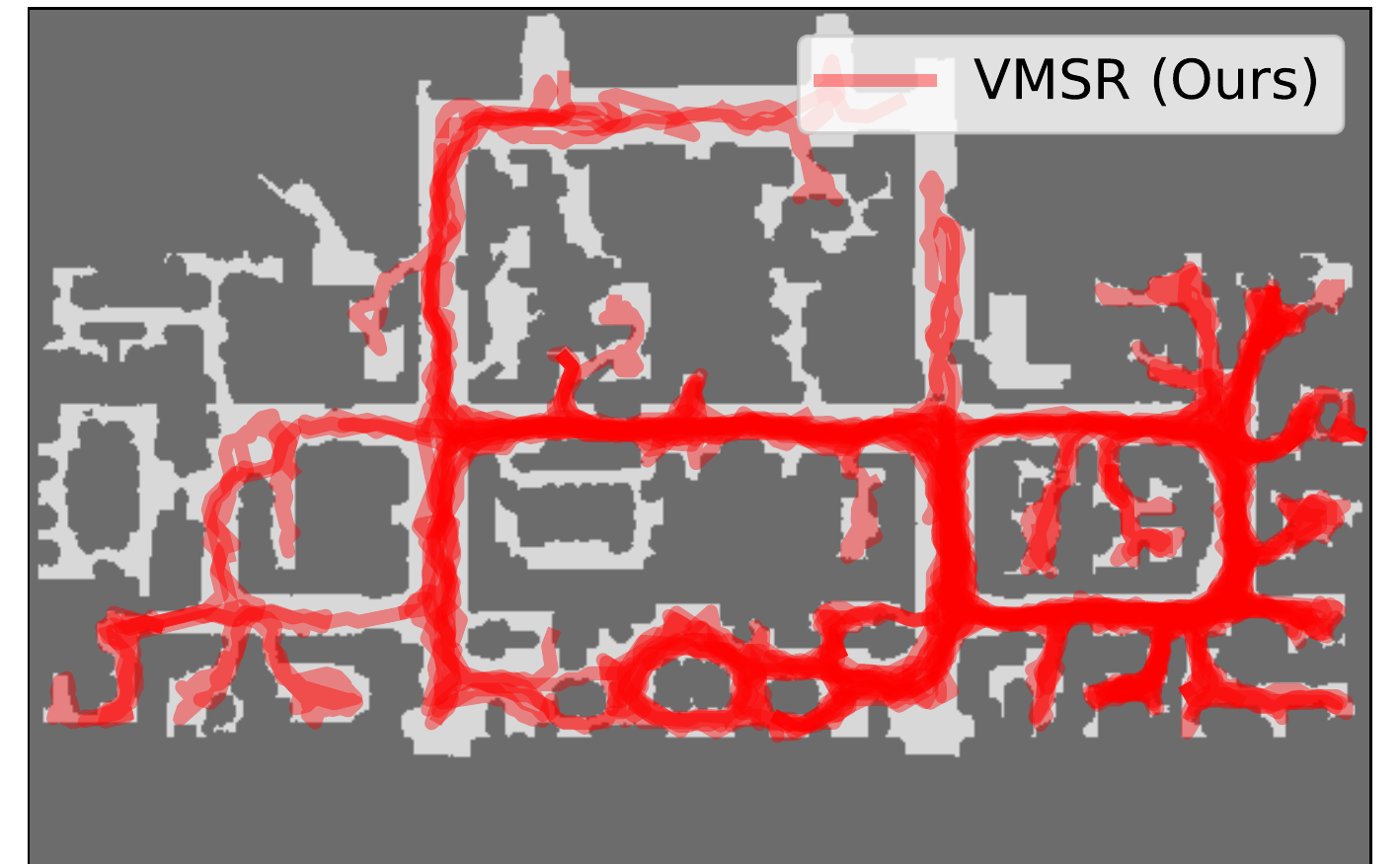}
\caption{\textbf{Coverage Visualization:} We show coverage of the overall space
after sampling 20 roll-outs from 11 different locations in the the test
environment \Etest. Note that \methodname covers more of the environment. It is
able to come out of rooms and different roll-outs go towards different areas.
Curiosity, diversity and Random policies spend most of their time inside rooms.
Policies that are biased to move forward do come out, but do not show diverse
behavior.  Visualizations show top view, however policies only use first person
views.}
\figlabel{coverage}
\end{figure}

%% file: ablation_plots.tex
\begin{figure*}
\centering
\insertWL{0.59}{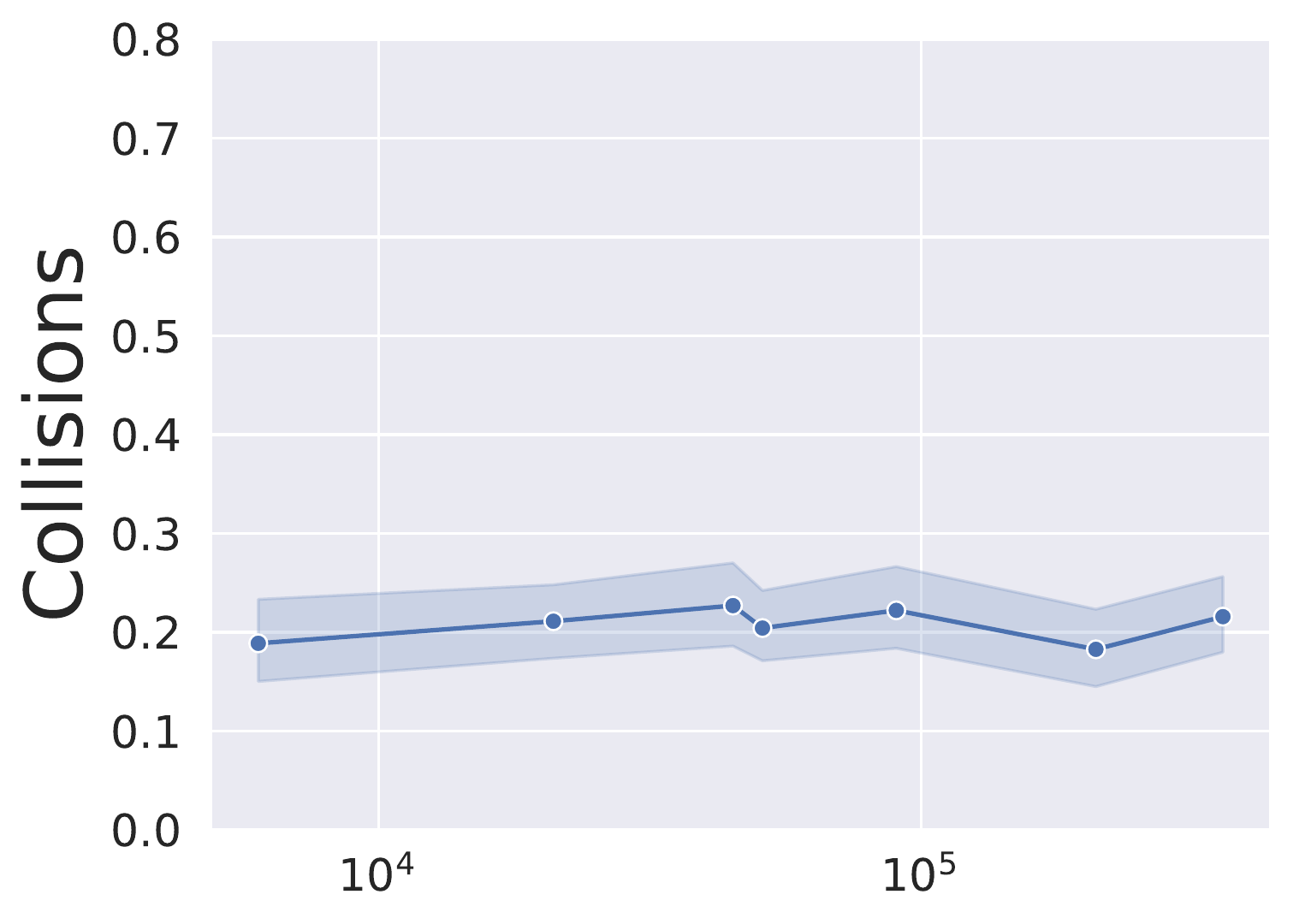}
\insertWL{0.55}{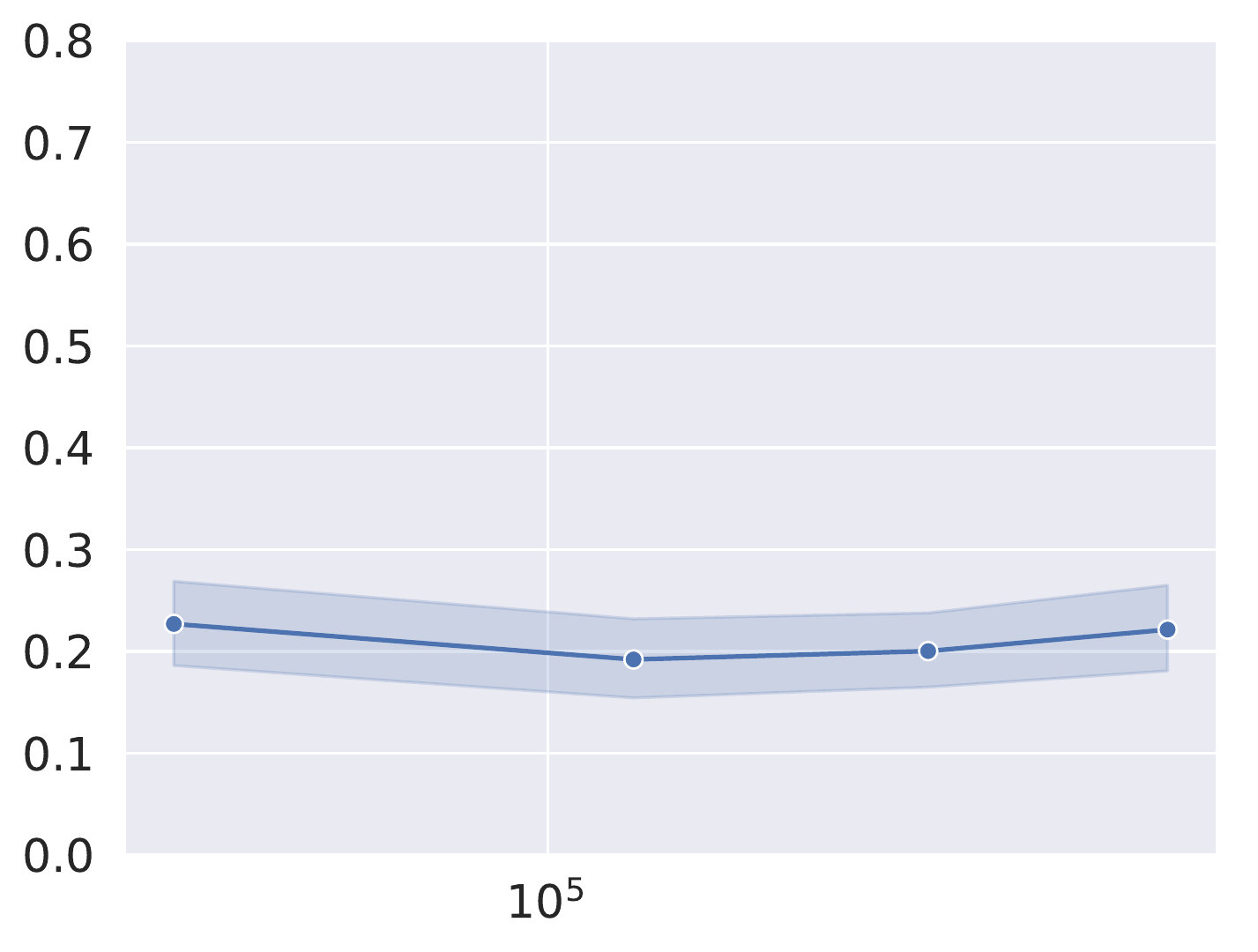}
\insertWL{0.55}{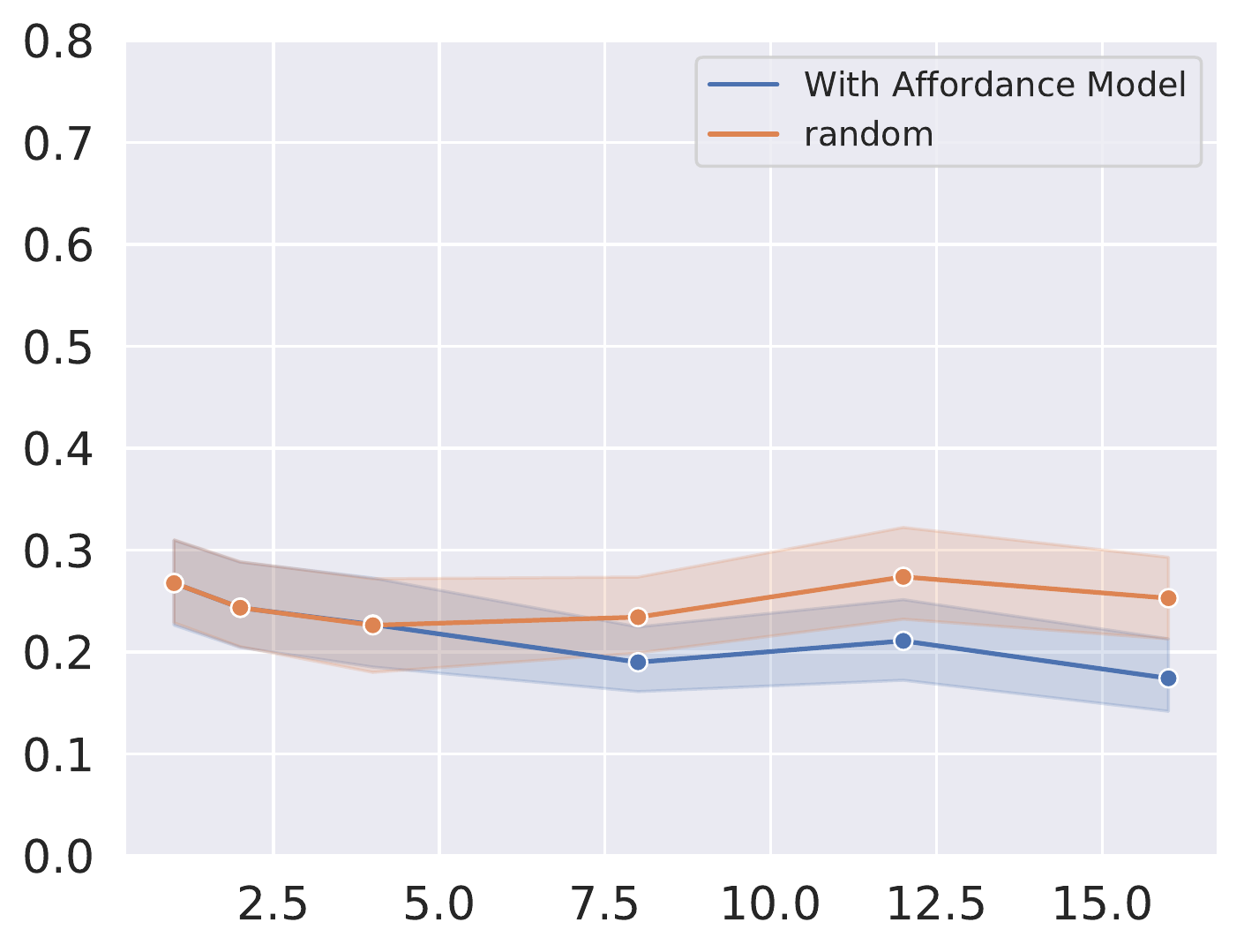}
\insertWL{0.55}{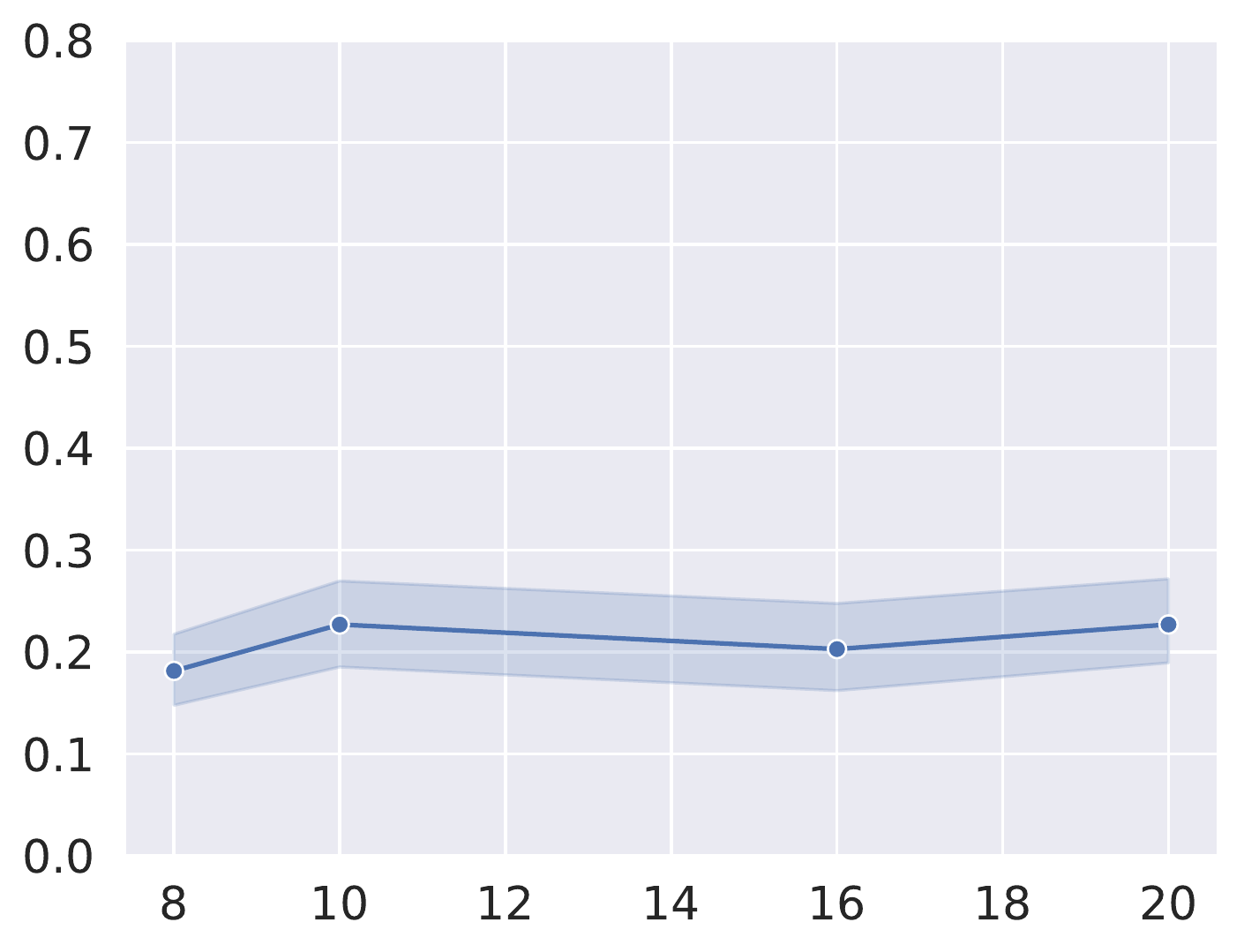}\\
\insertWL{0.59}{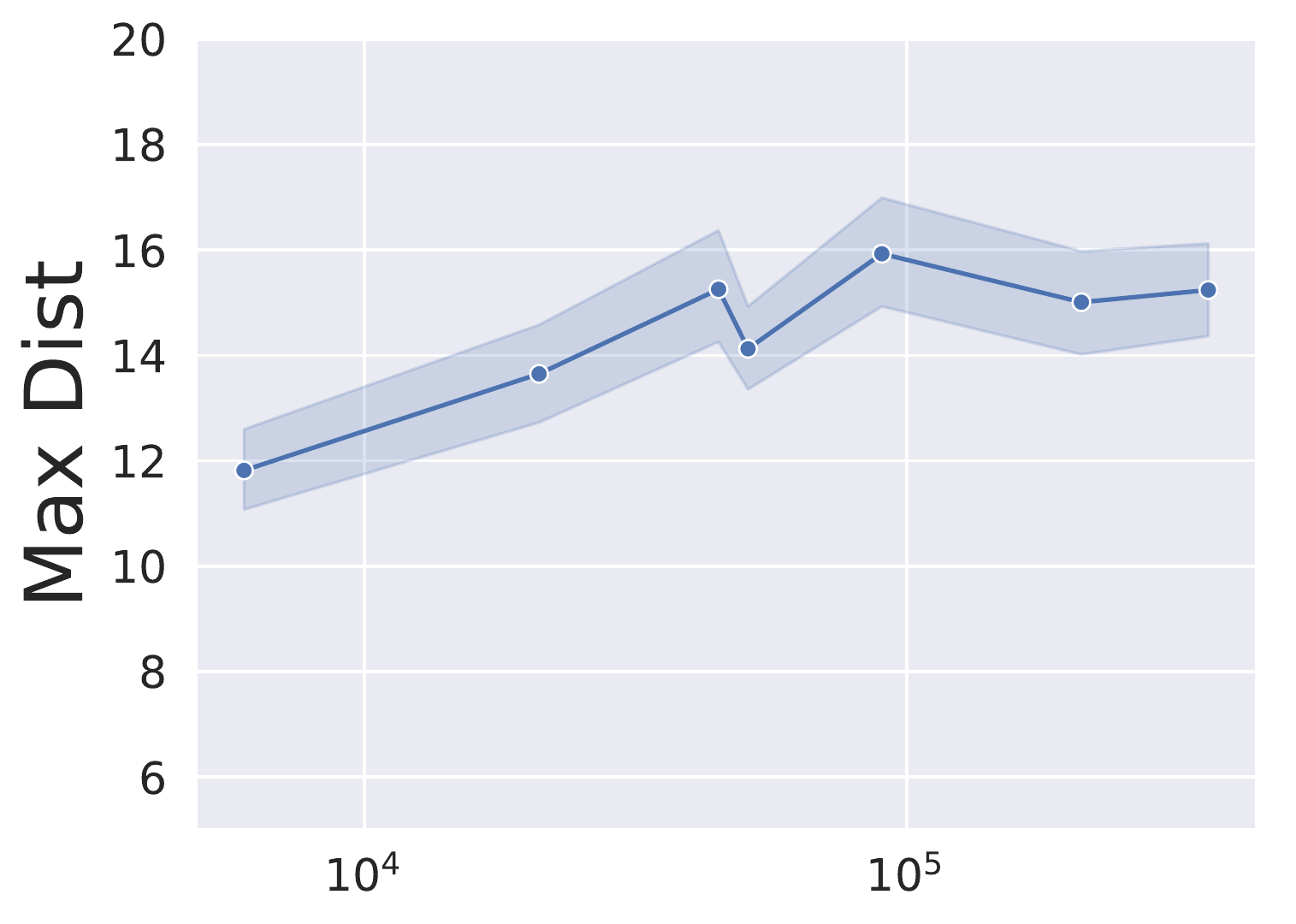}
\insertWL{0.55}{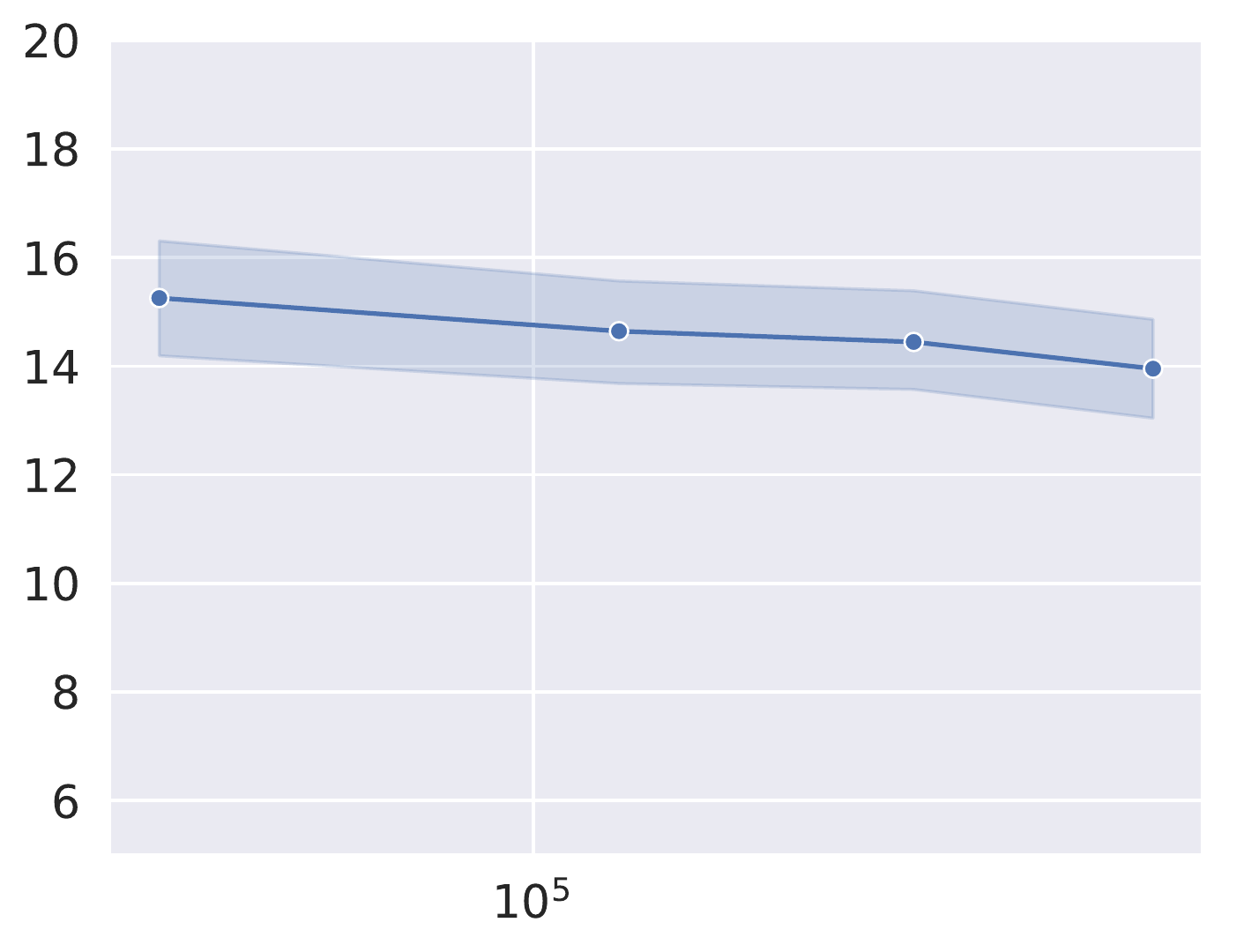}
\insertWL{0.55}{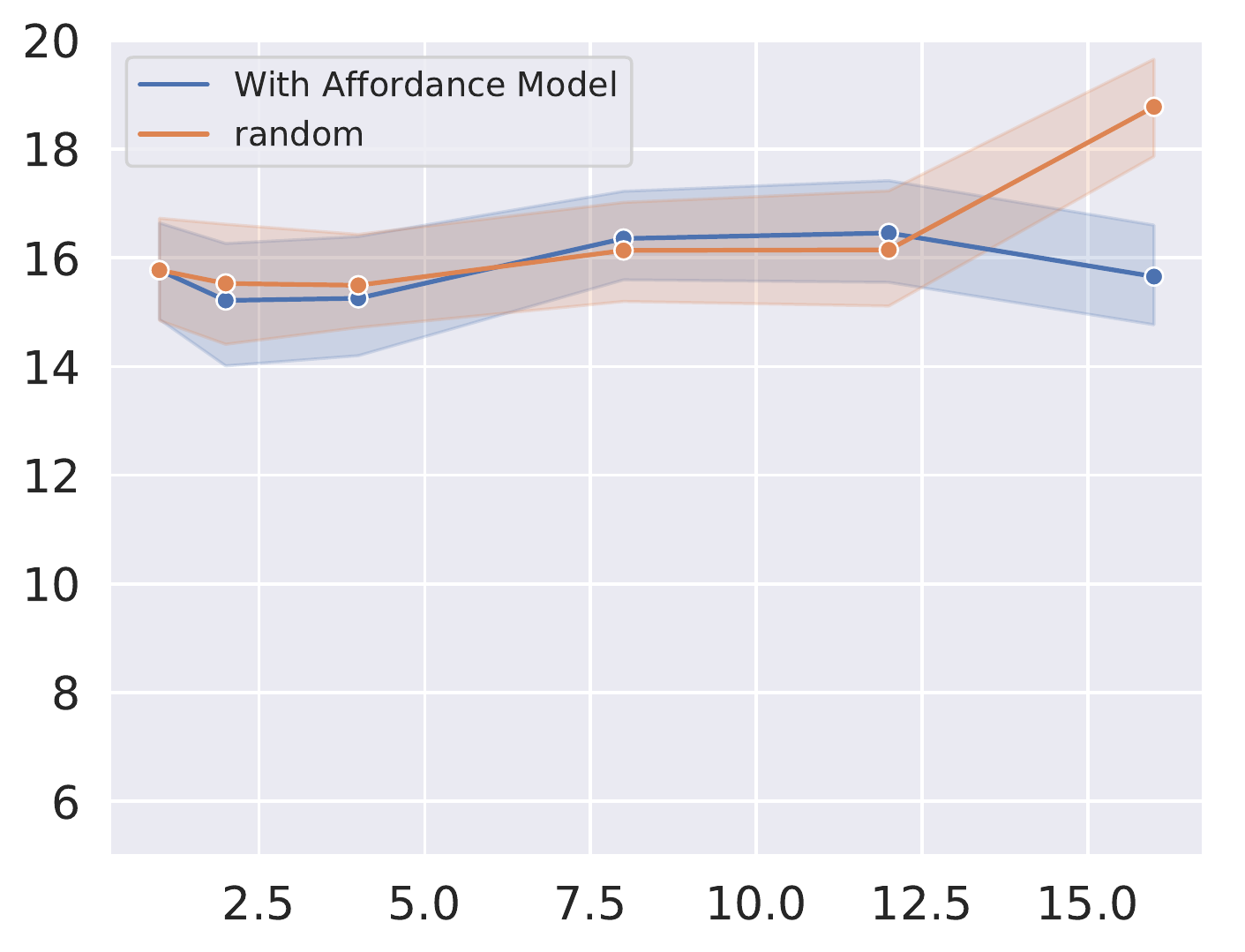}
\insertWL{0.55}{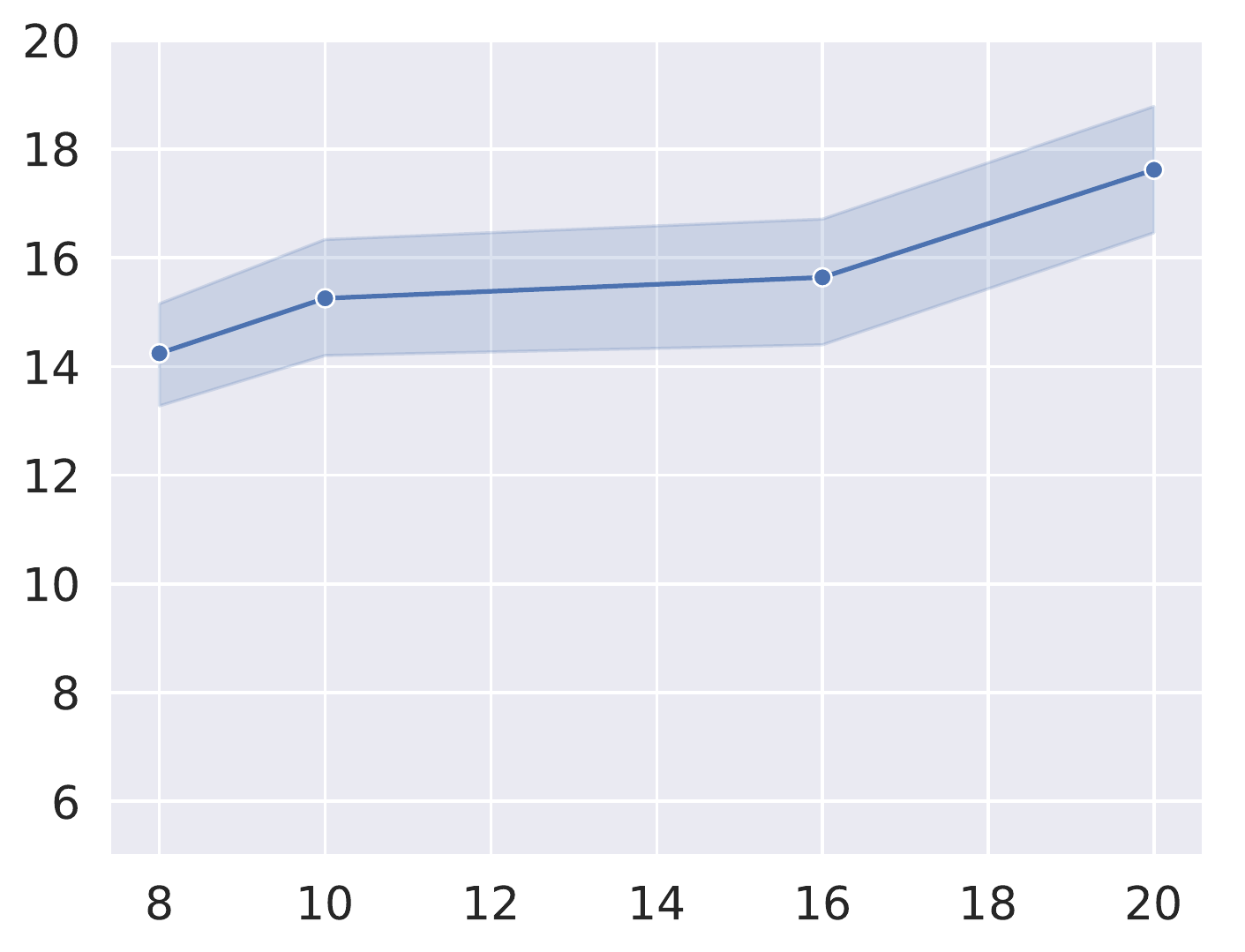}\\
\insertWL{0.59}{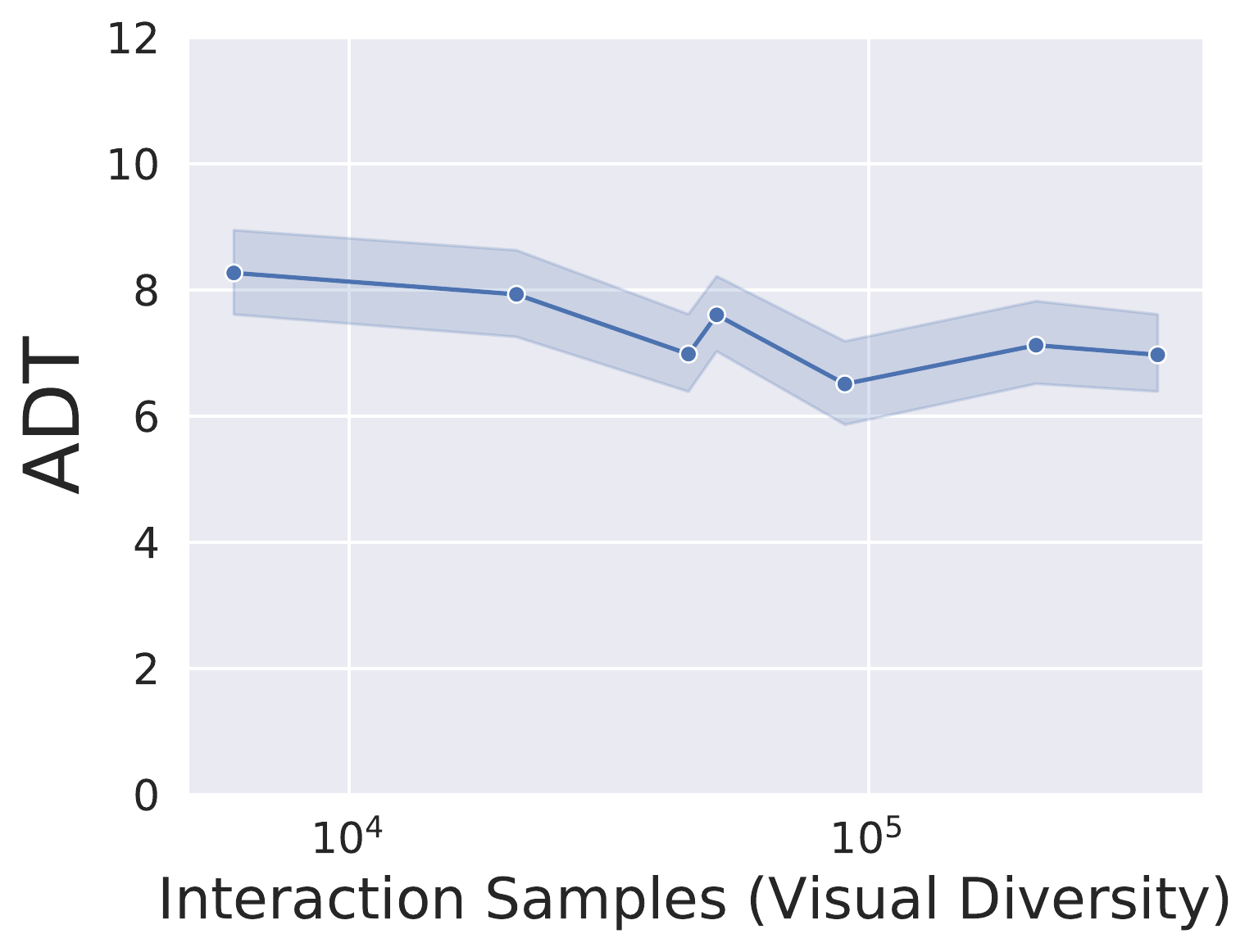}
\insertWL{0.55}{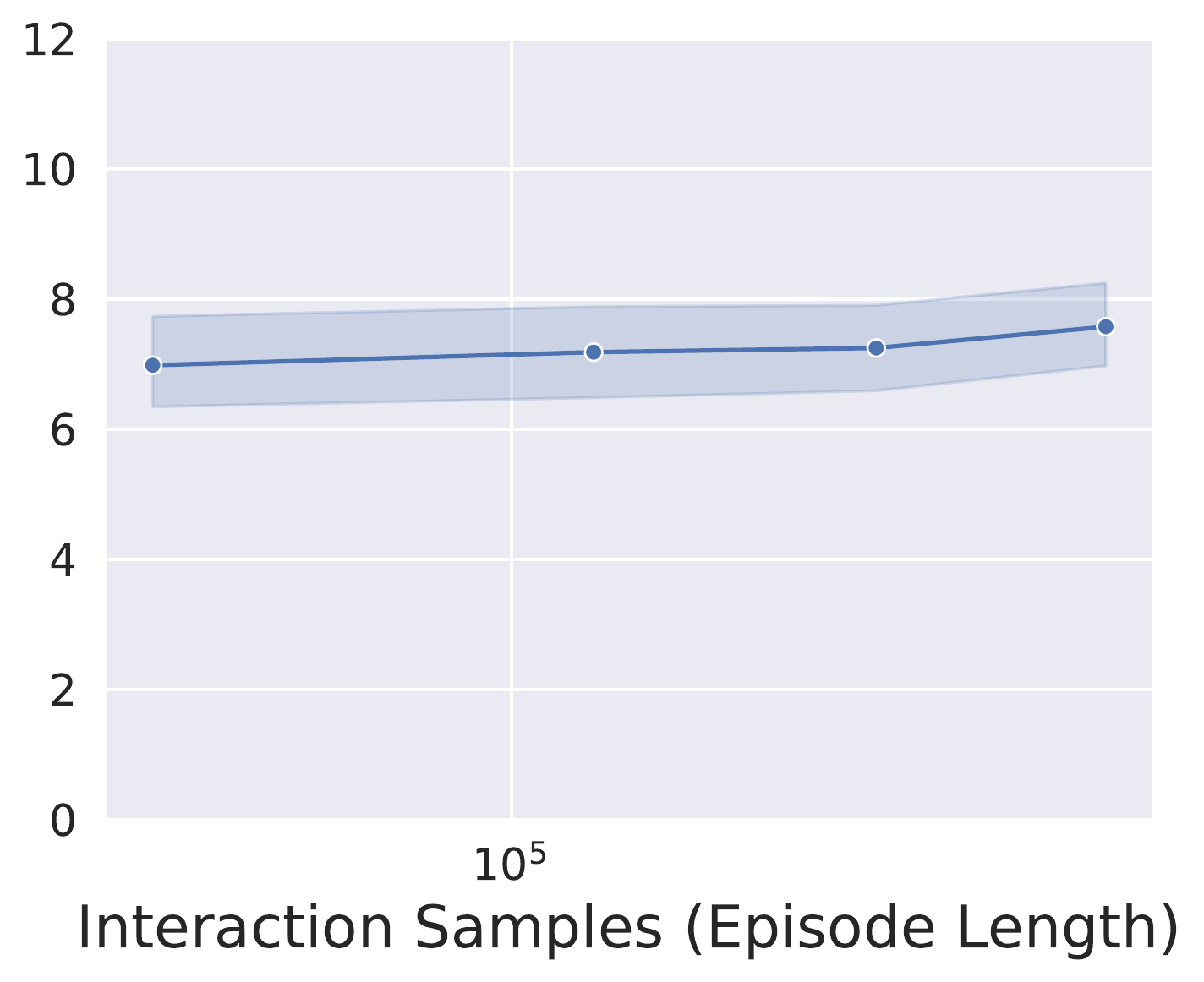}
\insertWL{0.55}{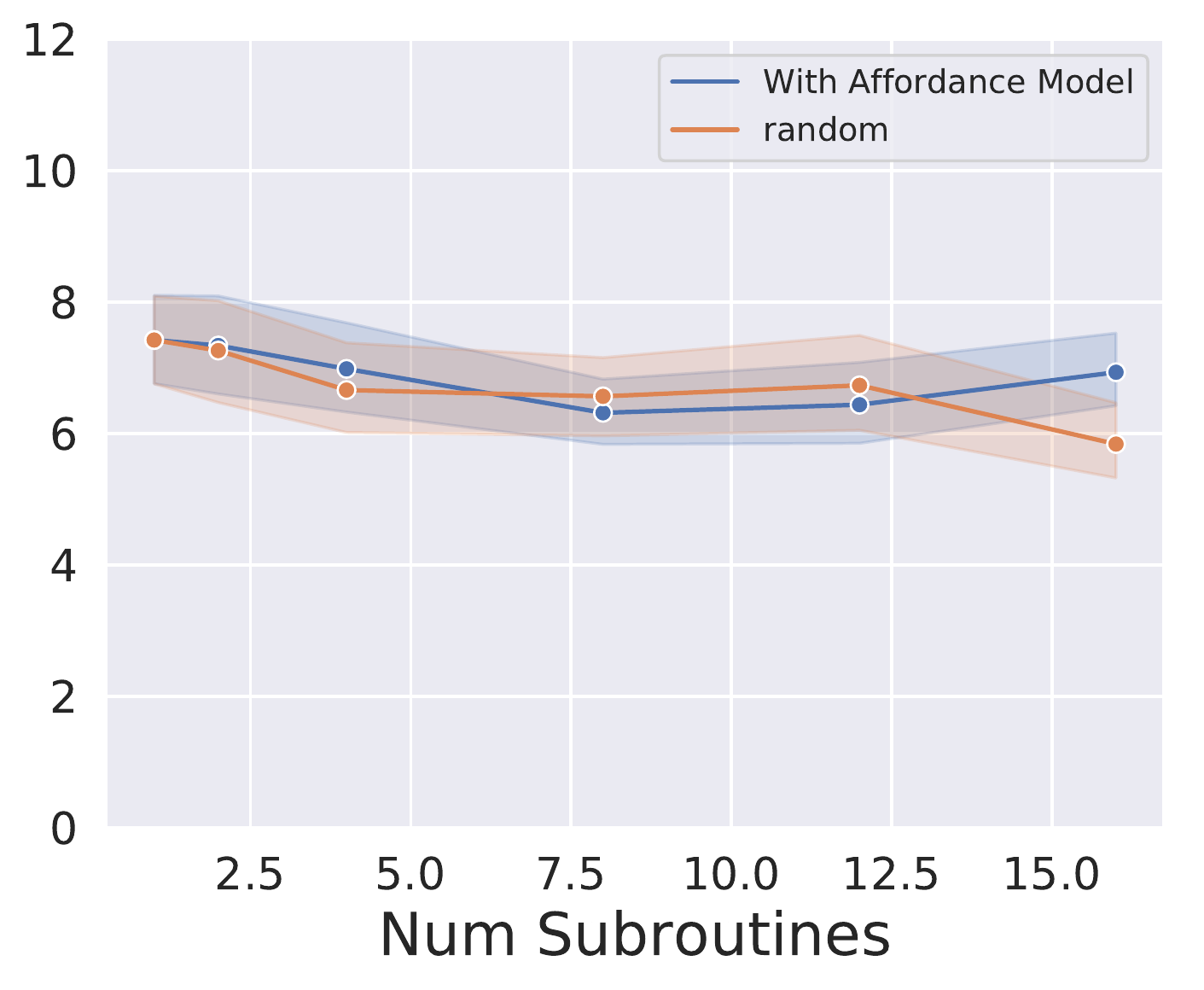}
\insertWL{0.55}{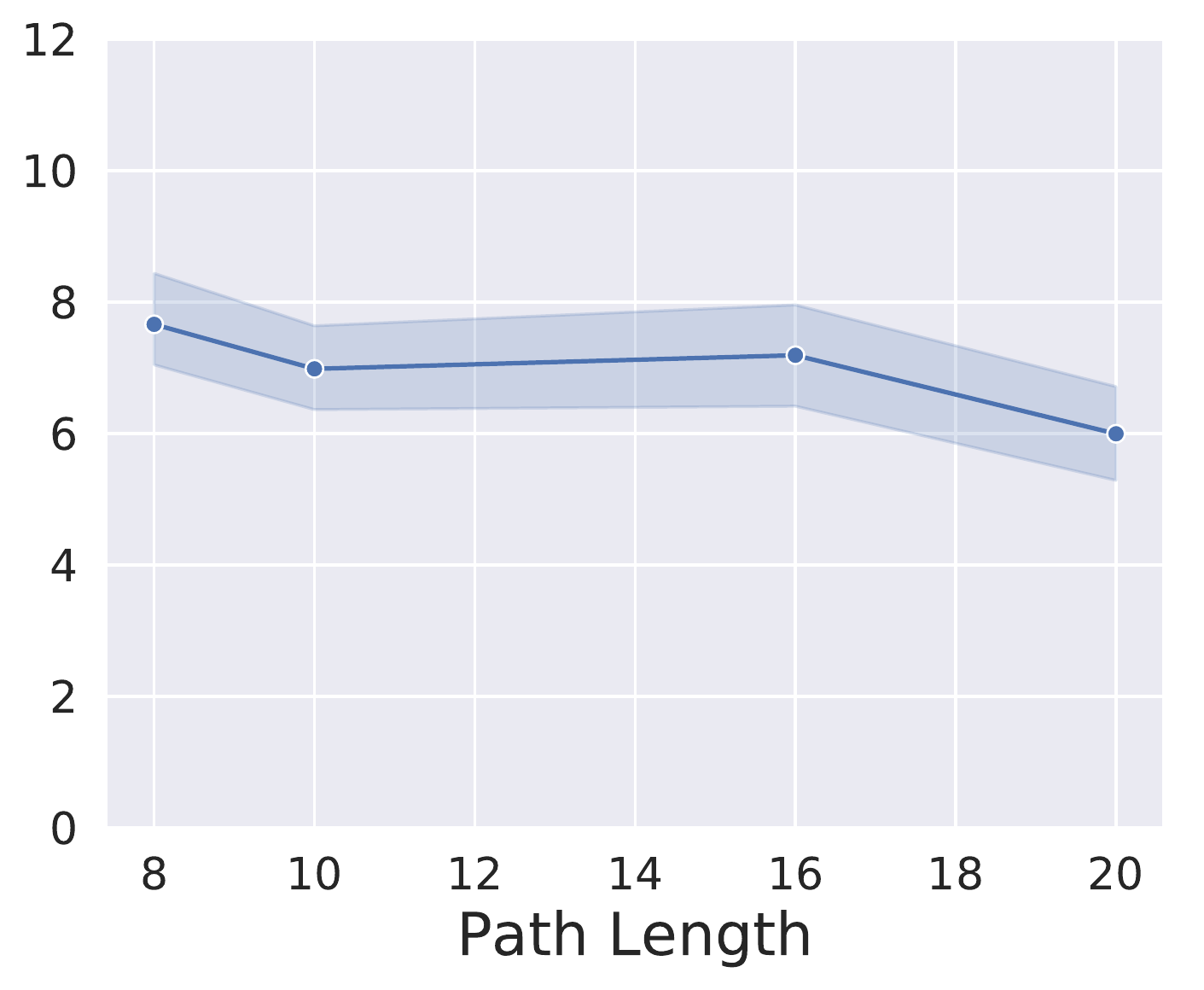}
\caption{\textbf{Dependence on active environment interaction samples, length of reference videos and number of subroutines specified}:
\textbf{Column 1 and 2}: We plot the exploration metrics against the number of
self supervision interaction samples. There are two orthogonal ways of achieving this --
increasing the number of restarts while keeping each episode length fixed (Col
1) and increasing the length of each self supervision episode while keeping the
number of restarts fixed (Col 2). We see that visual diversity improves
performance on Max Dist metric, but saturates at 45K 
interaction samples (1500 restarts with 30
steps each).  Performance roughly remains the same 
as we increase the episode length. \textbf{Column 3}: We change the
number of subroutines learned on the x-axis and compare the use of affordance
model for sampling subroutines to randomly sampling subroutines. Affordance
model shows improvement in collision rate over random sampling, indicating that
the affordance model better respects the constraints of the  physical space. We
don't see an improvement in the exploration metric or max distance metric.
\textbf{Column 4}: We observe improvements as we increase the path length of the
reference trajectories. Longer trajectories presumably allow \methodname to
learn more complex subroutines.}
\figlabel{ablations}
\end{figure*}

%% file: rl_setup.tex
\section{RL Experimental Setup}
We use \Etest for RL experiments. We use A2C to train all our algorithms on Point Goal task and Area Goal task. 
\begin{itemize}
\item \textbf{Area Goal:} The task is to find the nearest washroom. \Etest contains 2 washroom, and we start the agent 10-23 steps away from the nearest washroom. We randomly start the agent at a different location for every episode.
\item \textbf{Point Goal:} We specify the goal coordinates relative to the start position, and randomly sample the start and the goal locations every episode. The goal is 10-17 steps away from the start location. 
\end{itemize}

%% file: supp-settings.tex
\begin{table*}
\centering
\caption{Split of environments between different sets used in the paper. These
environments are from Stanford Building Parser Dataset (SBPD)
\citep{armeni20163d} and Matterport 3D Dataset (MP3D) \citep{Matterport3D}. We
fix a step size ($x$) and rotation angle ($\theta$) for each area by randomly
sampling from the list. For elevation angle and height of the robot, we
resample a value from the given ranges for \textit{every video}.}
\setlength{\tabcolsep}{6pt}
\renewcommand{\arraystretch}{1.2} 
\footnotesize
\resizebox{1.0\linewidth}{!}{
\begin{tabular}{lp{7cm}cccc}
\toprule
Split & Environments & \multicolumn{4}{c}{Agent Settings} \\
\cmidrule(l){3-6}
      & & Step Sizes & Rotation Angles & Elevations & Height \\ 
      & & ($x$ in $cm$) & ($\theta$) & ($\phi$) & ($h$ in $cm$) \\ 
\midrule
 
\Einv 
& \texttt{area1, area6, B6ByNegPMKs, Vvot9Ly1tCj} 
& 20, 50, 80 & 36$^\circ$, 24$^\circ$, 18$^\circ$ & [-25$^\circ$, 5$^\circ$] & [90, 150] 
\\
\Evideo 
& \texttt{area5a, area5b, p5wJjkQkbXX, VFuaQ6m2Qom, 2n8kARJN3HM, SN83YJsR3w2} 
& 30, 60, 90 & 40$^\circ$, 30$^\circ$, 24$^\circ$, 20$^\circ$ & [-35$^\circ$, -5$^\circ$] & [80, 160] 
\\
\Eval 
& \texttt{area3}
& 40 & 30$^\circ$ & -15$^\circ$ & 120 
\\ 
\Etest 
& \texttt{area4}
& 40 & 30$^\circ$ & -15$^\circ$ & 120 
\\ 
\bottomrule
\end{tabular}}
\tablelabel{agent-settings}
\end{table*}

%% file: vmsr-corl.bbl
\begin{thebibliography}{43}
\providecommand{\natexlab}[1]{#1}
\providecommand{\url}[1]{\texttt{#1}}
\expandafter\ifx\csname urlstyle\endcsname\relax
  \providecommand{\doi}[1]{doi: #1}\else
  \providecommand{\doi}{doi: \begingroup \urlstyle{rm}\Url}\fi

\bibitem[Fikes and Nilsson(1971)]{fikes1971strips}
R.~E. Fikes and N.~J. Nilsson.
\newblock {STRIPS}: A new approach to the application of theorem proving to
  problem solving.
\newblock \emph{Artificial intelligence}, 1971.

\bibitem[Sutton et~al.(1999)Sutton, Precup, and Singh]{sutton1999between}
R.~S. Sutton, D.~Precup, and S.~Singh.
\newblock Between mdps and semi-mdps: A framework for temporal abstraction in
  reinforcement learning.
\newblock \emph{Artificial intelligence}, 1999.

\bibitem[Xu et~al.(2017)Xu, Gao, Yu, and Darrell]{xu2017end}
H.~Xu, Y.~Gao, F.~Yu, and T.~Darrell.
\newblock End-to-end learning of driving models from large-scale video
  datasets.
\newblock In \emph{CVPR}, pages 2174--2182, 2017.

\bibitem[Peng et~al.(2018{\natexlab{a}})Peng, Kanazawa, Malik, Abbeel, and
  Levine]{peng2018sfv}
X.~B. Peng, A.~Kanazawa, J.~Malik, P.~Abbeel, and S.~Levine.
\newblock Sfv: Reinforcement learning of physical skills from videos.
\newblock In \emph{SIGGRAPH Asia 2018}, page 178. ACM, 2018{\natexlab{a}}.

\bibitem[Peng et~al.(2018{\natexlab{b}})Peng, Abbeel, Levine, and van~de
  Panne]{peng2018deepmimic}
X.~B. Peng, P.~Abbeel, S.~Levine, and M.~van~de Panne.
\newblock Deepmimic: Example-guided deep reinforcement learning of
  physics-based character skills.
\newblock \emph{ACM Transactions on Graphics (TOG)}, 37\penalty0 (4):\penalty0
  143, 2018{\natexlab{b}}.

\bibitem[LaValle(2006)]{Lavalle06book}
S.~M. LaValle.
\newblock \emph{Planning Algorithms}.
\newblock Cambridge University Press, Cambridge, U.K., 2006.
\newblock Available at http://planning.cs.uiuc.edu/.

\bibitem[Thrun et~al.(2005)Thrun, Burgard, and Fox]{thrun2005probabilistic}
S.~Thrun, W.~Burgard, and D.~Fox.
\newblock \emph{Probabilistic robotics}.
\newblock MIT press, 2005.

\bibitem[Hauser et~al.(2008)Hauser, Bretl, Harada, and
  Latombe]{hauser2008using}
K.~Hauser, T.~Bretl, K.~Harada, and J.-C. Latombe.
\newblock Using motion primitives in probabilistic sample-based planning for
  humanoid robots.
\newblock In \emph{Algorithmic Foundation of Robotics}. 2008.

\bibitem[Schaal(2006)]{schaal2006dynamic}
S.~Schaal.
\newblock Dynamic movement primitives-a framework for motor control in humans
  and humanoid robotics.
\newblock In \emph{Adaptive motion of animals and machines}, pages 261--280.
  Springer, 2006.

\bibitem[Ijspeert et~al.(2013)Ijspeert, Nakanishi, Hoffmann, Pastor, and
  Schaal]{ijspeert2013dynamical}
A.~J. Ijspeert, J.~Nakanishi, H.~Hoffmann, P.~Pastor, and S.~Schaal.
\newblock Dynamical movement primitives: learning attractor models for motor
  behaviors.
\newblock \emph{Neural computation}, 25\penalty0 (2):\penalty0 328--373, 2013.

\bibitem[Zhu et~al.(2017)Zhu, Mottaghi, Kolve, Lim, Gupta, Fei-Fei, and
  Farhadi]{zhu2016target}
Y.~Zhu, R.~Mottaghi, E.~Kolve, J.~J. Lim, A.~Gupta, L.~Fei-Fei, and A.~Farhadi.
\newblock Target-driven visual navigation in indoor scenes using deep
  reinforcement learning.
\newblock In \emph{ICRA}, 2017.

\bibitem[Mirowski et~al.(2017)Mirowski, Pascanu, Viola, Soyer, Ballard, Banino,
  Denil, Goroshin, Sifre, Kavukcuoglu, et~al.]{mirowski2016learning}
P.~Mirowski, R.~Pascanu, F.~Viola, H.~Soyer, A.~Ballard, A.~Banino, M.~Denil,
  R.~Goroshin, L.~Sifre, K.~Kavukcuoglu, et~al.
\newblock Learning to navigate in complex environments.
\newblock In \emph{ICLR}, 2017.

\bibitem[Gupta et~al.(2017)Gupta, Davidson, Levine, Sukthankar, and
  Malik]{gupta2017cognitive}
S.~Gupta, J.~Davidson, S.~Levine, R.~Sukthankar, and J.~Malik.
\newblock Cognitive mapping and planning for visual navigation.
\newblock In \emph{CVPR}, 2017.

\bibitem[Sadeghi(2019)]{sadeghi2019divis}
F.~Sadeghi.
\newblock Divis: Domain invariant visual servoing for collision-free goal
  reaching.
\newblock \emph{RSS}, 2019.

\bibitem[Sadeghi and Levine(2016)]{sadeghi2016cad2rl}
F.~Sadeghi and S.~Levine.
\newblock Cad2rl: Real single-image flight without a single real image.
\newblock \emph{arXiv preprint arXiv:1611.04201}, 2016.

\bibitem[Levine et~al.(2016)Levine, Finn, Darrell, and Abbeel]{levine2016end}
S.~Levine, C.~Finn, T.~Darrell, and P.~Abbeel.
\newblock End-to-end training of deep visuomotor policies.
\newblock \emph{JMLR}, 2016.

\bibitem[Levine et~al.(2018)Levine, Pastor, Krizhevsky, Ibarz, and
  Quillen]{levine2018learning}
S.~Levine, P.~Pastor, A.~Krizhevsky, J.~Ibarz, and D.~Quillen.
\newblock Learning hand-eye coordination for robotic grasping with deep
  learning and large-scale data collection.
\newblock \emph{The International Journal of Robotics Research}, 37\penalty0
  (4-5):\penalty0 421--436, 2018.

\bibitem[Pathak et~al.(2017)Pathak, Agrawal, Efros, and
  Darrell]{pathak2017curiosity}
D.~Pathak, P.~Agrawal, A.~A. Efros, and T.~Darrell.
\newblock Curiosity-driven exploration by self-supervised prediction.
\newblock In \emph{ICML}, 2017.

\bibitem[Eysenbach et~al.(2019)Eysenbach, Gupta, Ibarz, and
  Levine]{eysenbach2018diversity}
B.~Eysenbach, A.~Gupta, J.~Ibarz, and S.~Levine.
\newblock Diversity is all you need: Learning skills without a reward function.
\newblock In \emph{ICLR}, 2019.

\bibitem[Argall et~al.(2009)Argall, Chernova, Veloso, and
  Browning]{argall2009survey}
B.~D. Argall, S.~Chernova, M.~Veloso, and B.~Browning.
\newblock A survey of robot learning from demonstration.
\newblock \emph{Robotics and Autonomous systems}, 2009.

\bibitem[Billard et~al.(2008)Billard, Calinon, Dillmann, and
  Schaal]{billard2008robot}
A.~Billard, S.~Calinon, R.~Dillmann, and S.~Schaal.
\newblock Robot programming by demonstration.
\newblock In \emph{Springer Handbook of robotics}. 2008.

\bibitem[Hausman et~al.(2017)Hausman, Chebotar, Schaal, Sukhatme, and
  Lim]{hausman2017multi}
K.~Hausman, Y.~Chebotar, S.~Schaal, G.~Sukhatme, and J.~J. Lim.
\newblock Multi-modal imitation learning from unstructured demonstrations using
  generative adversarial nets.
\newblock In \emph{NIPS}, 2017.

\bibitem[Aytar et~al.(2018)Aytar, Pfaff, Budden, Paine, Wang, and
  de~Freitas]{aytar2018playing}
Y.~Aytar, T.~Pfaff, D.~Budden, T.~L. Paine, Z.~Wang, and N.~de~Freitas.
\newblock Playing hard exploration games by watching youtube.
\newblock \emph{arXiv preprint arXiv:1805.11592}, 2018.

\bibitem[Torabi et~al.(2018)Torabi, Warnell, and Stone]{torabi2018behavioral}
F.~Torabi, G.~Warnell, and P.~Stone.
\newblock Behavioral cloning from observation.
\newblock In \emph{IJCAI}, 2018.

\bibitem[Edwards et~al.(2018)Edwards, Sahni, Schroeker, and
  Isbell]{edwards2018imitating}
A.~D. Edwards, H.~Sahni, Y.~Schroeker, and C.~L. Isbell.
\newblock Imitating latent policies from observation.
\newblock \emph{arXiv preprint arXiv:1805.07914}, 2018.

\bibitem[Pathak et~al.(2018)Pathak, Mahmoudieh, Luo, Agrawal, Chen, Shentu,
  Shelhamer, Malik, Efros, and Darrell]{pathak2018zero}
D.~Pathak, P.~Mahmoudieh, G.~Luo, P.~Agrawal, D.~Chen, Y.~Shentu, E.~Shelhamer,
  J.~Malik, A.~A. Efros, and T.~Darrell.
\newblock Zero-shot visual imitation.
\newblock In \emph{ICLR}, 2018.

\bibitem[Finn et~al.(2017)Finn, Yu, Zhang, Abbeel, and Levine]{finn2017one}
C.~Finn, T.~Yu, T.~Zhang, P.~Abbeel, and S.~Levine.
\newblock One-shot visual imitation learning via meta-learning.
\newblock \emph{arXiv preprint arXiv:1709.04905}, 2017.

\bibitem[Yu et~al.(2018)Yu, Finn, Xie, Dasari, Zhang, Abbeel, and
  Levine]{yu2018one}
T.~Yu, C.~Finn, A.~Xie, S.~Dasari, T.~Zhang, P.~Abbeel, and S.~Levine.
\newblock One-shot imitation from observing humans via domain-adaptive
  meta-learning.
\newblock \emph{arXiv preprint arXiv:1802.01557}, 2018.

\bibitem[Dayan and Hinton(1993)]{dayan1993feudal}
P.~Dayan and G.~E. Hinton.
\newblock Feudal reinforcement learning.
\newblock In \emph{NIPS}, 1993.

\bibitem[Barto and Mahadevan(2003)]{barto2003recent}
A.~G. Barto and S.~Mahadevan.
\newblock Recent advances in hierarchical reinforcement learning.
\newblock \emph{Discrete event dynamic systems}, 2003.

\bibitem[Vezhnevets et~al.(2017)Vezhnevets, Osindero, Schaul, Heess, Jaderberg,
  Silver, and Kavukcuoglu]{vezhnevets2017feudal}
A.~S. Vezhnevets, S.~Osindero, T.~Schaul, N.~Heess, M.~Jaderberg, D.~Silver,
  and K.~Kavukcuoglu.
\newblock Feudal networks for hierarchical reinforcement learning.
\newblock \emph{arXiv preprint arXiv:1703.01161}, 2017.

\bibitem[Levy et~al.(2017)Levy, Platt, and Saenko]{levy2017hierarchical}
A.~Levy, R.~Platt, and K.~Saenko.
\newblock Hierarchical actor-critic.
\newblock \emph{arXiv preprint arXiv:1712.00948}, 2017.

\bibitem[Fouhey et~al.(2014)Fouhey, Delaitre, Gupta, Efros, Laptev, and
  Sivic]{fouhey2014people}
D.~F. Fouhey, V.~Delaitre, A.~Gupta, A.~A. Efros, I.~Laptev, and J.~Sivic.
\newblock People watching: Human actions as a cue for single view geometry.
\newblock \emph{IJCV}, 2014.

\bibitem[Jordan and Rumelhart(1992)]{jordan1992forward}
M.~I. Jordan and D.~E. Rumelhart.
\newblock Forward models: Supervised learning with a distal teacher.
\newblock \emph{Cognitive science}, 16\penalty0 (3):\penalty0 307--354, 1992.

\bibitem[Agrawal et~al.(2016)Agrawal, Nair, Abbeel, Malik, and
  Levine]{agrawal2016learning}
P.~Agrawal, A.~V. Nair, P.~Abbeel, J.~Malik, and S.~Levine.
\newblock Learning to poke by poking: Experiential learning of intuitive
  physics.
\newblock In \emph{NIPS}, pages 5074--5082, 2016.

\bibitem[Jang et~al.(2016)Jang, Gu, and Poole]{jang2016categorical}
E.~Jang, S.~Gu, and B.~Poole.
\newblock Categorical reparameterization with gumbel-softmax.
\newblock \emph{arXiv preprint arXiv:1611.01144}, 2016.

\bibitem[Armeni et~al.(2016)Armeni, Sener, Zamir, Jiang, Brilakis, Fischer, and
  Savarese]{armeni20163d}
I.~Armeni, O.~Sener, A.~R. Zamir, H.~Jiang, I.~Brilakis, M.~Fischer, and
  S.~Savarese.
\newblock {3D} semantic parsing of large-scale indoor spaces.
\newblock In \emph{CVPR}, 2016.

\bibitem[Chang et~al.(2017)Chang, Dai, Funkhouser, Halber, Niessner, Savva,
  Song, Zeng, and Zhang]{Matterport3D}
A.~Chang, A.~Dai, T.~Funkhouser, M.~Halber, M.~Niessner, M.~Savva, S.~Song,
  A.~Zeng, and Y.~Zhang.
\newblock {Matterport3D}: Learning from {RGB-D} data in indoor environments.
\newblock In \emph{3DV}, 2017.

\bibitem[Kumar* et~al.(2018)Kumar*, Gupta*, Fouhey, Levine, and
  Malik]{kumar2018visual}
A.~Kumar*, S.~Gupta*, D.~Fouhey, S.~Levine, and J.~Malik.
\newblock Visual memory for robust path following.
\newblock In \emph{Advances in Neural Information Processing Systems}, 2018.

\bibitem[Swedish and Raskar(2018)]{swedish2018deep}
T.~Swedish and R.~Raskar.
\newblock Deep visual teach and repeat on path networks.
\newblock In \emph{CVPRW}, 2018.

\bibitem[Anderson et~al.(2018)Anderson, Chang, Chaplot, Dosovitskiy, Gupta,
  Koltun, Kosecka, Malik, Mottaghi, Savva, and Zamir]{anderson2018evaluation}
P.~Anderson, A.~Chang, D.~S. Chaplot, A.~Dosovitskiy, S.~Gupta, V.~Koltun,
  J.~Kosecka, J.~Malik, R.~Mottaghi, M.~Savva, and A.~Zamir.
\newblock On evaluation of embodied navigation agents.
\newblock \emph{arXiv preprint arXiv:1807.06757}, 2018.

\bibitem[Kingma and Ba(2014)]{kingma2014adam}
D.~P. Kingma and J.~Ba.
\newblock Adam: A method for stochastic optimization.
\newblock \emph{arXiv preprint arXiv:1412.6980}, 2014.

\bibitem[Murali et~al.(2019)Murali, Chen, Alwala, Gandhi, Pinto, Gupta, and
  Gupta]{pyrobot2019}
A.~Murali, T.~Chen, K.~V. Alwala, D.~Gandhi, L.~Pinto, S.~Gupta, and A.~Gupta.
\newblock Pyrobot: An open-source robotics framework for research and
  benchmarking.
\newblock \emph{arXiv preprint arXiv:1906.08236}, 2019.

\end{thebibliography}
